\documentclass{article}
\usepackage{arxiv}

\usepackage[utf8]{inputenc} 
\usepackage[T1]{fontenc}    
\usepackage{hyperref}       
\usepackage{url}            
\usepackage{booktabs}       
\usepackage{amsfonts}       
\usepackage{nicefrac}       
\usepackage{microtype}      
\usepackage{graphicx}
\usepackage{natbib}
\usepackage{doi}

\newcommand{\location}{\boldsymbol{x}}
\newcommand{\locb}{y}
\newcommand{\locc}{z}
\newcommand{\locationb}{\boldsymbol{\locb}}
\newcommand{\locationc}{\boldsymbol{\locc}}
\newcommand{\direction}{\boldsymbol{\omega}}
\newcommand{\dangle}{\theta}
\newcommand{\danglei}{\theta^{\regpar{\pathindex}}}
\newcommand{\voxelset}{\mathcal{V}}
\newcommand{\voxel}{v}
\newcommand{\linesegment}[2]{\overline{{#1}{#2}}}
\newcommand{\hemisphere}{\Omega} 
\newcommand{\sphere}{\mathcal{S}^2} 
\newcommand{\ppath}{\Bar{\boldsymbol{x}}}

\newcommand{\pathsize}{B}
\newcommand{\pathindex}{i}
\newcommand{\pathi}{\ppath^{\pathindex}}
\newcommand{\segind}{b}
\newcommand{\locationi}{\location^{\regpar{i}}}
\newcommand{\directioni}{\direction^{\regpar{i}}}
\newcommand{\pathset}{\mathcal{P}}
\newcommand{\geomterm}{G}
\newcommand{\lightloc}{\location_{\text{light}}}
\newcommand{\sensloc}{\location_{\nm}}
\newcommand{\sensdir}{\direction_{\nm}}
\newcommand{\sensbdir}{\direction_{\segind, d}}

\newcommand{\lastloc}{\locationi_{\text{last}}}
\newcommand{\ext}{\beta}
\newcommand{\exts}{\ext_{\text{s}}}
\newcommand{\exta}{\ext_{\text{a}}}
\newcommand{\extt}{\ext}
\newcommand{\extsat}{\exts\regpar{\location}}
\newcommand{\extaat}{\exta\regpar{\location}}
\newcommand{\exttat}{\extt\regpar{\location}}
\newcommand{\sinscat}{\varpi}
\newcommand{\sinscatpar}[1]{\sinscat^{\regpar{#1}}}
\newcommand{\extcloud}{\ext^{\text{c}}}
\newcommand{\extair}{\ext^{\text{a}}}
\newcommand{\extscloud}{\exts^{\text{c}}}
\newcommand{\extsair}{\exts^{\text{a}}}
\newcommand{\sinscatcloud}{\sinscat_{\text{c}}}
\newcommand{\sinscatair}{\sinscat_{\text{a}}}

\newcommand{\ffunc}{\mathsf{f}}
\newcommand{\brdf}{\ffunc_{\text{r}}}
\newcommand{\phasefunc}{\ffunc_{\text{p}}}
\newcommand{\phasefunccloud}{\ffunc^{\text{c}}_{\text{p}}}
\newcommand{\phasefuncair}{\ffunc^{\text{a}}_{\text{p}}}
\newcommand{\anglefunc}{\ffunc_{\text{s}}}
\newcommand{\distance}{l}
\newcommand{\ptype}{j}

\newcommand{\nm}{d}
\newcommand{\Np}{N} 
\newcommand{\Npar}{N_{\text{par}}} 
\newcommand{\Nm}{N_{\text{\nm}}} 
\newcommand{\Nr}{N_{\text{r}}} 
\newcommand{\Nu}{N_{\text{u}}} 

\newcommand{\sceneparam}{\boldsymbol{\mathcal{M}}}
\newcommand{\sceneparamt}{\sceneparam^{\iter}}
\newcommand{\sceneparef}{\sceneparam_{\text{ref}}}
\newcommand{\scenescalar}{m}
\newcommand{\lossfunc}{\mathcal{L}}
\newcommand{\gtmeasurements}{\boldsymbol{I}^{\text{gt}}}
\newcommand{\gtscalar}{I^{\text{gt}}}
\newcommand{\scorefunc}{S}
\newcommand{\correctfact}{r}
\newcommand{\iter}{t}
\newcommand{\maxiter}{\mathcal{T}}
\newcommand{\unknownindex}{v}
\newcommand{\gradest}{\boldsymbol{\mathcal{G}}}
\newcommand{\momentdecay}{\eta}
\newcommand{\stepsize}{\alpha}
\newcommand{\reldist}{\epsilon}
\newcommand{\relbias}{\delta}

\newcommand{\radiance}{L}
\newcommand{\radiances}{L_{\text{s}}}
\newcommand{\radiancee}{L_{\text{e}}}
\newcommand{\radianceat}{\radiance\regpar{\location,\direction}}
\newcommand{\radiancesat}{\radiances\regpar{\location,\direction}}
\newcommand{\radianceeat}{\radiancee\regpar{\location,\direction}}

\newcommand{\normal}{\textbf{n}} 

\newcommand{\contfunc}{f}
\newcommand{\segfunc}{g}
\newcommand{\pixmeasure}{W}
\newcommand{\ray}{\mathcal{R}}

\newcommand{\trans}[2]{T\regpar{{#1},{#2}}}

\newcommand{\ispdf}{\mu}
\newcommand{\forwardmodel}{\mathcal{F}}

\newcommand{\mcpar}{u}
\newcommand{\unipdf}{p_{\mathcal{U}}}
\newcommand{\nexteventpdf}{P_\text{le}}

\newcommand{\regpar}[1]{\left({#1}\right)}
\newcommand{\squarepar}[1]{\left[{#1}\right]}
\newcommand{\curlpar}[1]{\left\{ {#1}\right\}}
\newcommand{\norm}[1]{\left|\left| {#1} \right|\right|}
\newcommand{\abs}[1]{\left| {#1} \right|}
\newcommand{\derivx}[2]{\dfrac{\partial {#1}}{\partial {#2}}}
\newcommand{\deriv}[1]{\dfrac{\partial}{\partial {#1}}}

\newcommand{\simiid}{\sim}

\newcommand{\smokereldist}{25}
\newcommand{\smokerelbias}{6}

\newcommand{\jplreldist}{53.8}
\newcommand{\jplrelbias}{-1.5}
\newcommand{\smallcfreldist}{52.5}
\newcommand{\smallcfrelbias}{0.07}

\newcommand{\eq}[1]{Eq.~({#1})}
\newcommand{\eqs}[1]{Eqs.~({#1})}
\newcommand{\eqnopar}[1]{Eq.~{#1}}
\newcommand{\eqsnopar}[1]{Eqs.~{#1}}
\newcommand{\fig}{Fig.~}
\newcommand{\figs}[1]{Figs.~{#1}}
\newcommand{\alg}{Alg.~}

\usepackage{tikz}
\usepackage{comment}
\usepackage{color}
\usepackage{xcolor}
\usepackage{mathrsfs}
\usepackage{bbm}
\usepackage{mathtools}
\usepackage{gensymb}
\usepackage{acronym}
\usepackage[ruled,vlined]{algorithm2e}
\usepackage{cancel}
\usepackage{float}
\usepackage{tikz}
\usetikzlibrary{shapes,arrows}
\usepackage{longtable}
\usepackage{makecell}
\usepackage{amssymb}

\newacro{RTE}{Radiative Transfer Equation}
\newacro{MC}{Monte-Carlo}
\newacro{SRE}{Surface Rendering Equation}
\newacro{PDF}{Probability Density Function}
\newacro{BRDF}{Bidirectional Reflectance Distribution Function}
\newacro{I.I.D}{Independent and Identically Distributed}
\newacro{GPU}{Graphics Processing Unit}
\newacro{JIT}{Just In Time}
\newacro{ADAM}{Adaptive Moment Estimation}
\newacro{LES}{Large Eddy Simulation}
\newacro{AD}{Automatic Differentiation}
\newacro{SIMD}{Single Instruction Multiple Data}
\newacro{MLT}{Metropolis Light Transport}
\newacro{CT}{Computed Tomography}
\newacro{MCF}{Measurement Contribution function}
\newacro{SCF}{Segment Contribution function}
\newacro{PSF}{Path Score Function}
\newacro{IFOV}{Instantaneous Field of View}
\title{Accelerating Inverse Rendering\\ By Using a GPU and Reuse of Light Paths}

\author{ 
Ido Czerninski, Yoav Y.~Schechner \\
	Viterbi Faculty of Electrical and Computer Engineering\\
	Technion – Israel Institute of Technology\\
	\texttt{idocz@campus.technion.ac.il  yoav@ee.technion.ac.il} \\
}

\renewcommand{\shorttitle}{\textit{arXiv} Template}

\hypersetup{
pdftitle={A template for the arxiv style},
pdfsubject={},
pdfauthor={Ido Czerninski, Yoav Y.~Schechner},
pdfkeywords={First keyword, Second keyword, More},
}

\begin{document}
\maketitle

\begin{abstract}
Inverse rendering seeks to estimate scene characteristics from a set of data images. The dominant approach is based on differential rendering using Monte-Carlo. Algorithms as such usually rely on a forward model and use an iterative gradient method that requires sampling millions of light paths per iteration. This paper presents an efficient framework that speeds up existing inverse rendering algorithms. This is achieved by tailoring the iterative process of inverse rendering specifically to a \ac{GPU} architecture. For this cause, we introduce two interleaved steps - Path Sorting and Path Recycling. Path Sorting allows the \ac{GPU} to deal with light paths of the same size. Path Recycling allows the algorithm to use light paths from previous iterations to better utilize the information they encode. Together, these steps significantly speed up  gradient optimization. In this paper, we give the theoretical background for Path Recycling. We demonstrate its efficiency for volumetric scattering tomography and reflectometry (surface reflections).
\end{abstract}


\section{INTRODUCTION}
\label{sec:intro}
In computer graphics, rendering is referred to as the process of computationally generating an image of a given scene. Previous work \citep{kajiya1986rendering} showed that image formation can be estimated by \ac{MC} techniques, by tracing light paths from a light source to a camera in the scene. This method is called {\em Path Tracing}. Since light paths are independent, path tracing can be parallelized over computing resources. Path tracing allowed rendering of photorealistic images that are faithful to real-life lighting. Recent work showed that {\em path tracing} equations are differentiable, and therefore can be used to solve inverse problems. 

Methods that solve inverse problems are usually based on a differentiable forward model. A differential forward model can estimate how small perturbations of the scene parameters affect the estimated image. With this knowledge, inverse rendering can be formulated as an optimization problem that finds the scene parameters that best agree with a set of ground-truth measurements. However, forward models have not been optimized to serve as a black box in a gradient algorithm. Moreover, randomness in \ac{MC} implementation of the forward model inhibits \ac{GPU} architectures from fully utilizing the parallelism of the problem.  

In this paper, we counter these problems by exploiting the mathematical foundation of path tracing in a way that is tailored well to a \ac{GPU}. This is done by two elements: Path Sorting and Path Recycling. Path sorting is a simple step that extracts coherent work by sorting the light paths with respect to their size. This step assures that a \ac{GPU} better utilizes its parallelized structure. The problem with this step is that the path size is a random quantity, which is not known in advance. To counter this, we introduce a  path recycling method, which is the main contribution of this paper. Path recycling samples a set of light paths once and reuses them in future iterations. This step reveals the light path size, and by doing so, eliminates the randomness of the path size. Moreover, each gradient iteration only slightly changes the scene parameters. This maintains the usefulness of paths from previous iterations. 

We implemented a differentiable rendering engine that uses path sorting and path recycling to accelerate inverse rendering algorithms. We demonstrate our contribution on scattering tomography and reflectometry. Moreover, we take a step towards more realistic scattering tomography, by providing gradient derivation and implementation for multi-type particles. 

\ac{CT} addresses the challenge of reconstructing a volumetric medium. Tomography is used in various domains, such as medical imaging, atmospheric science, astrophysics, and geophysics. Specifically, medical \ac{CT} aims to see-through a human body in 3D, non-invasively . Another example from atmospheric science is cloud tomography. Cloud tomography aims to retrieve a 3D distribution of physical properties of a cloud by using the sun as the radiation source. Performing \ac{CT} that considers scattered radiation is complex due to the various paths the radiation can traverse.

In medical \ac{CT}, the complex problem is simplified by considering only the direct component of the radiation. This is made possible in case of X-ray \ac{CT} by using an anti-scatter grid. This approach, however, dismisses information encoded by the scattered radiation. In cloud retrieval, simplification is usually made by assuming that the atmosphere consists of laterally-homogeneous layers. This assumption does not capture the 3D nature of the atmosphere. Addressing these limitations has great importance. It can significantly reduce the required radiation dose in medical \ac{CT} and allow potentially more accurate climate prediction.

The paper is organized as follows: Section \ref{sec:related} reviews related work. Section \ref{sec:forward} gives the relevant background on the image formation model and how it can be estimated using \ac{MC} integration. Section \ref{sec:inverse} defines inverse rendering as an optimization problem and shows how the gradient of an error measure can be computed, to allow gradient-based solutions. Section \ref{sec:recycling} introduces path recycling and provides the relevant theoretical justifications. Section \ref{sec:simulations} shows simulation results of path recycling on two inverse rendering problems. Section \ref{sec:conclusion} discusses limitations and suggests future research directions.

\section{RELATED WORK}
\label{sec:related}
{\em Differentiable rendering.} The dominant approach to inverse rendering is to differentiate through the rendering process for gradient-based optimization.  Differential rendering has been used in various domains. \citep{loper2014opendr} suggested an approximated differential rendering framework that uses \ac{AD}. \ac{AD} is a method to automatically compute gradients of a computer program. Refs \citep{gkioulekas2013inverse, gkioulekas2016evaluation} use symbolic differentiation, which uses analytical derivatives, to retrieve volumetric properties of a medium. \citep{geva2018x}  used symbolic differentiable rendering to perform X-ray \ac{CT} that considers scattered radiation. \citep{loeub2020monotonicity} suggested symbolic differentiable rendering together with a monotonicity prior, to preform cloud tomography.

\citep{nimier2019mitsuba} introduced Mitsuba 2 - a physics-based rendering engine that runs on a \ac{GPU} and supports various lighting effects such as polarized light, caustics, gradient-index optics, and volumetric scattering. Moreover, it supports differentiable rendering by using a reverse mode \ac{AD}, which is equivalent to the backpropagation method from deep learning. While it solves a variety of problems, the reverse mode of Mitsuba 2 requires storing a computation transcript for backpropagating the gradient back to the scene parameters. This creates a significant memory bottleneck when differentiating over a large set of scene parameters. \citep{nimier2020radiative} countered this by observing that analytical derivatives can be addressed as the solution of a modified light transport problem. They achieved up to $\times1000$ speed-up compared to Mitsuba 2. 

{\em Wavefront vs mega-kernels}. There are two approaches for implementing path tracing on a \ac{GPU}: mega-kernel and wavefront. A Mega-kernel traces several paths as wholes using a single kernel execution. Wavefront rendering traces a batch of paths, one segment at a time, using multiple micro-kernels for each segment. \citep{laine2013megakernels} suggested a wavefront approach, arguing that mega-kernels do not suit the parallelism structure and resources of a \ac{GPU} . The wavefront approach has become widely used \citep{lee2017vectorized, fascione2018manuka}. Mitsuba 2 also traces paths using the wavefront approach. However, tracing multiple paths while recording a transcript at the same time exhausts the \ac{GPU} memory capacity. Therefore, \citep{nimier2020radiative} discarded the wavefront approach and transferred Mitsuba 2 to the mega-kernel formulation. This may imply that while a mega-kernel approach has drawbacks in forward rendering, it can be preferable for differential rendering. Our implementation follows a similar mega-kernel approach.

{\em Coherent rendering.} Extracting coherent work is the process of dividing computations of a computer program into parallelable segments, each executing similar commands. Modern computing architectures that are based on \ac{SIMD}, such as \ac{GPU}s, require coherent work to fully utilize their resources. However, the randomness of \ac{MC} path tracing is inherently incoherent. Coherent rendering counters this randomness by extracting coherent work as a subset of the rendering process. Since the wavefront approach divides the work into multiple steps, it naturally enables more freedom for extracting coherent work between steps.

\citep{afra2016local} generated coherent work on a wavefront path tracer by sorting path segments with respect to the material type. \citep{nimier2019mitsuba} introduces a novel \ac{MLT} method that directly generates coherent work without the need of a sorting process. Nevertheless, by transferring Mitsuba 2 into a mega-kernel, \citep{nimier2020radiative} achieved a significant speed-up but lost coherency. 
Our path recycling framework is capable of {\em creating coherent work by sorting the paths according to their size.} This is without using a memory-hungry wavefront approach.

{\em Scattering Tomography}. Recent work addressed the recovery limitations stated of Sec.~\ref{sec:intro}. \citep{levis2015airborne} suggested solving scattering \ac{CT} using a deterministic approach, which uses a spherical harmonic representation of the \ac{RTE}. This method is capable of solving cloud tomography while considering complicated physical quantities, such as the effective radius of the droplets in the cloud. However, this method requires high memory resources, which does not scale to large cloud fields.

Another approach is based on \ac{MC} radiative transfer. \citep{holodovsky2016situ} and \citep{loeub2020monotonicity} suggested an approach based on {\em path tracing}, which fits a volumetric model to the detectors' measurements using gradient-based optimization. An advantage of this approach is that it considers multiple scattering events while allowing an arbitrary medium as its input. \citep{geva2018x} showed that medical \ac{CT} can be performed more accurately with less radiation dose, by using a similar approach. 

While previous work suggested effective ways to perform scattering tomography, they suffer from high computational demands. This arises from sampling many light paths repeatedly in each gradient iteration. In this paper, we address the above problem by suggesting an efficient framework for \ac{MC} based optimization.

\section{BACKGROUND}
\label{sec:forward}
This section provides essential theoretical background of image formation and how it can be estimated by \ac{MC} integration. This background is essential for understanding the basis of various inverse rendering algorithms that come next. This section is inspired by \citep{novak2018monte}.

\subsection{Optical Properties of Surfaces and Scattering Media}
Before presenting the RTE, we describe ray propagation through a scattering medium and on reflection by an opaque surface. In this section, $\location$, $\locationb$ and $\locationc$ are 3D locations and $\direction$ is a direction vector.

\subsubsection{Bidirectional Reflectance Distribution Function}
\begin{figure}[t]
  \centering
  \includegraphics[width=0.6\linewidth]{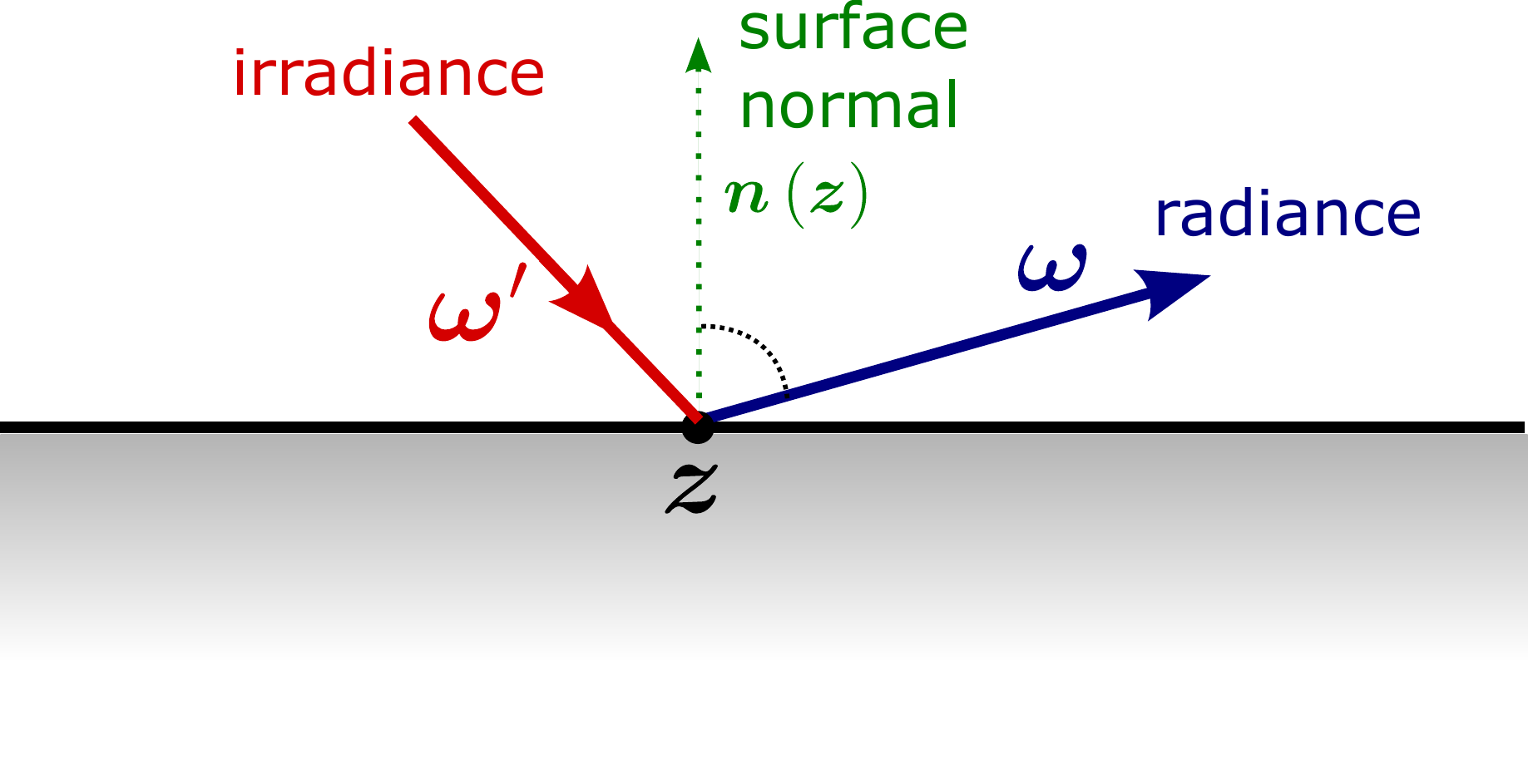}
    \caption{Illustration of the notations in \eq{\ref{eq:brdf_conservation}}. In the graphics literature, $\direction'$ is often defined as the negative irradiance direction. We define $\direction'$ as the irradiance direction, in order to be consistent with definitions in the literature on scattering.}
    \label{fig:BRDF}
\end{figure}

The \ac{BRDF} of a surface is a function of location $\locationc$, irradiance direction $\direction'$ and radiance direction $\direction$. Let $\hemisphere$ be the unit hemisphere and $\normal\regpar{\locationc}$ be the surface normal at $\locationc$ (illustrated in \fig\ref{fig:BRDF}). The \ac{BRDF} is denoted by $\brdf\regpar{\locationc, \direction'\rightarrow\direction}$.  Upon hitting on opaque surface, light can be either reflected or absorbed. Hence, \citep{veach1997robust},
\begin{equation}
\label{eq:brdf_conservation}
\intop_\hemisphere \brdf\regpar{\locationc, \direction'\rightarrow\direction}\regpar{\normal\regpar{\locationc} \cdot \direction}d\direction \leq 1.
\end{equation}

\subsubsection{Scattering and Absorption Coefficient.}
A scattering medium consists of microscopic particles which can absorb or scatter light. The medium has a scattering coefficient $\exts\regpar{\location}\squarepar{m^{-1}}$ and an absorption coefficient $\exta\regpar{\location}\squarepar{m^{-1}}$. The total extinction coefficient $\extt\regpar{\location}=\exts\regpar{\location}+\exta\regpar{\location}$ expresses the total radiance attenuation due to both scattering and absorption events. 

A medium may consist of $\Npar$ types of particles. Had the volume element contained exclusively only particles particles of type $\ptype$, the respective scattering and absorption coefficients would have been $\exts^{\regpar{\ptype}}\regpar{\location}$ and $\exta^{\regpar{\ptype}}\regpar{\location}$. Then  
\begin{equation}
\label{eq:ext_particles}
\extt\regpar{\location}=\sum_{\ptype=1}^{\Npar}{\extt^{\regpar{\ptype}}\regpar{\location}}= \sum_{\ptype=1}^{\Npar}{\squarepar{\exts^{\regpar{\ptype}}\regpar{\location}+\exta^{\regpar{\ptype}}\regpar{\location}}}.
\end{equation}

\subsubsection{Single Scattering Albedo.}
The single scattering albedo of the $\ptype$'th particle type is the ratio:
\begin{equation}
\label{eq:sinscat}
\sinscatpar{j} \triangleq \frac{\exts^{\regpar{\ptype}}\regpar{\location}}{\extt^{\regpar{\ptype}}\regpar{\location}}.
\end{equation}
In this paper, we assume that the single scattering albedo of a particle type is space-invariant. We define the {\em effective} single scattering albedo as follows:
\begin{equation}
\label{eq:eff_sinscat}
\sinscat \regpar{\location} \triangleq  \frac{\sum_{\ptype=1}^{\Npar}{\sinscatpar{\ptype}\extt^{\regpar{\ptype}}\regpar{\location}}}{\exttat} = 
\frac{\extsat}{\exttat} 
\end{equation}
Notice that an effective single scattering albedo depends on the location $\location$.

\subsubsection{Phase Function.}
Light propagating in direction $\direction'$ scatters to a new direction $\direction$. The probability density for this direction is set by a {\em phase function} denoted $\phasefunc\regpar{\direction\rightarrow\direction'}$. Let there be $\Npar$ types of scattering particles in the medium. The phase function associated with the $\ptype$'th particle type is $\phasefunc^{\regpar{\ptype}}$. Then, the total phase function is:
\begin{equation}
\label{eq:phasefunc_particles}
\phasefunc\regpar{\location, \direction'\rightarrow \direction}=\frac{\sum_{\ptype=1}^{\Npar}{\exts^{\regpar{\ptype}}\regpar{\location}\cdot \phasefunc^{\regpar{\ptype}}\regpar{\direction'\rightarrow\direction}}}{\sum_{\ptype=1}^{\Npar}{\exts^{\regpar{\ptype}}\regpar{\location}}} .
\end{equation}
Notice that in this case, the phase function depends on the location $\location$. The domain of the phase function is the unit sphere $\sphere$. It satisfies
\begin{equation}
\label{eq:phase_conservation}
\intop_{\sphere} \phasefunc\regpar{\location, \direction'\rightarrow\direction} d\direction = 1.
\end{equation}
Usually, the phase function is a function of $\cos\regpar{\dangle}=\direction\cdot\direction'$. Thus, we occasionally use the notation $\phasefunc\regpar{\location, \cos\regpar{\dangle}}$.

\subsubsection{Volumetric Emission.}
Volumetric emission of light by a medium can be caused, for example, by fluorescence. Fluorescence requires absorption of excitation light by the medium. After being excited by a photon, a fluorescent molecule relaxes to a lower energy by emitting a different photon. Another example of volumetric emission is through a chemical process, as combustion.

\subsection{The Radiative Transfer Equation}
\begin{figure}[t]
  \centering
  \includegraphics[width=0.9\linewidth]{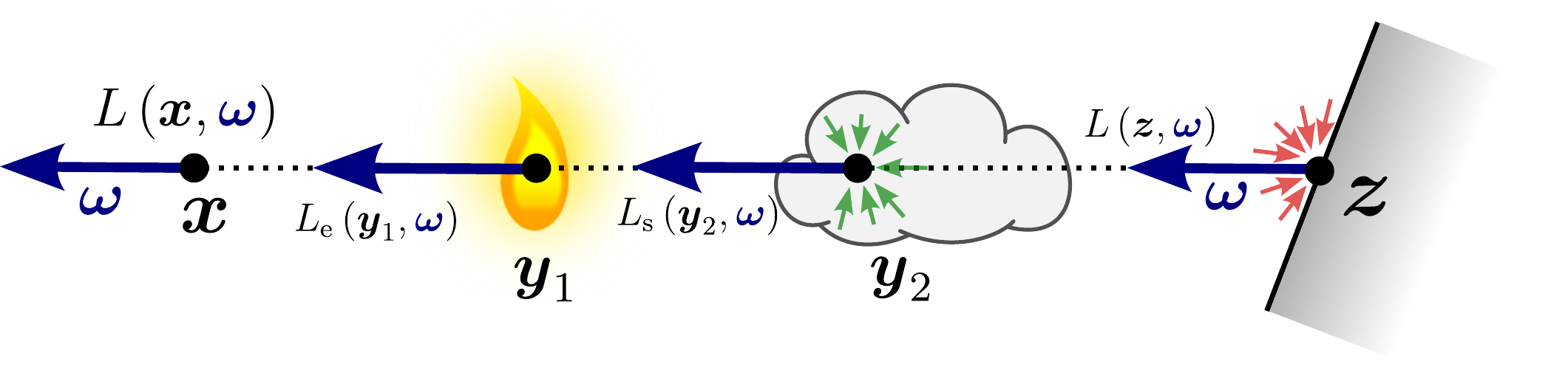}
    \caption{Illustration of \eq{\ref{eq:volume_rendering}}. This figure demonstrates how the radiance at location $\location$ and direction $\direction$ can be evaluated. A surface patch $\locationc$ collects irradiance from all directions on the unit hemisphere  (red arrows, \eqnopar{\ref{eq:surface_rendering}}) and reflects it towards $\direction$. Then, all volume elements $\locationb_2$ scatter light from all the directions $\direction'$ on the unit sphere (green arrows, \eqnopar{\ref{eq:inscattered_radiance})} towards $\direction$. In addition, emissive volume elements $\locationb_1$ contribute radiance in direction $\direction$.}
    \label{fig:volume_rendering}
\end{figure}

In this section, we first deal only with a scattering medium. Afterwards, we consider the effect of an opaque surface. The \ac{RTE} describes how radiance $\radiance(\location,\direction)$ changes in an infinitesimal volume element of a participating medium, due to absorption, emission and scattering events. Integrating incoming radiance from all directions of the unit sphere $\sphere$, the scattered radiance is
\begin{equation}
\label{eq:inscattered_radiance}
\radiancesat = \intop_{\sphere}{\phasefunc\regpar{\location,\direction'\rightarrow\direction} \radiance\regpar{\location,\direction'} d\direction'}.
\end{equation}
A ray starts at $-\infty$ and points in direction $\direction$ (Fig. \ref{fig:volume_rendering}). A location on a ray is
\begin{equation}
\label{eq:integrated_ray}
\locationb = \location + \locb\direction,  \quad  \locb\in\left(-\infty, 0\right).
\end{equation}
 The transmittance of the medium between points $\location$, $\locationb$ is
\begin{equation}
\label{eq:transmittance}
\trans{\location}{\locationb} = \exp{\regpar{-\intop_{\locb}^0{\extt\regpar{\location+\locb'\direction}d\locb'}}}.
\end{equation} 
Let $\radiancee$ be the radiance emitted by a unit volume in the medium. The \ac{RTE} is:
\begin{equation}
\label{eq:RTE}
\regpar{d\direction\cdot\nabla}\radianceat = -\radianceat\squarepar{\extaat + \extsat} +\extsat\radiancesat + \extaat\radianceeat .
\end{equation}
Integrating \eq{\ref{eq:RTE}} along (\ref{eq:integrated_ray}) results in an explicit formula for the radiance:
\begin{equation}
\label{eq:integrated_RTE}
\radianceat = \intop_{-\infty}^0{\trans{\location}{\locationb}\bigg[\exts\regpar{\locationb}\radiances\regpar{\locationb,\direction}} + \exta\regpar{\locationb}\radiancee\regpar{\locationb,\direction} \bigg]d\locb\quad \squarepar{\frac{\text{W}}{\text{sr}\cdot\text{m}^{2}}}.
\end{equation}
 
\eq{\ref{eq:integrated_RTE}} can be extended to consider surfaces by introducing the \ac{SRE} \citep{kajiya1986rendering}. Let $\locationc$ be a surface point with normal $\normal\regpar{\locationc}$. Let $\radiancee$ be the radiance emitted by the surface. The radiance at point $\locationc$ towards direction $\direction$ is given by:
 
\begin{equation}
\label{eq:surface_rendering}
\radiance\regpar{\locationc,\direction} = \radiancee\regpar{\locationc,\direction} +  \intop_{\hemisphere}{\brdf\regpar{\direction'\rightarrow\direction}\radiance\regpar{\locationc,\direction'}\abs{\normal\regpar{\locationc}\cdot\direction'}d\direction'}
\end{equation}
\eq{\ref{eq:surface_rendering}} can be used as a boundary condition on the integral in \eq{\ref{eq:integrated_RTE}}. Instead of integrating in \eq{\ref{eq:integrated_RTE}} the radiance from $-\infty$, the radiance can be integrated from the nearest surface point $\locationc=\location+\locc\direction$ (assuming there is one). Then, \eq{\ref{eq:integrated_RTE}} is modified as follows: 
\begin{equation}
\label{eq:volume_rendering}
\radianceat = \intop_{\locc^{-}}^0{\trans{\location}{\locationb}\bigg[\exts\regpar{\locationb}\radiances\regpar{\locationb,\direction}} + \exta\regpar{\locationb}\radiancee\regpar{\locationb,\direction} \bigg]d\locb +
\trans{\location}{\locationc}\radiance\regpar{\locationc,\direction}.
\end{equation} 

\begin{figure}[t]
  \centering
  \includegraphics[width=0.6\linewidth]{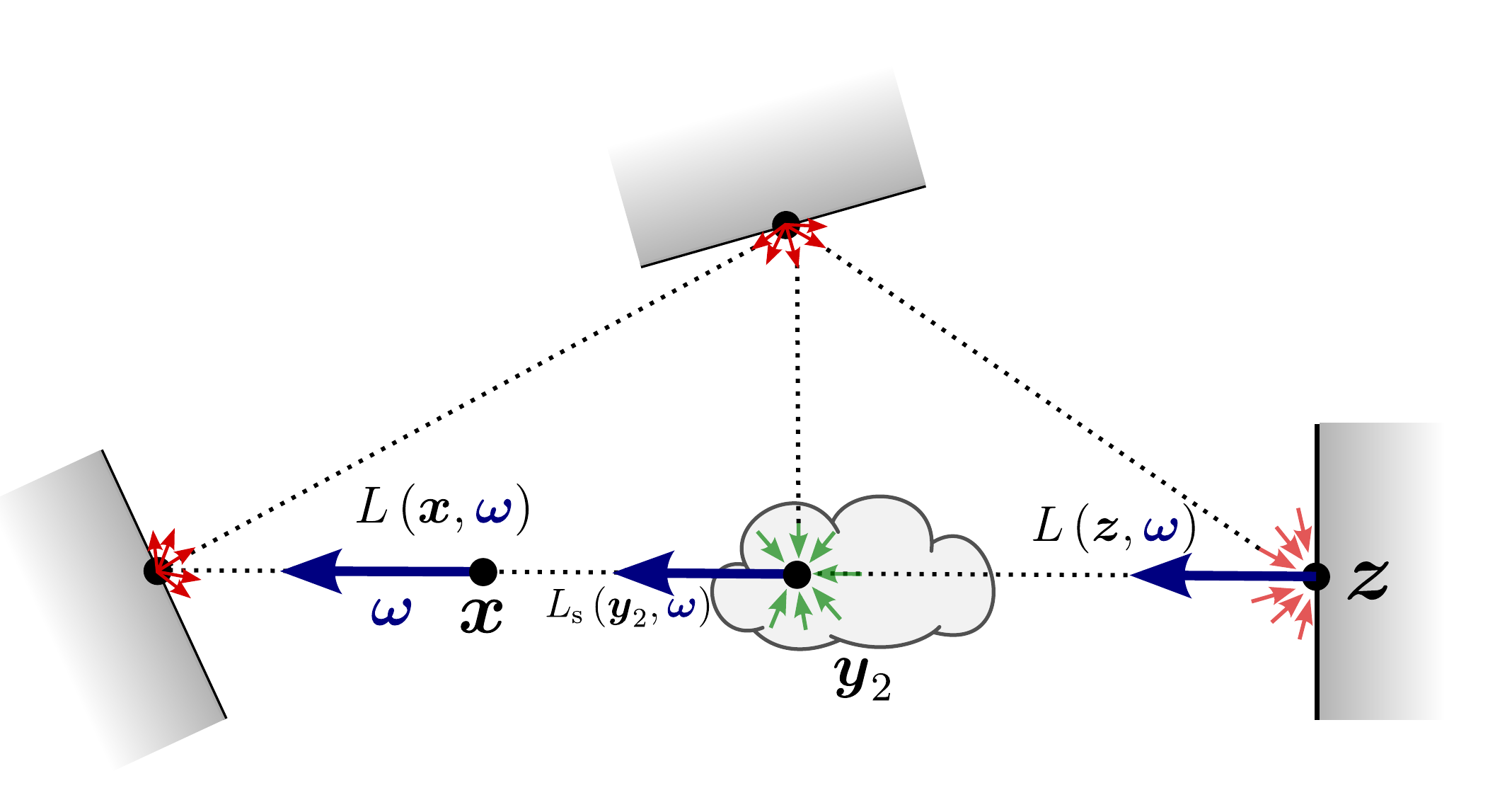}
    \caption{Recursion in the rendering equation. A portion of the light that exits $\location$ towards $\direction$ is reflected twice by the surfaces in the scene. Hence, the radiance $\radianceat$ may affect $\radiances\regpar{\locationb_2,\direction}$ and $\radiance\regpar{\locationc,\direction}$, which in turn affect $\radianceat$.
    While this figure demonstrates a recursion caused only by surface reflections, volumetric scattering, or a combination of the two, also causes a recursion.}  
    \label{fig:volume_recursion}
\end{figure}

\eq{\ref{eq:volume_rendering}} considers both a scattering medium and opaque surfaces as illustrated in \fig\ref{fig:volume_rendering}. There is a recursive dependency between $\radianceat$, $\radiances\regpar{\locationb_2,\direction}$ and $\radiances\regpar{\locationc,\direction}$ as illustrated in \fig\ref{fig:volume_recursion}.  Hence, \eq{\ref{eq:volume_rendering}} cannot be solved analytically. A solution can be sought by a path integral formulation \citep{veach1997robust} of light transport.

\subsection{Path Integral Formulation}
\begin{figure}[b]
  \centering
  \includegraphics[width=0.6\linewidth]{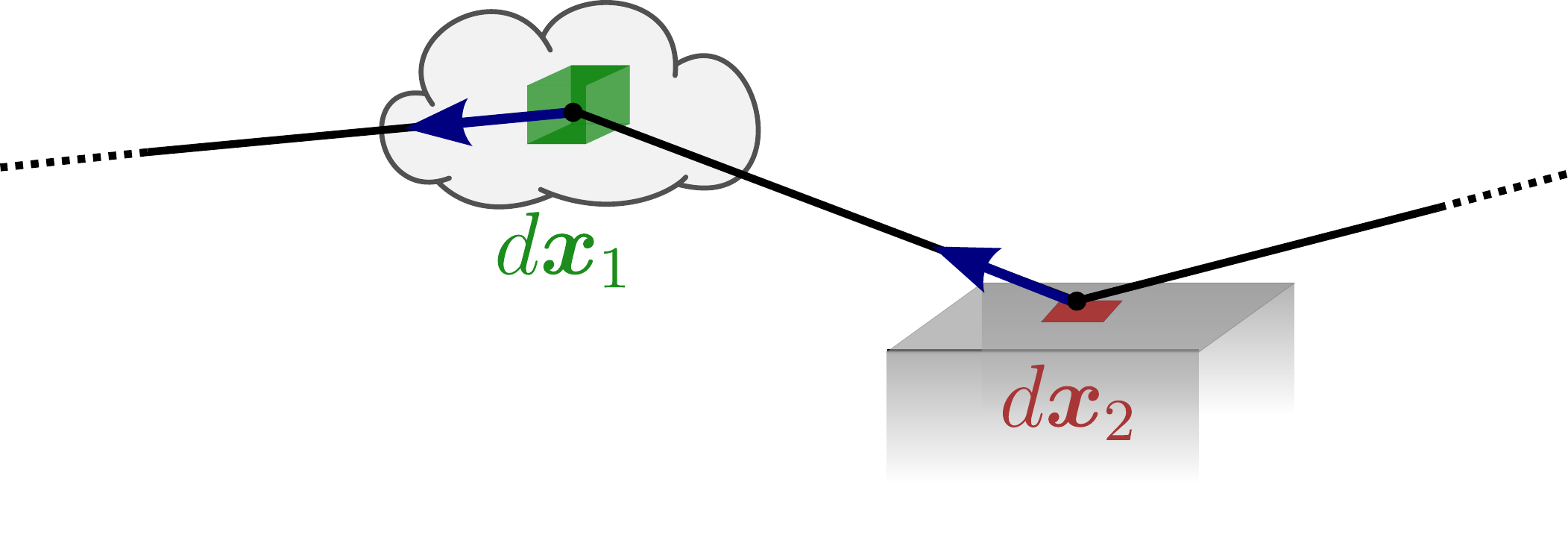}
    \caption{Illustration of a path differential. When a light path is reflected by a surface, the path integral uses a differential surface (red). When a light path is scattered by a medium, the path integral uses a differential volume (green).}  
    \label{fig:path_differential}
\end{figure}
Let $\nm$ be a single-pixel detector positioned at $\sensloc$, collecting radiation flowing in an \ac{IFOV} around direction $\sensdir$. The intensity $I_{\nm}$ measured by detector $\nm$ can be estimated by recursively \citep{veach1997robust} expanding \eq{\ref{eq:volume_rendering}}. This results in a path integral formulation of the RTE.

Let $\ppath=\regpar{\location_0,\location_1,...,\location_{\pathsize}}$ be a light path of an arbitrary size $\pathsize$. Let $\pathset$  be the set of all possible paths. The detector sensitivity  response to radiance carried by light path $\ppath$ is defined by a \ac{MCF} $\contfunc_{\nm}\regpar{\ppath}$ per pixel $\nm$. Then, the path integral formulation for pixel $\nm$ is:
\begin{equation}
\label{eq:path_integral_formulation}
I_{\nm} = \intop_{\pathset}\contfunc_{\nm}(\ppath)d\ppath .
\end{equation}
Here,  
\begin{equation}
    d\ppath=\prod_{\segind=0}^{\pathsize}{d\location_{\segind}}
\end{equation}
is the path differential, which consists of a product of differential volume and surface elements depending on whether $\location_{\segind}$ is in a medium or on a surface (illustrated in \fig{\ref{fig:path_differential}}). 

Now, we show how the \ac{MCF} can be evaluated. Let
\begin{equation}
\direction_{\segind} = \frac{\location_{\segind+1}-\location_{\segind}}{\norm{\location_{\segind+1}-\location_{\segind}}}
\end{equation}
be the normalized direction from $\location_{\segind-1}$ to $\location_{\segind}$.  Let $\pixmeasure_{\nm}\regpar{\location_{\pathsize-1},\location_{\pathsize}}$ be the response function of pixel $\nm$, which indicates the pixel sensitivity to the last path segment. For example, A utopean detector at $\location_{\nm}$ that collects radiation in direction $\sensdir$ \citep{gkioulekas2016evaluation} has:
\begin{equation}
\label{eq:wd_example}
\pixmeasure_{\nm}\regpar{\location_{\pathsize-1},\location_{\pathsize}}=\delta\regpar{\location_{\pathsize}-\sensloc}\delta\regpar{\direction_{\pathsize-1}-\sensdir}
\end{equation}
Notice that $\pixmeasure_{\nm}$ in \eq{\ref{eq:wd_example}} nulls the contribution of any path that does not reach $\sensloc$ at direction $\sensdir$.

\begin{figure*}[t]
  \centering
  \includegraphics[width=0.9 \linewidth]{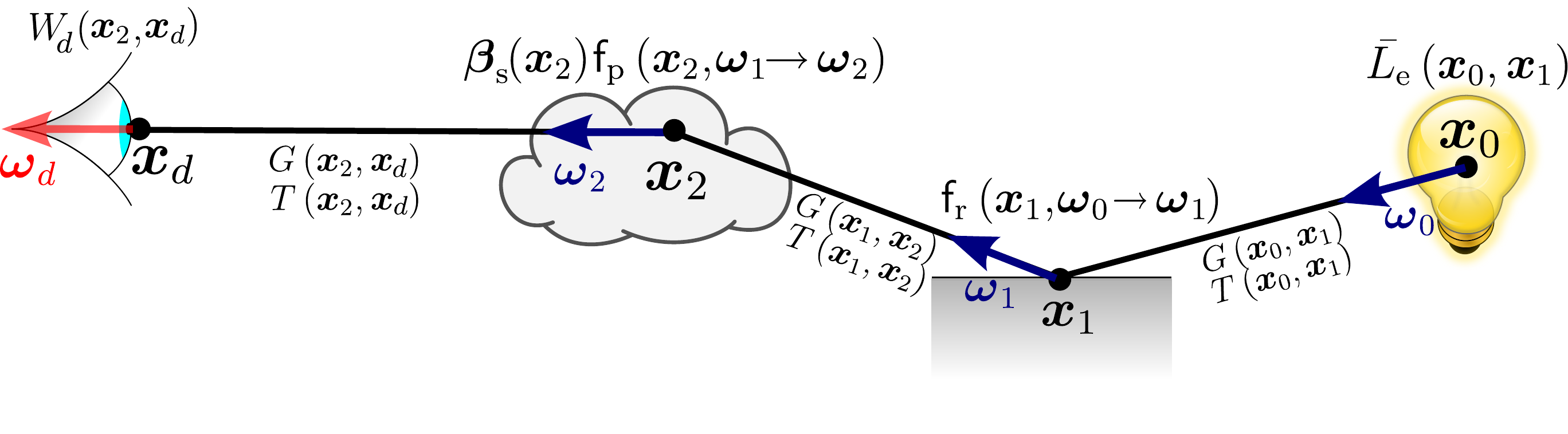}
    \caption{Illustration of $\contfunc_{\nm}\regpar{\ppath}$ for $\ppath=\curlpar{\location_0,\location_1,\location_2,\location_3}$. Here, $\direction_{\segind}$ is the direction from $\location_{\segind-1}$ towards $\location_{\segind}$. The first location $\location_0$ starts at the light source and contributes a factor of $\Bar{\radiancee}\regpar{\location_0,\location_1}=\radiancee\regpar{\location_0,\direction_0}$. The path propagation to $\location_1$ contributes a factor of $G\regpar{\location_0,\location_1}\trans{\location_0}{\location_1}.$ At $\location_1$, the path hits a surface and reflects towards $\direction_1$, which contributes a factor of $\brdf\regpar{\location_1,\direction_0\rightarrow\direction_1}$. The scattering event at $\location_2$ contributes a factor of $\exts\regpar{\location_2}\phasefunc\regpar{\location_2,\direction_1\rightarrow\direction_2}$. Finally, the path hits the detector at $\location_{3}$ and contributes a factor of $\pixmeasure_{\nm}\regpar{\location_2,\location_3}$. }
    \label{fig:path_integral}
\end{figure*}

Now, we provide definition for functions that define the \ac{MCF} $\contfunc_{\nm}\regpar{\ppath}$. Let
\begin{equation}
\label{eq:general_radiance}
\Bar{\radiancee}\regpar{\location_0,\location_1}= 
\begin{cases}
\radiancee\regpar{\location_0, \direction_0} & \text{if } \location \in \text{surface} \\
\exta\regpar{\location_0}\radiancee\regpar{\location_0, \direction_0} & \text{if } \location \in \text{medium}
\end{cases}
\end{equation}
be emitted radiance, considering both surface and volumetric events.
Let 
\begin{equation}
\label{eq:geomterm}
\geomterm\regpar{\location_{\segind},\location_{\segind+1}}=\dfrac{1}{\norm{\location_{\segind}-\location_{\segind+1}}^2}\cdot
\begin{cases}
\abs{\normal\regpar{\location_{\segind}}\cdot\direction_{\segind}}\abs{\normal\regpar{\location_{\segind+1}}\cdot\direction_{\segind}} & \text{if } \location\in \text{surface} \\
1 & \text{if } \location \in  \text{medium}.
\end{cases}  
\end{equation}
be a geometric term $\geomterm$ that converts a differential solid angle (\eqsnopar{\ref{eq:inscattered_radiance},\ref{eq:surface_rendering}}) to a differential area or volume (\eqnopar{\ref{eq:path_integral_formulation}}). Let
\begin{equation}
\label{eq:fs}
\anglefunc\regpar{\location_{\segind-1},\location_{\segind},\location_{\segind+1}}=
\begin{cases}
\brdf\regpar{\location_{\segind},\direction_{\segind-1}\rightarrow\direction_{\segind}} & \locationb \in \text{surface} \\
\exts\regpar{\location_{\segind}}\phasefunc\regpar{\location_{\segind},\direction_{\segind-1}\rightarrow\direction_{\segind}} &  \locationb \in  \text{medium}
\end{cases}
\end{equation}
be a directional function that considers both reflecting and scattering events.
Using \eqs{\ref{eq:general_radiance}-\ref{eq:fs}}, the \ac{MCF} can be defined as a multiplication of \ac{SCF}s, demonstrated in \fig\ref{fig:path_integral}:

Define the \ac{SCF} as follows,
$\forall \segind\in\curlpar{1,...,\pathsize-1}:$
\begin{equation}
\label{eq:gk}
\segfunc_{\segind}\regpar{\location_{\segind-1},\location_{\segind},\location_{\segind+1}}= \anglefunc\regpar{\location_{\segind-1},\location_{\segind},\location_{\segind+1}} \trans{\location_{\segind}}{\location_{\segind+1}}G\regpar{\location_{\segind},\location_{\segind+1}},
\end{equation}
and
\begin{equation}
\label{eq:g0K}
\segfunc_0\regpar{\location_{-1},\location_{0},\location_{1}}  =  \Bar{\radiancee}\regpar{\location_{0},\location_{1}}\trans{\location_0}{\location_{1}}\geomterm\regpar{\location_0,\location_{1}}.
\end{equation}
Then\footnote{Notice that $\location_{-1}$ and $\location_{\pathsize+1}$ do not exist. We use them for the ease of notation.},
\begin{equation}
\label{eq:cont_func}
\contfunc_{\nm}\regpar{\ppath} = \pixmeasure_{\nm}\regpar{\location_{\pathsize-1},\location_{\pathsize}} \prod_{\segind=0}^{\pathsize-1}{\segfunc_{\segind}\regpar{\location_{\segind-1},\location_{\segind},\location_{\segind+1}}}.
\end{equation}

In this notation, only light paths that start at the light source and end at the detector, meaning $\location_{\pathsize}=\sensloc$, may contribute to the intensity of pixel $\nm$. It is also possible to define $\contfunc_{\nm}\regpar{\ppath}$ such that a path starts at a pixel and ends at a light source as can be seen in \eqs{\ref{eq:g0K},\ref{eq:cont_func}}. While the two definitions are theoretically equivalent, they yield two different estimation algorithms.

At first sight, the path integral formulation of the light transport (\eqnopar{\ref{eq:cont_func}}) does not seem any easier to solve than \eq{\ref{eq:volume_rendering}}. This is because \eq{\ref{eq:cont_func}} requires every path that connects the light sources to a pixel. However, using the theory of  \ac{MC} integration \citep{veach1997robust}, we can estimate \eq{\ref{eq:cont_func}} by sampling light paths and summing their contribution. 


\subsection{Monte-Carlo Estimation of the Path Integral}
\label{sec:MC_est_path_integral}
Using the knowledge of \ac{MC} integration (Appendix \ref{sec:Monte-Carlo}), an estimation technique for \eq{\ref{eq:path_integral_formulation}} can be derived. The derivation here focuses only on a scattering medium. Recall \eqs{\ref{eq:general_radiance}-\ref{eq:fs}}. In this setting, \eq{\ref{eq:cont_func}} turns into:
\begin{multline}
\label{eq:cont_func_scattering}
\contfunc_{\nm}(\ppath) =  \pixmeasure_{\nm}\regpar{\location_{\pathsize-1},\location_{\pathsize}}\Bar{\radiancee}\regpar{\location_{0},\location_{1}}\trans{\location_0}{\location_{1}}\geomterm\regpar{\location_0,\location_{1}}\cdot \\ \cdot \prod_{\segind=1}^{\pathsize-1}{\exts\regpar{\location_{\segind}}\phasefunc\regpar{\location_{\segind},\dangle_{\segind-1,\segind}} \trans{\location_{\segind}}{\location_{\segind+1}}G\regpar{\location_{\segind},\location_{\segind+1}}}.
\end{multline}
Let $\ispdf_{\nm}$ be a \ac{PDF} that is associated with pixel $\nm$ and defined over the path space $\pathset$. Let $\curlpar{\pathi}_{{\pathindex}=1}^{\Np}$ be $\Np$ samples drawn from $\ispdf_{\nm}$. By employing \ac{MC} integration on \eq{\ref{eq:path_integral_formulation}}, we get the following estimator:
\begin{equation}
\label{eq:forward_estimator}
I\approx\frac{1}{\Np}\sum_{\pathindex=1}^{\Np}{\frac{\contfunc_{\nm}\regpar{\pathi}}{\ispdf_{\nm}\regpar{\pathi}}}.
\end{equation}
In the rest of this paper, we refer to $\ispdf_{\nm}$ as the {\em rendering technique}. The realization of \eq{\ref{eq:forward_estimator}} depends on having:
\begin{itemize}
  \item An explicit method that evaluates $\ispdf_{\nm}\regpar{\ppath} \quad \forall \ppath\in\pathset$.
  \item An explicit process that generates paths according to $\ispdf_{\nm}$.
\end{itemize}
Moreover, we want $\ispdf_{\nm}$ to sample $\contfunc_{\nm}$ in a way that reduces the estimation error. This can be obtained by choosing $\ispdf_{\nm}$ that is similar to $f$ (Appendix \ref{sec:IS}). 

In this section, we describe $\mu$ which results in the traditional {\em path tracing} algorithm for participating media. \citep{veach1997robust}. {\em Path tracing} draws consecutive segments and not a path as a whole through an explicit use of $\ispdf$.
There are two approaches in {\em path tracing}. One approach is forward path tracing, which sets the first vertex of the path at the light source. The other approach is backward path tracing, which sets the first vertex at the detector location. We focus on the first approach.  

Assume a single isotropic point light source at $\lightloc$.
The first vertex $\location_0$ of the path is set at the light source. A direction $\direction_0$ is uniformly sampled:
\begin{equation}
    \location_0 = \lightloc, \quad \direction_0 \sim \mathcal{U}\squarepar{\sphere}, \quad   \ray_0 =\regpar{\location_0, \direction_0},
\end{equation}
and $\ray_0$ is an initial ray.
 The \ac{PDF} of $\ray_0$ is:
\begin{equation}
    \ispdf_{\text{ray}}\regpar{\ray_0} = \frac{1}{4\pi}.
\end{equation}
To each segment $\curlpar{\location_{\segind-1},\location_{\segind},\location_{\segind+1}}$ of a path, we associate a ray as follows:
\begin{equation}
    \ray_{\segind} \triangleq \regpar{\location_{\segind},\direction_{\segind}},
\end{equation}
Let $\dangle_{\segind-1,\segind}=\arccos\regpar{\direction_{\segind-1}\cdot\direction_{\segind}}$ be the scattering angle of the $\segind$'th segment.
Now, we need to define a sampling process that samples a ray $\ray_{\segind}$, given the previous ray $\ray_{\segind-1}$ in a path. Sampling a new ray consists of two steps. The first step samples the distance $\distance_{\segind}$ to the next point $\location_{\segind}=\location_{\segind-1}+\distance_{\segind}\direction_{\segind-1}$. The second step samples the scattering direction $\direction_{\segind}$.
Let $\ispdf_l\regpar{l_k|\location_{\segind-1},\direction_{\segind-1}}$ and $\ispdf_{\direction}\regpar{\direction_{\segind}|\direction_{\segind-1}, \location_{\segind}}$ be the distance \ac{PDF} and the direction \ac{PDF} respectively. Then, the ray \ac{PDF} is defined as follows:
\begin{equation}
\label{eq:semgent_sampling}
\ispdf_{\text{ray}}\regpar{\ray_{\segind}|\ray_{\segind-1}} = \ispdf_{\distance}\regpar{\distance_{\segind}|\location_{\segind-1},\direction_{\segind-1}}
\cdot \ispdf_{\direction}\regpar{\direction_{\segind}|\direction_{\segind-1},\location_{\segind}}\cdot\geomterm\regpar{\location_{\segind-1},\location_{\segind}} .
\end{equation}
Here the geometric term $\geomterm$ (see Table \ref{table:cont_funct}) converts solid angles to volume measures as requested by \eq{\ref{eq:path_integral_formulation}}.
The total path \ac{PDF} is:
\begin{equation}
\label{eq:path_sampling}
\ispdf_{\nm}\regpar{\ppath} =\ispdf_{\text{ray}}\regpar{\ray_0}\prod_{\segind=1}^{\pathsize}{\ispdf_{\text{ray}}\regpar{\ray_{\segind}|\ray_{\segind-1}}}.
\end{equation}
{\em Path tracing} samples a new ray by applying {\em importance sampling}. {\em Importance sampling} \citep{kajiya1986rendering, veach1997robust} is a technique to reduce the variance of an \ac{MC} estimator (Appendix \ref{sec:IS}). This is done by sampling the paths from a \ac{PDF} that is proportional to the integrand of \eq{\ref{eq:path_integral_formulation}}. Recall from \eq{\ref{eq:gk}-\ref{eq:cont_func}} that for a scattering medium, which does not contain surfaces, the \ac{SCF} is:
\begin{equation}
\label{eq:gk_scatter}
\segfunc_{\segind}\regpar{\location_{\segind-1}, \location_{\segind}, \location_{\segind+1}}=\exts\regpar{\location_{\segind}}\phasefunc\regpar{\location_{\segind},\dangle_{\segind-1,\segind}}
\trans{\location_{\segind-1}}{\location_{\segind}} \geomterm\regpar{\location_{\segind-1},\location_{\segind}} ,
\end{equation}
as presented in \eq{\ref{eq:gk}}. Notice that \eq{\ref{eq:gk_scatter}} uses $\dangle_{\segind-1,\segind}$ as the argument for the phase function. The rest of this paper follows this notation. To apply importance sampling on \eq{\ref{eq:path_integral_formulation}}, {\em path tracing} samples a new ray as follows (illustrated in \fig\ref{fig:segment_sampling}):
\begin{figure}[t]
  \centering
  \includegraphics[width=0.6\linewidth]{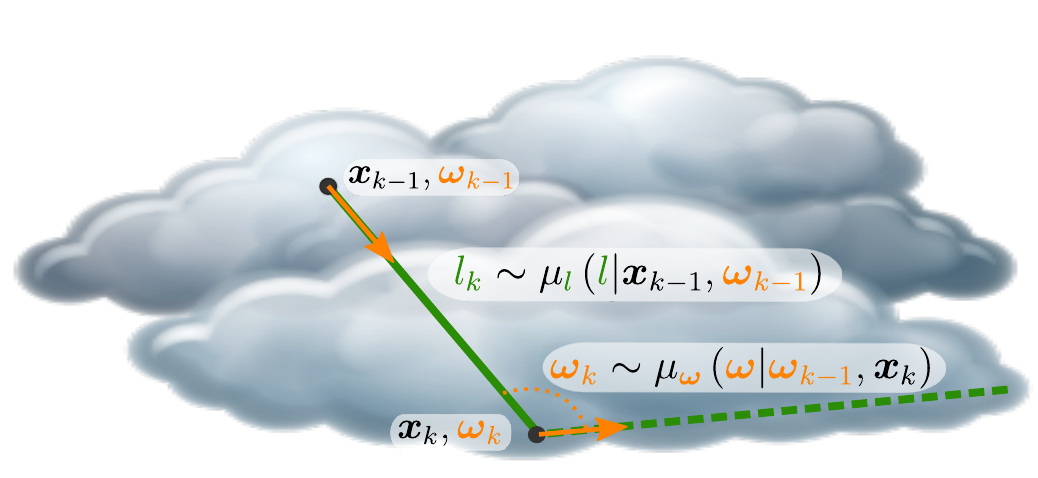} 
    \caption{Ray sampling. Given $\location_{\segind-1}, \direction_{\segind-1}$, we need to sample $\location_{\segind}$, $\direction_{\segind}$. First, we sample the distance $\distance_{\segind}$ to the next point (green segment). Then, we set $\location_{\segind}=\location_{\segind-1}+~\distance_{\segind}\direction_{\segind-1}$ and sample the scattering direction $\direction_{\segind}$ (orange arrow). These steps are repeated until a terminal condition is reached.}
    \label{fig:segment_sampling}
\end{figure}    

\subsubsection{Distance Sampling}
\label{sec:distance_sampling}
\begin{enumerate}
    \item Sample $\tau$ from an exponential \ac{PDF}, i.e., $\tau\sim\exp\regpar{1}$
    \item Find $\distance_{\segind}$ that satisfies:    
    \begin{equation}
    \trans{\location_{\segind-1}}{\location_{\segind-1}+\distance_{\segind}\direction_{\segind-1}}=\exp(-\tau).
    \end{equation}
    \item Set $\location_{\segind}=\location_{\segind-1}+\distance_{\segind}\direction_{\segind-1}$.
\end{enumerate}
\subsubsection{Direction Sampling}
\label{sec:direction_sampling}
\begin{enumerate}
    \item Sample the scattering particle index $\ptype \in \curlpar{1,...,\Npar}$ with probability:
    \begin{equation}
    P\regpar{\ptype}=\frac{\extt^{\regpar{\ptype}}\regpar{\location_{\segind}}}{\sum_{j'=1}^{\Npar}{\extt^{\regpar{j'}}\regpar{\location_{\segind}}}}.
    \end{equation}
    \item Sample the scattering direction from the phase function of the $\ptype$'th particle type\footnote{Scattering direction refers to both elevation and azimuth.}: $$\direction_{\segind}\sim\phasefunc^{\regpar{\ptype}}\regpar{\cos\dangle_{\segind-1,\segind}}.$$
\end{enumerate}
The resulting path \ac{PDF} (Appendix \ref{sec:ptpdf}) is :
\begin{equation}
\label{eq:final_mu}
\ispdf_{\nm}\regpar{\ppath} =\frac{1}{4\pi}\trans{\location_{0}}{\location_{1}}\geomterm\regpar{\location_{0},\location_{1}}\prod_{\segind=1}^{\pathsize-1}{\extt\regpar{\location_{\segind}}\phasefunc\regpar{\location_{\segind},\dangle_{\segind-1,\segind}} \trans{\location_{\segind}}{\location_{\segind+1}}\geomterm\regpar{\location_{\segind},\location_{\segind+1}}}.
\end{equation}
Path sampling terminates when the path hits the detector or leaves the medium.

The similarity of the resulting \ac{PDF} $\ispdf_{\nm}$ (\eqnopar{\ref{eq:final_mu}}) to the \ac{MCF} $\contfunc_{\nm}$ (\eqnopar{\ref{eq:cont_func_scattering}}), significantly reduces the estimation error. With an analytical form of $\ispdf_{\nm}$, we can write the {\em path tracing} estimator. Let $\pathsize_{\pathindex}$ be the number of scattering events of the $\pathindex$'th path. Let $\lastloc=\locationi_{\pathsize_{\pathindex}-1}$ be the last location of the last scattering event of the $\pathindex$'th path. Then (Appendix \ref{sec:ptpdf}):
\begin{equation}
\label{eq:explicit_foward}
I_{\nm}\approx \frac{4\pi}{\Np}\sum_{\pathindex=1}^{\Np}{
\radiancee\regpar{\location_0,\directioni_0}\pixmeasure_{\nm}\regpar{\lastloc,\sensloc} \cdot \prod_{\segind=1}^{\pathsize_{\pathindex}}{
\sinscat\regpar{\locationi_{\segind}}}}
.
\end{equation}
Notice that the geometric term $\geomterm$ from \eq{\ref{eq:gk},\ref{eq:g0K},\ref{eq:final_mu}} is canceled out in the division between $\contfunc_{\nm}$ and $\ispdf_{\nm}$.
\subsubsection{Local estimation}
\ac{MC} can efficiently estimate $I_{\nm}$ using local estimation \citep{marshak20053d}. Local estimation expresses the probability that a photon scatters from point $\location_{\segind}$ towards the light source without interacting again.

\begin{figure}[t]
  \centering
  \includegraphics[width=0.7\linewidth]{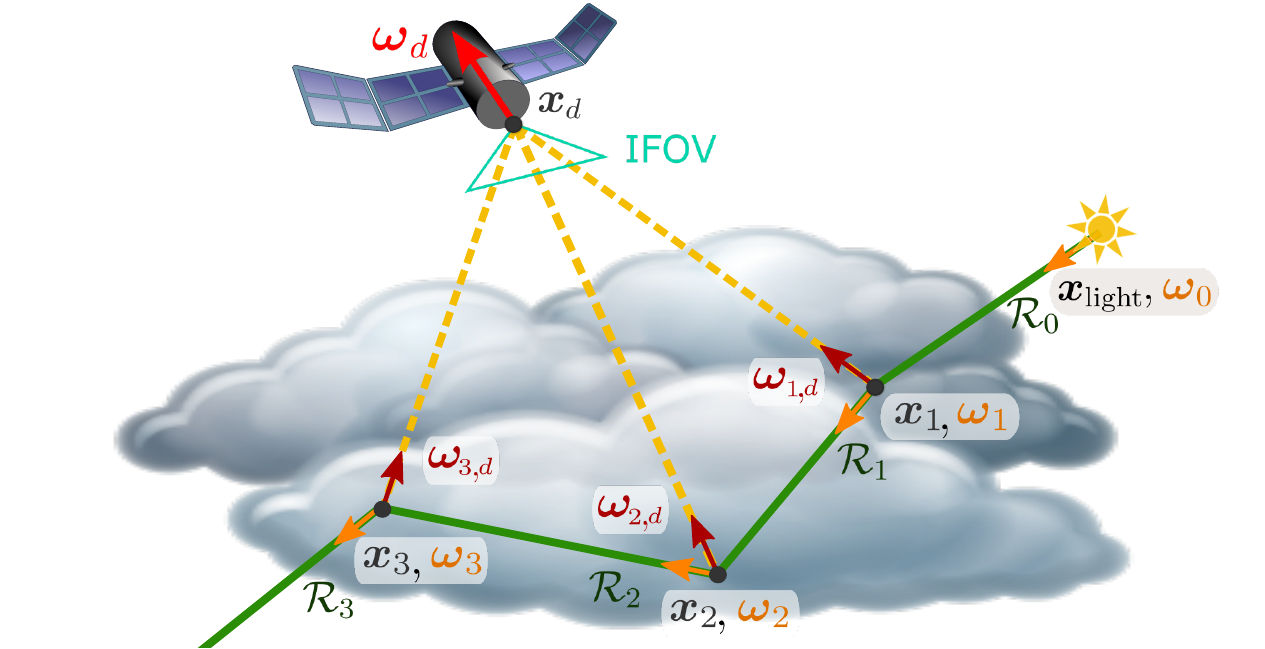}
    \caption{Illustration of the {\em path tracing} forward model. A light source is at $\lightloc$ and a detector at $\sensloc$. The detector collects radiance in an \ac{IFOV} around $\sensdir$. First, an initial direction $\direction_0$ is sampled uniformly from the unit sphere (orange arrow). Then, a next scattering point $\location_1$ is sampled on ray from $\lightloc$ towards $\direction_0$ (green segment). At scattering, an imaginary ray (red arrow) is sent to the detector for calculating the {\em local estimation} (yellow dashed line). Light scatters to a new direction $\direction_1$ (orange arrow). These steps are repeated until the light path leaves the medium. Notice that a simulation of multiple detectors differ only in the {\em local estimation}. Hence, this approach is efficient for  simulating multiple measurements in parallel.}
    \label{fig:fowrad_simulation}
\end{figure}

Only light paths which end at the detector contribute to the estimation. When the domain is large in three dimensions, reaching a small detector location is a rare event. \ac{MC} solves this problem by sending an imaginary ray to the detector at each scattering event. Then, the path contribution is multiplied by the probability that the light path scatters directly to the detector without scattering again. Let
\begin{equation}
    \sensbdir= \frac{\location_{\segind}-\sensloc}{\norm{\location_{\segind}-\sensloc}}
\end{equation}
be the direction from $\location_{\segind}$ towards the detector. Let $\dangle_{\segind,d}=\arccos\regpar{\direction_{\segind-1}\cdot\sensbdir}$ be the scattering angle from $\location_{\segind}$ towards the detector. Then, the local estimation probability density is given by:
\begin{equation}
    \nexteventpdf\regpar{\location_{\segind},\dangle_{\segind,d}}=\phasefunc\regpar{\location_{\segind},\cos\dangle_{\segind,d}} \trans{\location_{\segind}}{\sensloc} \geomterm\regpar{\location_{\segind},\sensloc}.
\end{equation}
The {\em local estimation} technique changes \eq{\ref{eq:explicit_foward}} as follows:
\begin{equation}
   I_{\nm} \approx \frac{4\pi}{\Np}\sum_{\pathindex=1}^{\Np}{\radiancee\regpar{\location_0,\directioni_0}  Q_{\nm}\regpar{\pathi}},
\end{equation}
where
\begin{equation}
\label{eq:Qd}
    Q_{\nm}\regpar{\ppath} = \sum_{\segind=1}^{\pathsize-1}{\prod_{\segind'=1}^{\segind}{\sinscat\regpar{\locationi_{\segind'}} \nexteventpdf\regpar{\locationi_{\segind'},\danglei_{\segind,d}}}  \pixmeasure_{\nm}\regpar{\sensloc, \locationi_{\segind'}}}.
\end{equation}
Each summand of \eq{\ref{eq:Qd}} corresponds to a different local estimation event. Each event is attenuated by the single scattering albedo, the probability of reaching the detector without scattering again and the detector response function.
An illustration of {\em path tracing} together with local estimation is given by \fig\ref{fig:fowrad_simulation}.
\section{THE INVERSE PROBLEM}
\label{sec:inverse}
A forward model estimates pixel values that would have been measured in a real scene. Inverse rendering fits given pixel values to unknown corresponding scene parameters. This paper considers two tasks. One task is retrieving a 3D model of a medium extinction coefficient. The other task is reflectometry, which seeks to estimates a \ac{BRDF} of a surface. This section starts with a general formulation of inverse rendering as an optimization problem. Then, this section follows by a method to estimate the gradient of the optimization loss. This section is built upon \citep{loeub2020monotonicity}.
\subsection{Problem Definition}
\label{sec:inverse_problem_definition}
Let $\sceneparam = \squarepar{\scenescalar_1, \scenescalar_2,...,\scenescalar_{\Nu}}$ be a vector of $\Nu$ scene unknowns of interest and $\sceneparam^{*}$ be their true values. For example, in scattering tomography, $\sceneparam$ may be the extinction coefficient $\extt$ in all voxel of a grid. In reflectometry, $\sceneparam$ may be a parameterization of a $\brdf$ for example. Let $\unknownindex$ be an index of a variable in $\sceneparam$. Let $\gtmeasurements=\squarepar{\gtscalar_1,\gtscalar_2, ..., \gtscalar_{\Nm}}$ be a vector of $\Nm$ ground-truth intensity measurements. Let $\forwardmodel\regpar{\sceneparam}=\squarepar{\forwardmodel_1\regpar{\sceneparam},...,\forwardmodel_{\Nm}\regpar{\sceneparam}}$ be a vector of forward model measurements. Each model measurement $\nm$ corresponds to an element in $\gtmeasurements$, but uses  \eq{\ref{eq:path_integral_formulation}} based on $\ispdf_{\nm}$. 

Let $\contfunc_{\nm}\regpar{\ppath|\sceneparam}$ be the \ac{MCF} that corresponds to the $\nm$'th measurement and the scene parameters $\sceneparam$. Then, the $\nm$'th element of the forward model vector is given by the path integral:
\begin{equation}
\label{eq:forward_model}
\forwardmodel_{\nm}\regpar{\sceneparam}\triangleq\intop_{\pathset}{\contfunc_{\nm}\regpar{\ppath|\sceneparam}d\ppath}.
\end{equation}
Define an appearance loss as
\begin{equation}
\label{eq:appearance_loss}
\lossfunc\regpar{\sceneparam|\gtmeasurements}=\frac{1}{2}\norm{\forwardmodel\regpar{\sceneparam} - \gtmeasurements} ^2
= \frac{1}{2}\sum_{\nm=1}^{\Nm}{\abs{\forwardmodel_{\nm}\regpar{{\sceneparam}}-\gtmeasurements_{\nm}}^2}.
\end{equation}
We define the following optimization problem for finding $\sceneparam^*$:
\begin{equation}
\label{eq:opt_problem}
 \Hat{\sceneparam}=\arg\min_{\sceneparam}{\lossfunc\regpar{\sceneparam|\gtmeasurements}}\approx\sceneparam^* .
\end{equation}
To solve \eq{\ref{eq:opt_problem}} using a gradient-based optimization algorithm, we need to compute the gradient:
\begin{equation}
\label{eq:loss_derivative}
\frac{\partial \lossfunc\regpar{\sceneparam|\gtmeasurements}}{\partial \scenescalar_{\unknownindex}} = \sum_{\nm=1}^{\Nm}{\squarepar{\forwardmodel_{\nm}\regpar{{\sceneparam}}-\gtmeasurements_{\nm}}}\frac{\partial \forwardmodel_{\nm}\regpar{\sceneparam}}{\partial \scenescalar_{\unknownindex}}.
\end{equation}
The term $\squarepar{\forwardmodel_{\nm}\regpar{{\sceneparam}}-\gtmeasurements_{\nm}}$ is the estimation error, which can be obtained by running the forward model. The term $\frac{\partial \forwardmodel_{\nm}\regpar{\sceneparam}}{\partial \scenescalar_{\unknownindex}}$ is the derivative of the model with respect to an unknown. Notice that for computing the gradient, we need to evaluate \eq{\ref{eq:loss_derivative}}  $\forall \unknownindex\in\curlpar{1,2,...,\Nu}$. Let
\begin{equation}
\label{eq:total_derivative_factor}
\scorefunc_{\unknownindex, \nm}\regpar{\ppath|\sceneparam} \triangleq \frac{\deriv{\scenescalar_{\unknownindex}} \pixmeasure_{\nm}\regpar{\location_{\pathsize-1},\location_{\pathsize}}}{ \pixmeasure_{\nm}\regpar{\location_{\pathsize-1},\location_{\pathsize}}}+ \sum_{\segind=0}^{\pathsize-1}{\frac{\deriv{\scenescalar_{\unknownindex}}\segfunc_{\segind}\regpar{\location_{\segind-1}, \location_{\segind}, \location_{\segind+1}}}{\segfunc_{\segind}\regpar{\location_{\segind-1}, \location_{\segind}, \location_{\segind+1}}}}.
\end{equation}
be the \ac{PSF}\footnote{In optics, the acronym \ac{PSF} often refers to a point spread function. } \citep{shem2020towards}. 

Then (Appendix \ref{sec:partial_deriv_app}),
\begin{equation}
\label{eq:int_derivative_full}
\frac{\partial \forwardmodel_{\nm}\regpar{\sceneparam}}{\partial \scenescalar_{\unknownindex}} =  \intop_{\pathset}{\contfunc_{\nm}\regpar{\ppath|\sceneparam}\cdot \scorefunc_{\unknownindex,\nm}\regpar{\ppath|\sceneparam} d\ppath}.
\end{equation}
is the path integral formulation for computing the partial derivative of the forward model with respect to the parameter $\scenescalar_{\unknownindex}$. For \eqs{\ref{eq:total_derivative_factor},\ref{eq:int_derivative_full}} to be well defined, we need to know the partial derivative of the \ac{SCF}s with respect to $\scenescalar_{\unknownindex}$. 
\subsection{Gradient Estimation}
\label{sec:grad_est}
Let $\ispdf_{\nm}\regpar{\ppath|\sceneparam}$ be the rendering technique that corresponds to the $\nm$'th measurement and the scene parameters $\sceneparam$. Similarly to \eq{\ref{eq:path_integral_formulation}}, \eq{\ref{eq:int_derivative_full}} is formulated as an integral over the path space $\pathset$. We can use this analogy to estimate  $\frac{\partial \forwardmodel_{\nm}\regpar{\sceneparam}}{\partial \scenescalar_{\unknownindex}}$ using the same rendering technique $\ispdf_{\nm}\regpar{\ppath|\sceneparam}$. Let us sample $\Np$ light paths according to: $$\curlpar{\pathi}_{\pathindex=1}^{\Np}\simiid\ispdf_{\nm}\regpar{\ppath|\sceneparam}.$$ 
Then, similarly to \eq{\ref{eq:forward_estimator}},
\begin{equation}
\label{eq:gradient_simualtion}
\derivx{\hat{\forwardmodel_{\nm}}\regpar{\sceneparam}}{\scenescalar_{\unknownindex}}\triangleq\frac{1}{\Np}\sum_{\pathindex=1}^{\Np}{\frac{\contfunc_{\nm}\regpar{\pathi|\sceneparam}\cdot\scorefunc_{\unknownindex,\nm}\regpar{\pathi|\sceneparam}}{\ispdf_{\nm}\regpar{\pathi|\sceneparam}}}
\end{equation}
is a \ac{MC} estimation to \eq{\ref{eq:int_derivative_full}}. Recall estimation of the forward model (\eqnopar{\ref{eq:forward_estimator}}): 
\begin{equation}
\label{eq:forward_simualtion}
\hat{\forwardmodel_{\nm}}\regpar{\sceneparam}\triangleq\frac{1}{\Np}\sum_{\pathindex=1}^{\Np}{\frac{\contfunc_{\nm}\regpar{\pathi|\sceneparam}}{\ispdf_{\nm}\regpar{\pathi|\sceneparam}}},
\end{equation}
\eqs{\ref{eq:gradient_simualtion}, \ref{eq:forward_simualtion}} then lead to estimation of \eq{\ref{eq:loss_derivative}} using:
\begin{equation}  
\label{eq:loss_gradient}
\derivx{\hat{\lossfunc}\regpar{\sceneparam|\gtmeasurements}}{\scenescalar_{\unknownindex}}=\sum_{\nm=1}^{\Nm}{\squarepar{\hat{\forwardmodel_{\nm}}\regpar{{\sceneparam}}-\gtmeasurements_{\nm}}}\frac{\partial \hat{\forwardmodel_{\nm}}\regpar{\sceneparam}}{\partial \scenescalar_{\unknownindex}} .
\end{equation}

\subsection{Inverse Rendering Algorithm}
Using \eq{\ref{eq:loss_gradient}}, a general inverse rendering algorithm based on \ac{MC} path tracing can be defined (\alg\ref{alg:inverse_rendering}). The algorithm receives $\Nm$ ground-truth measurements $\gtmeasurements=\squarepar{\gtmeasurements_1,\gtmeasurements_2,...,\gtmeasurements_{\Nm}}$ and returns an estimation  $\Hat{\sceneparam}$. 

\begin{algorithm}
\DontPrintSemicolon
\SetAlgoLined
    \SetKwInOut{Input}{Input}
    \SetKwInOut{Output}{Output}
\Input{$\gtmeasurements$ - $\Nm$ ground-truth measurements\\ $\stepsize$ - step size}
\Output{$\Hat{\sceneparam}$ - Estimated scene parameters}
 initialize $\sceneparam^{0}$\;
 \For{$\iter=1,2,...,\mathcal{T}$}{
 $\forall \nm\in\curlpar{1,...,\Nm}:$
  Sample $\curlpar{\pathi}_{\pathindex=1}^{\Np} \simiid \ispdf_{\nm}\regpar{\ppath|\sceneparamt}$\; 
    
  $\boldsymbol{I}^{\regpar{\iter}}\leftarrow\hat{\forwardmodel}\regpar{\sceneparamt}$ \tcp*[l]{run forward simulation}
  \For{$\unknownindex = 1,2,...,\Nu$}{
     $\gradest\leftarrow\derivx{\hat{\forwardmodel}\regpar{\sceneparamt}}{\scenescalar_{\unknownindex}}$  \tcp*[l]{run gradient simulation}
     $\scenescalar_{\unknownindex}^{\regpar{\iter+1}} \leftarrow \scenescalar^{\regpar{\iter}}_{\unknownindex} - \stepsize\cdot\sum_{\nm=1}^{\Nm}{\regpar{\boldsymbol{I}^{\regpar{\iter}}_{\nm}-\gtmeasurements_{\nm}}\gradest_{\nm}}$ \tcp*[l]{gradient update}} 
    
}
return $\sceneparam^{\regpar{\maxiter}}$

 \caption{\ac{MC} Inverse Rendering}
 \label{alg:inverse_rendering}
\end{algorithm}
To conclude, solving an inverse rendering problem requires:
\begin{enumerate}
    \item A rendering technique for estimating the forward model - \eq{\ref{eq:forward_simualtion}}
    \item An explicit form of the \ac{PSF} - \eq{\ref{eq:total_derivative_factor}}
\end{enumerate}
\subsection{Scattering Tomography}
\begin{figure}[b]
  \centering
  \includegraphics[width=0.6\linewidth]{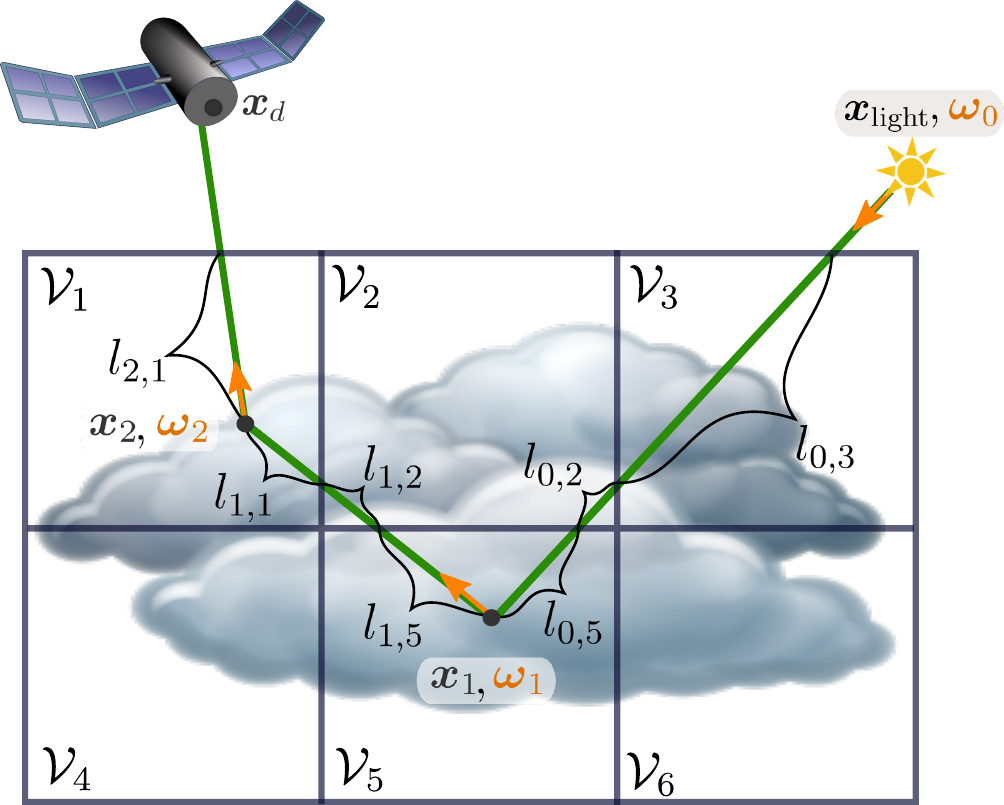}
    \caption{Demonstration of $\voxelset_{\voxel}$ and $\distance_{\segind,\voxel}$ on a light path.}
    \label{fig:inverse_notations}
\end{figure}
In scattering tomography, we are interested to estimate properties of the particles in the scene. In this paper, we assume two  particle types, air molecules and cloud droplets. Let $\extcloud(\location)$,$\sinscatcloud$ and $\extair(\location)$,$\sinscatair$ be the total extinction coefficients and single scattering albedo of cloud droplets and air molecules, respectively. In this setting, $\extair(\location), \sinscatcloud, \sinscatair$ are known. We seek to estimate $\extcloud(\location)$. First, we follow Section \ref{sec:inverse} to define scattering tomography as an inverse problem. 

Let $\sceneparam=\squarepar{\extcloud_1, \extcloud_2,...,\extcloud_{\Nu}}$ be a discrete grid representation of the cloud total extinction coefficient where $\Nu$ is the number of voxels (volume elements in the grid). We assume that the cloud extinction coefficient is a constant in voxel $\unknownindex$, denoted $\extcloud_\unknownindex$. For computing the gradient of the gradient of the forward model (\eqnopar{\ref{eq:int_derivative_full}}), we need to evaluate $\scorefunc_{\unknownindex,\nm}(\ppath|\sceneparam)$ for each  unknown index $\unknownindex \in \squarepar{1,2,...,\Nu}$.

The line segment between $\location_{\segind}$ and $\location_{\segind+1}$ is denoted by $\linesegment{\location_{\segind}}{\location_{\segind+1}}$, and contains all
points satisfying $\curlpar{\location=\lambda \location_{\segind}+(1-\lambda )\location_{\segind+1}):\lambda \in\squarepar{0,1}}$. The domain of voxel $\voxel$ is denoted by $\voxelset_{\voxel}$. The line segment $\linesegment{\location_{\segind}}{\location_{\segind+1}}$ traverses
several voxels. Denote by $\distance_{\segind,\voxel}$ the length of the intersection of $\voxelset_{\voxel}$ with $\linesegment{\location_{\segind}}{\location_{\segind+1}}$ demonstrated in \fig\ref{fig:inverse_notations}. Then (Appendix \ref{sec:psf_st}),
\begin{equation}
\scorefunc_{\unknownindex, \nm}\regpar{\ppath|\sceneparam} = -\sum_{\segind=0}^{\pathsize}{\distance_{\segind,\voxel}}+
\sum_{\segind=0}^{\pathsize-1}{
\begin{cases}
\regpar{\extcloud_{\voxel}+\dfrac{\sinscatair\phasefunc^{\text{a}}\regpar{\dangle_{\segind-1,\segind}}}{\sinscatcloud\phasefunc^{\text{c}}\regpar{\dangle_{\segind-1,\segind}}}\extair_{\voxel}} ^ {-1} & \text{if $\location_{\segind}\in\voxelset_{\voxel}$} \\
0 & \text{else}
\end{cases}}.
\end{equation}

\section{PATH RECYCLING}
\begin{figure}[t]
  \centering
  \includegraphics[width=0.8\linewidth]{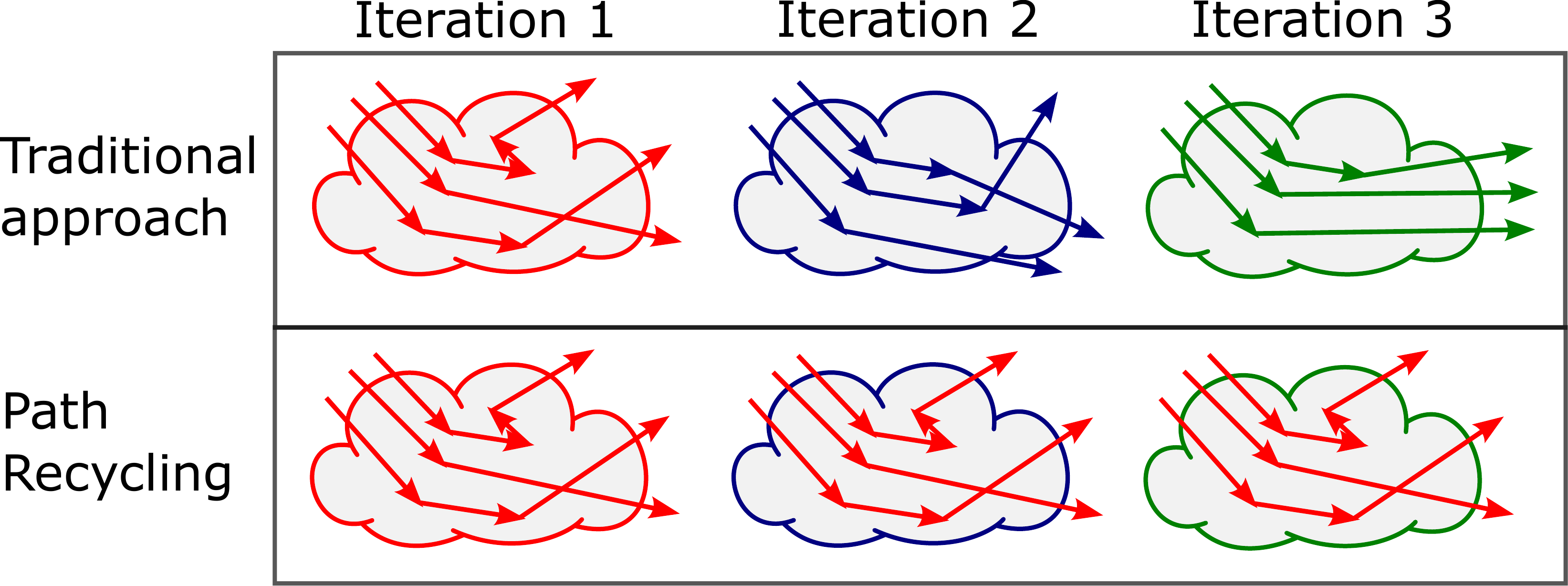} 
    \caption{Path recycling illustration.  Each cloud is of a different iteration. The cloud contour color represents its changes throughout the iterations. The path colors correspond to the cloud that is used for sampling the paths. [Top] The traditional approach samples a new set of paths in each iteration. [Bottom] Our path recycling method uses the same set of paths for several iterations.}
    \label{fig:path_recycling_figure}
\end{figure}    

\label{sec:recycling}
This paper presents a novel technique for \ac{MC} inverse rendering that efficiently utilizes the information encoded by the light paths. Previous methods sample a new set of light paths in each iteration of the optimization, resulting in a long optimization run-time. We suggest storing the light paths in a designated data structure and recycling them in the next $\Nr$ iterations. Since the scene parameters $\sceneparamt$ change between iterations, light paths from previous iterations are not trivially valid for the current iteration. In this section, we introduce a correction factor that corrects this discrepancy. \fig\ref{fig:path_recycling_figure} illustrates the basic concept of path recycling. Recycling makes image rendering {\em less} efficient than traditional methods, increasing noise variance per light path, thus run-time. However, recall that we are interested here, fundamentally, in {\em inverse problems}. For inverse problems, recycling is more efficient than traditional path sampling.
\subsection{Problem Definition}
Let $\sceneparamt$ be the scene parameters at the current iteration $\iter$ and $\sceneparef=\sceneparam^{\iter'}$ be the scene parameters from a previous iteration $\iter'<\iter$.
Again, we are interested in estimating $\forwardmodel_{\nm}\regpar{\sceneparamt}$ and $\derivx{\forwardmodel_{\nm}\regpar{\sceneparamt}}{m}$ as shown in Section \ref{sec:grad_est}. However, in this framework, we only have light paths that correspond to a different parameter vector:
$$\curlpar{\ppath}^{\Np}_{\pathindex=1}\simiid \ispdf_{\nm}\regpar{\ppath|\sceneparef}.$$
\subsection{Our Method}
\begin{figure}[t]
\centering
\tikzstyle{decision} = [diamond, draw, fill=blue!20, 
    text width=4.5em, text badly centered, node distance=3cm, inner sep=0pt]
\tikzstyle{block} = [rectangle, draw, fill=blue!20, 
    text width=7.5em, text centered, rounded corners, minimum height=2em]
\tikzstyle{line} = [draw, -latex']
\tikzstyle{cloud} = [draw, ellipse,fill=red!20, node distance=3cm,
    minimum height=2em]
    
\begin{tikzpicture}[node distance = 2cm, auto]
    \node [block] (init) {Initialize scene parameters};
    \node [block, below of=init] (sample_first) {Sample paths};
    \node [block, below of=sample_first] (forward) {Run differentiable forward model};
    \node [block, below of=forward] (update) {Gradient update};
    \node [block, left of=update, node distance=3cm] (resample) {Resample paths};
    \node [decision, below of=update] (decide) {Should resample?};
    \node [block, below of=decide, node distance=3cm] (stop) {Return estimated scene parameters};
    \path [line] (init) -- (sample_first);
    \path [line] (sample_first) -- (forward);
    \path [line] (forward) -- (update);
    \path [line] (update) -- (decide);
    \path [line] (decide.west) -| node[near start] {yes} (resample);
    \draw [line] (decide.east) -- + (2cm,0)  node[near start]{no} |-  (forward);
    \path [line] (decide) -- node {at convergence}(stop);
    \path [line] (resample) |- (forward);

\end{tikzpicture}
\caption{Flowchart of the path recycling algorithm.}
\label{fig:path_recycling_diagram}
\end{figure}
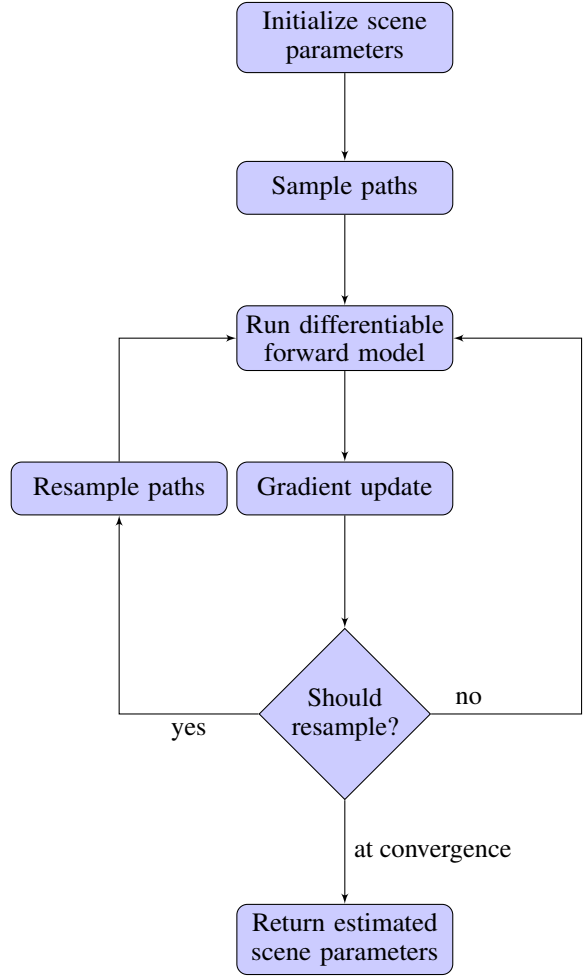

The basic concepts of \ac{MC} integration allow us to correct the discrepancies. Recall that \ac{MC} provides freedom to choose a rendering technique. We suggest using $\ispdf_{\nm}\regpar{\cdot|\sceneparef}$ as our technique for rendering the parameters $\sceneparamt$. In other words, we can define
\begin{equation}
\label{eq:forward_simualtion_recycling}
\hat{\forwardmodel_{\nm}}\regpar{\sceneparamt|\sceneparef}\triangleq\frac{1}{\Np}\sum_{\pathindex=1}^{\Np}{\frac{\contfunc_{\nm}\regpar{\pathi|\sceneparamt}}{\ispdf_{\nm}\regpar{\pathi|\sceneparef}}}
\end{equation}
as the estimator for $\forwardmodel_{\nm}\regpar{\sceneparamt}$ given $\Np$ light paths drawn from $\ispdf_{\nm}\regpar{\cdot|\sceneparef}$.
Denote a path recycling correction factor:
\begin{equation}
\label{eq:correction factor}
\correctfact_{\nm}\regpar{\ppath,\sceneparamt|\sceneparef}\triangleq\frac{\ispdf_{\nm}\regpar{\ppath|\sceneparamt}}{\ispdf_{\nm}\regpar{\ppath|\sceneparef}}.
\end{equation}
Then:
\begin{equation}
\label{eq:forward_simualtion_recycling_factor}
\hat{\forwardmodel_{\nm}}\regpar{\sceneparamt|\sceneparef}=\frac{1}{\Np}\sum_{\pathindex=1}^{\Np}{{\frac{\contfunc_{\nm}\regpar{\pathi|\sceneparamt}}{\ispdf_{\nm}\regpar{\pathi|\sceneparamt}}}}\correctfact_{\nm}\regpar{\pathi,\sceneparamt|\sceneparef} .
\end{equation}
The correction for each path is a ratio between \ac{PDF}s. The partial derivative of the forward model can be equivalently estimated by:
\begin{equation}
\label{eq:gradient_simualtion_recycling}
\derivx{\hat{\forwardmodel_{\nm}}\regpar{\sceneparamt|\sceneparef}}{\scenescalar_{\unknownindex}}\triangleq\frac{1}{\Np}\sum_{\pathindex=1}^{\Np}{\frac{\contfunc_{\nm}\regpar{\pathi|\sceneparamt}\scorefunc_{\unknownindex,\nm}\regpar{\pathi|\sceneparamt}}{\ispdf_{\nm}\regpar{\pathi|\sceneparamt}}} \correctfact_{\nm}\regpar{\pathi,\sceneparamt|\sceneparef}.
\end{equation}
The estimation error of \eqs{\ref{eq:forward_simualtion_recycling_factor},\ref{eq:gradient_simualtion_recycling}} increases together with the discrepancies between $\sceneparamt$ and $\sceneparef$. However, in a gradient-based optimization, each iteration only slightly changes the unknowns in the descent direction. This property maintains the usefulness of paths from previous iterations. We employ path recycling (\eqsnopar{\ref{eq:forward_simualtion_recycling_factor},\ref{eq:gradient_simualtion_recycling}})  in \alg\ref{alg:inverse_recycling}, to speed-up \alg\ref{alg:inverse_rendering}. A flowchart diagram of path recycling is provided by \fig\ref{fig:path_recycling_diagram}.
\subsection{Path Sorting.}
\begin{algorithm}[t]
\DontPrintSemicolon
\SetAlgoLined
    \SetKwInOut{Input}{Input}
    \SetKwInOut{Output}{Output}
\Input{$\gtmeasurements$ - A set of ground-truth measurements\\ $\stepsize$ - step size}
\Output{$\Hat{\sceneparam}$ - Estimated scene parameters}
 initialize $\sceneparam^{0}$\;
 \For{$\iter=1,2,...,\maxiter$}{
 \If(\tcp*[h]{every $\Nr$ iteration}){$\regpar{\iter\mod\Nr==0}$}{
  $\forall \nm\in{1,...,\Nm} :$ Sample $\curlpar{\pathi}_{\pathindex=1}^{\Np} \underset{\text{\ac{I.I.D}}}{\sim} \ispdf_{\nm}\regpar{\ppath|\sceneparamt}$ \;
  $\forall \nm\in{1,...,\Nm} :$ Sort $\curlpar{\pathi}_{\pathindex=1}^{\Np}$ \;
  $\sceneparef \leftarrow \sceneparamt$ \tcp*[l]{set reference}
    }
  $\boldsymbol{I}^{\regpar{\iter}}\leftarrow\hat{\forwardmodel}\regpar{\sceneparamt|\sceneparef}$ \tcp*[l]{forward simulation}
   \For{$\unknownindex = 1,2,...,\Nu$}{
     $\gradest\leftarrow\derivx{\hat{\forwardmodel}\regpar{\sceneparamt|\sceneparef}}{\scenescalar_{\unknownindex}}$  \tcp*[l]{gradient simulation}
    $\scenescalar_{\unknownindex}^{\regpar{\iter+1}} \leftarrow \scenescalar^{\regpar{\iter}}_{\unknownindex} - \stepsize\cdot\sum_{\nm=1}^{\Nm}{\regpar{\boldsymbol{I}_{\nm}-\gtmeasurements_{\nm}}\gradest_{\nm}}$ \tcp*[l]{gradient update}} 
    
}
return $\sceneparam^{\regpar{\maxiter}}$
 \caption{Path Recycling Inverse Rendering}
 \label{alg:inverse_recycling}
\end{algorithm}

Recall that $\pathsize$ is a path size. Light path sampling can be parallelized since the light paths are independent. However, a \ac{GPU} architecture cannot trivially benefit from this parallelism. A \ac{GPU} executes tasks in {\em groups}. Each group has 32 {\em threads} such that each thread represents a different path sampling instance. To fully utilize the \ac{GPU} capabilities, all the threads within the same group have to execute the same commands. Unfortunately, different light paths have different $\pathsize$. In traditional methods,$\pathsize$ is known only after sampling ends. However, in path recycling, the light paths have already been sampled and can be sorted according to $\pathsize$. This lets us follow a similar approach as \citep{afra2016local}, without using the wavefront approach. This enables much better utilization of the \ac{GPU} architecture. As a result, path sorting speeds up the runtime by up to 3 times. We note that our sorting procedure is simple. Further research for more complex sorting procedures may speed up the runtime even more. 


\section{SIMULATIONS}
\label{sec:simulations}
This section demonstrates how path recycling significantly speeds up recovery. Moreover, to show that the approach is not limited to scattering tomography, we give a basic reflectometry example. We implemented a 3D rendering engine to solve both scattering tomography and reflectometry. To exploit the parallelism of the problems, we implemented it on a \ac{GPU}. We use Numba \citep{lam2015numba}, which is a \ac{JIT} compilation library for python. Numba efficiently compiles Python codes to \ac{GPU} kernels. We implemented a modified version of \ac{ADAM} \citep{kingma2014adam} with the following parameters: first moment decay $\momentdecay_1=0.9$, second moment decay $\momentdecay_2=0.999$, and step size $\stepsize=10^7$.
To assess the reconstruction quality, we use two measures following \citep{loeub2020monotonicity}:
\begin{equation}
\reldist = \frac{\norm{\sceneparam^*-\Hat{\sceneparam}}_1}{{\norm{\sceneparam^*}_1}},\quad
\relbias = \frac{\norm{\sceneparam^*}_1-||\Hat{\sceneparam}||_1}{{\norm{\sceneparam^*}_1}} .
\end{equation}
\subsection{Scattering Tomography}


\subsubsection{Scene Settings.} 
\begin{figure}[t]
  \centering
   \includegraphics[width=0.6\linewidth]{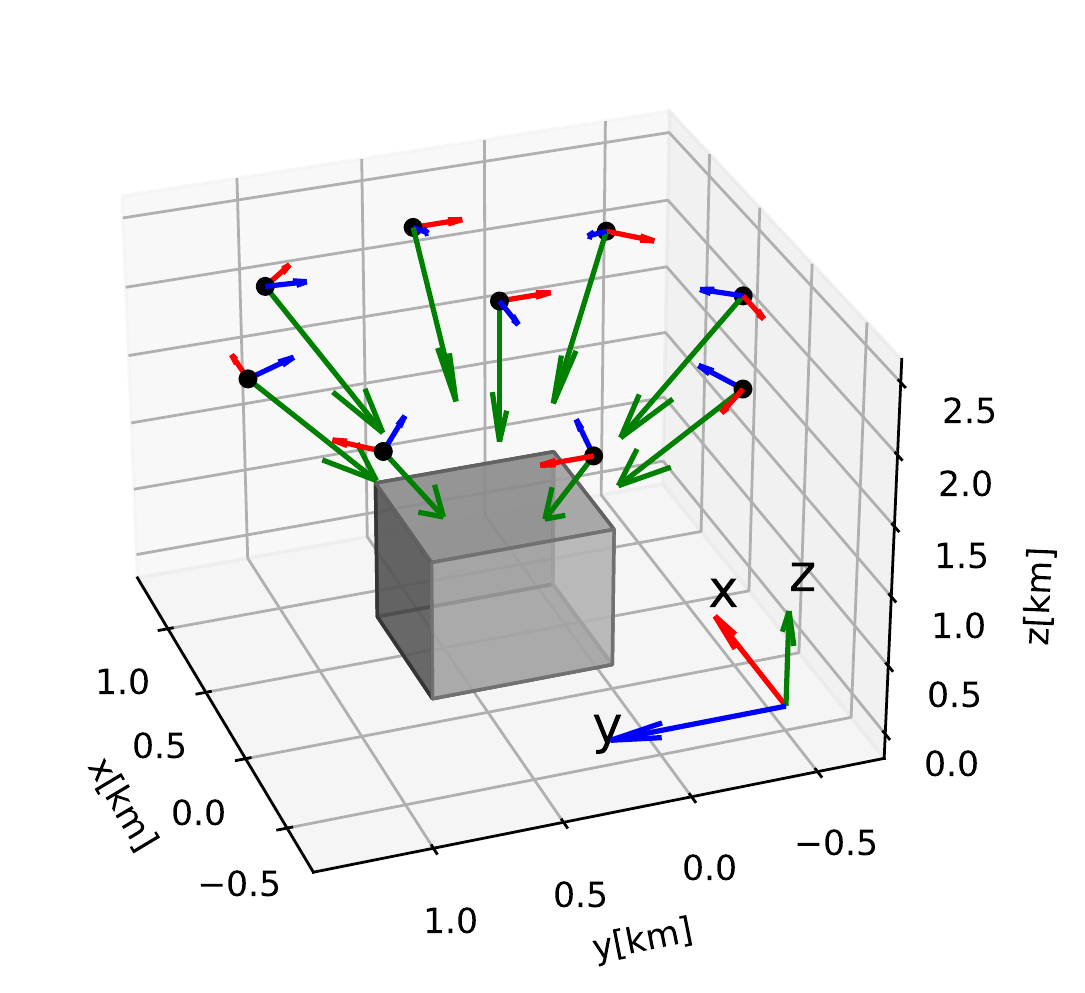} 
    \caption{Camera formation. This figure illustrates the geometry in a single cloud scene. One camera is at the zenith and the other eight are on a horizontal ring, $29^{\circ}$ from the zenith.}
    \label{fig:camera_setting}
\end{figure}

This paper considers three scattering scenes. 
All scenes have a constant air extinction coefficient $\extair=0.04\squarepar{\frac{1}{km}}$ and single scattering albedos $\sinscatcloud=0.99, \sinscatair=0.912$. The scenes are illuminated by the sun at the zenith, using parallel rays. The air phase function is modeled by Rayleigh scattering \citep{frisvad2011importance} and the cloud phase function is modeled by a Henyey-Greenstein function \citep{binzoni2006use}. The scenes are captured by nine perspective cameras. All cameras have the same intrinsic setting. One camera is at the zenith and eight cameras reside on a horizontal ring. This formation is illustrated in \fig\ref{fig:camera_setting}. Additional scene settings are provided in Table \ref{table:scene_settings}.
\begin{table}[b]
\centering
\begin{tabular}{c | c | c | c | c | c | c} 
 \hline
 Scene & Bounding Box & \thead{Max \\Optical \\Depth} & Voxels & Pixels & \thead{Viewing \\zenith \\angle} & Radius\\ [0.7ex] 
 \hline\hline 

 Plume & $2.56\times1\times 2.56 \squarepar{m^{3}}$  & $10 \squarepar{\frac{1}{m}}$& $128\times 50 \times 128$  & $200\times200$ & $90^{\circ}$ & $3.84m$ \\
 \hline

 Single Cloud & $0.64\times0.72\times 1 \squarepar{km^{3}}$  &$127 \squarepar{\frac{1}{km}}$ & $32\times 36 \times 25$  & $76\times76$  & $29^{\circ}$  & $2km$\\
  \hline
  
 Cloud Field & $1.58\times1.68\times 1.08 \squarepar{km^{3}}$  &$127 \squarepar{\frac{1}{km}}$&  $79\times 84 \times 27$   & $86\times86$ & $33^{\circ}$ &   $2.16km$  \\ \hline
\end{tabular}
\caption{Scene properties. The viewing zenith angle describes the ring of the 8 cameras. The radius is the distance of all cameras from the center of the bounding box.}
\label{table:scene_settings}
\end{table}

The first scene is of a plume generated using the Mitsuba fluid solver \citep{Mitsuba}. The second scene contains a single cloud, which was generated by a \ac{LES} \citep{chung2014large}. \ac{LES} is a tool used by atmospheric scientists as a standard for computationally generating cloud fields from first principles. Rendering the cloud requires more computer power than the plume since the light scatters more frequently in the cloud, as seen in the optical depths in Table.~\ref{table:scene_settings}. The third scene contains a cloud field, generated by \ac{LES}. 

The reconstruction process is initialized by space carving \citep{kutulakos2000theory}. The estimated photo-hull is then set with the average extinction coefficient of the cloud. The optimization is performed in a coarse to fine manner \citep{loeub2020monotonicity}. We start the optimization with a down-sampled version of the ground truth images having $30\times30$ pixels and use $N=10^6$ light paths. The image resolution and the number of light paths increase each time the loss function reaches saturation. At the highest resolution, $N=5\times10^7$. All reconstructions are performed on a Tesla V100 DGXS 32GB.


\begin{figure}[t]
  \centering
   \includegraphics[width=0.9\linewidth]{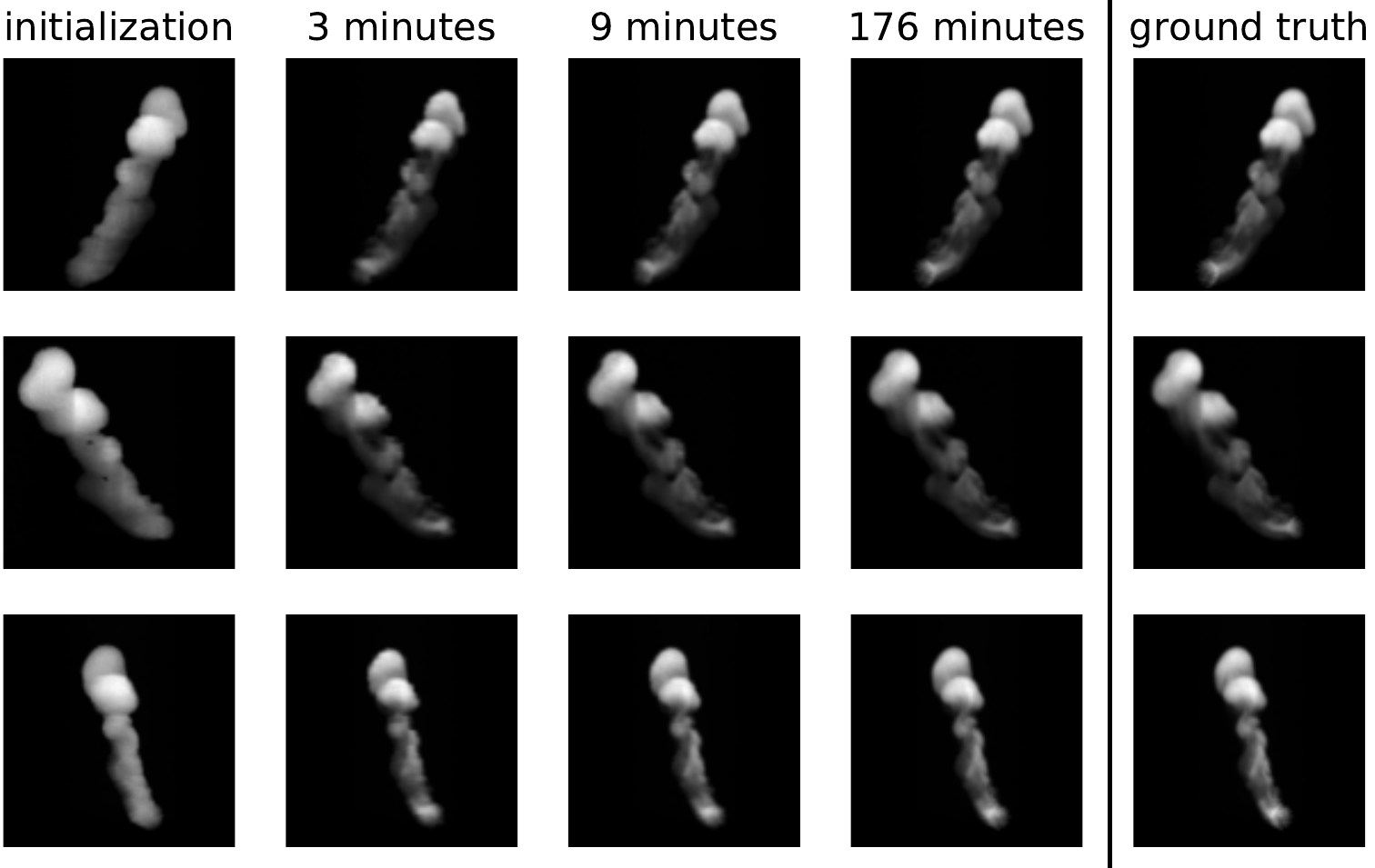} 
    \caption{Plume scene convergence. Each row shows the convergence of one viewpoint. The bottom row shows a viewpoint that was not used during the optimization. Each column shows results after a different stage of the optimization. The right-most column shows the corresponding ground truth images. }
    \label{fig:smoke_convergence_grid}
\end{figure}




\begin{figure}[b!]
  \centering
  \begin{minipage}[b]{0.4\textwidth}
  \includegraphics[width=\textwidth]{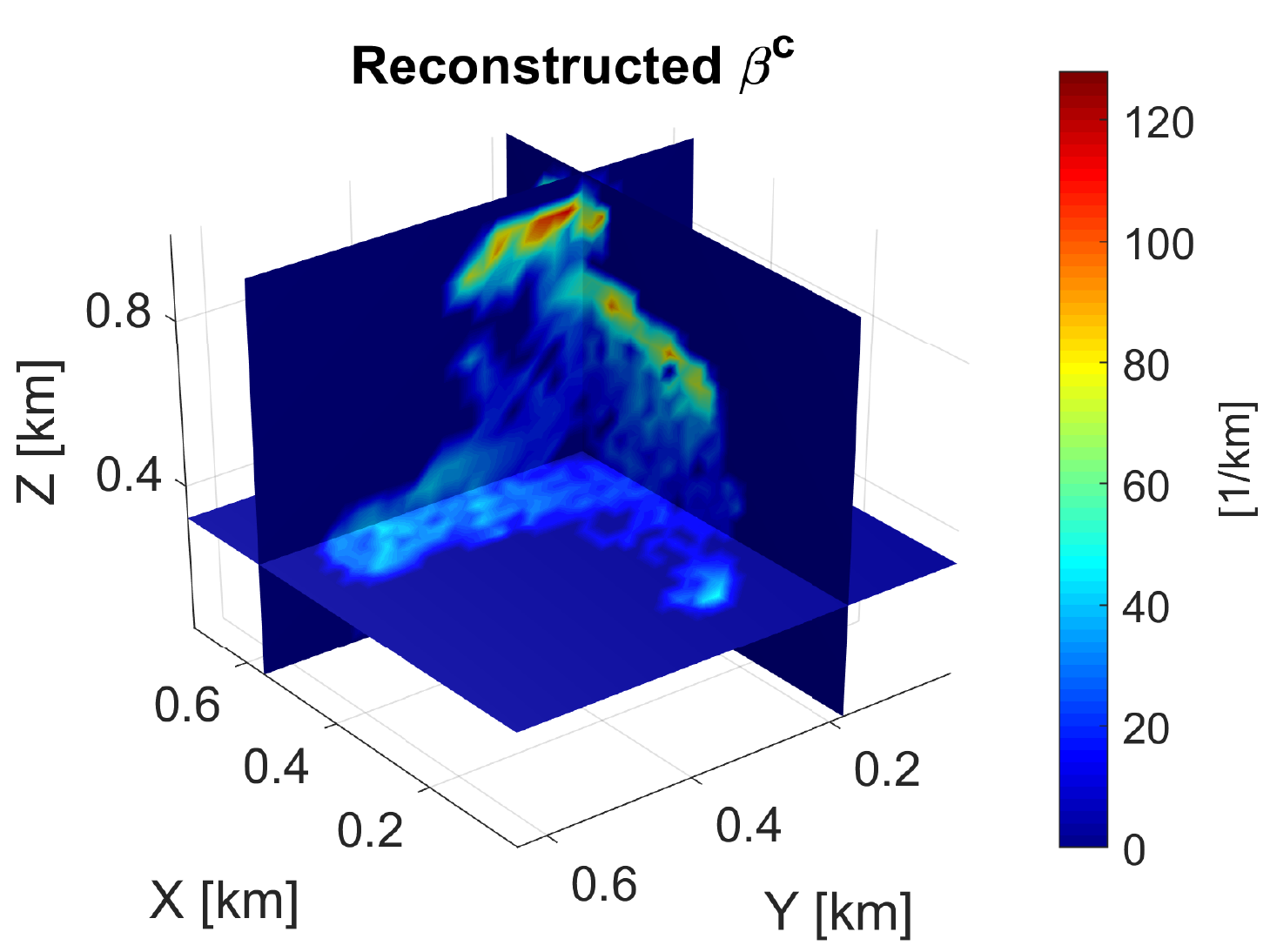}
  \end{minipage}
    \begin{minipage}[b]{0.3\textwidth}
  \includegraphics[trim=0 0 100 0,clip, width=\textwidth]{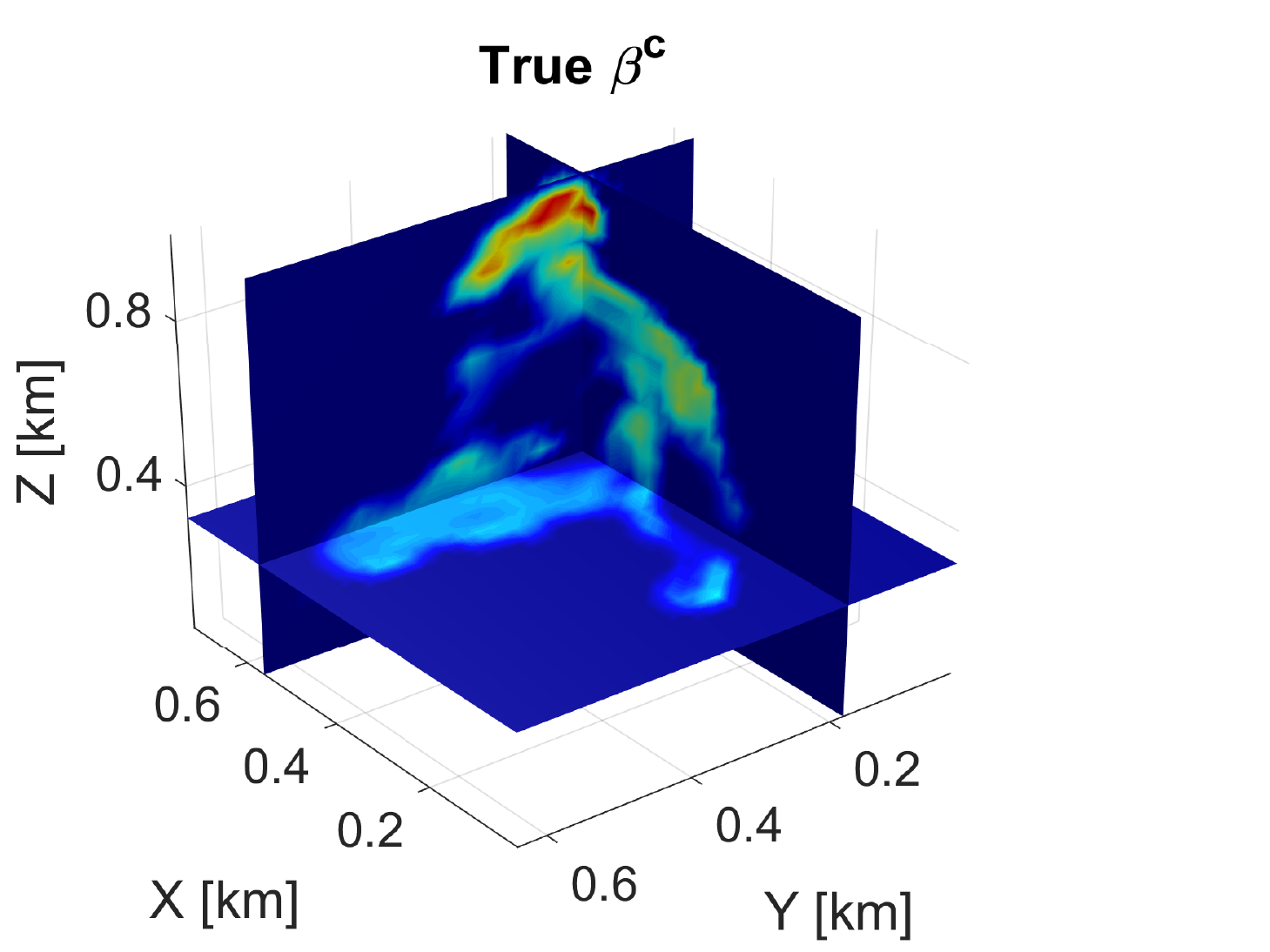}
  \end{minipage}
      \begin{minipage}[b]{0.28\textwidth}
  \includegraphics[width=\textwidth]{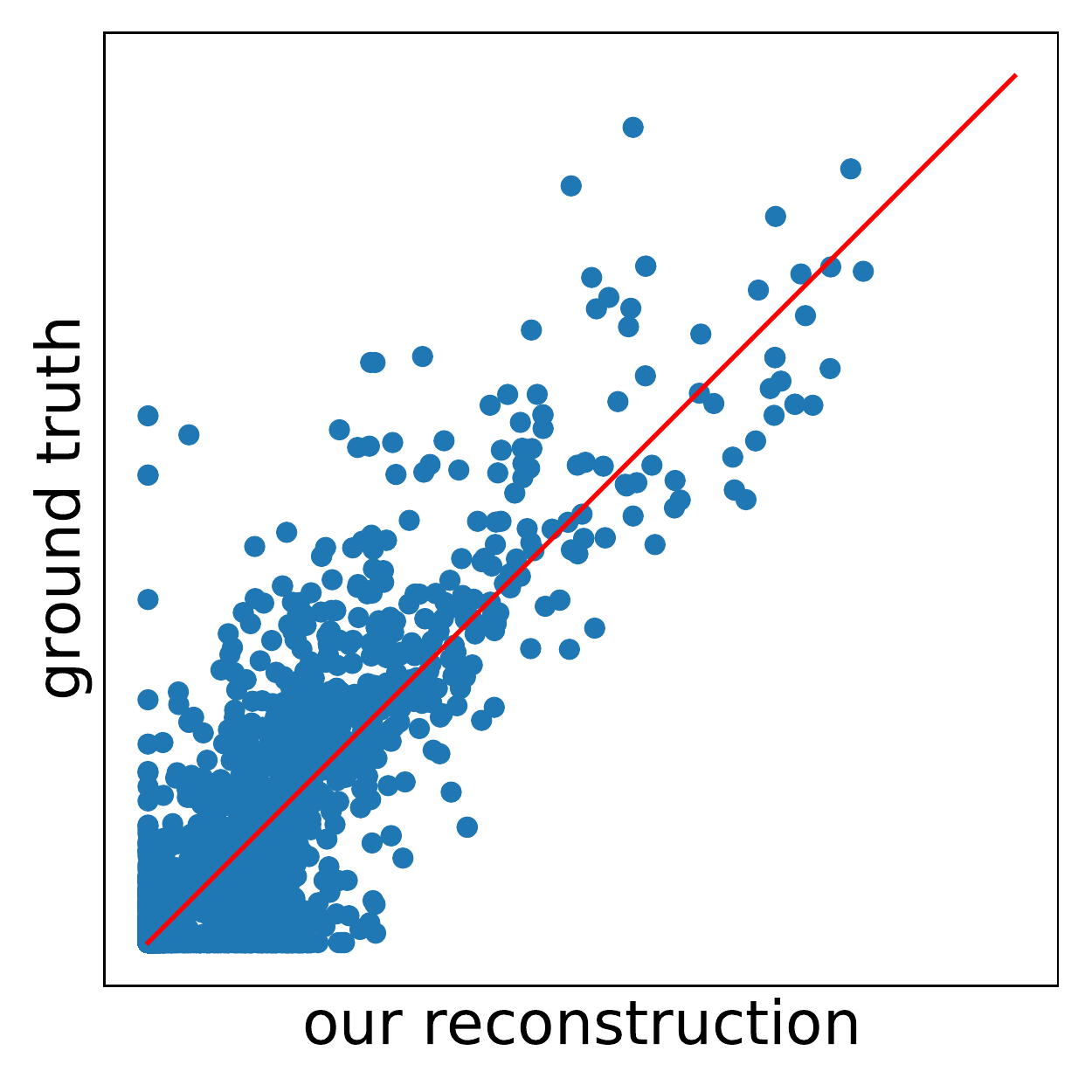}
  \end{minipage}
    \caption{3D slices of the single cloud scene. [Left] The reconstructed medium. [Middle] The true medium. [Right] Scatter plot of 30\% of the data points, randomly selected.}
    \label{fig:cloud_plane}
\end{figure}

\begin{figure}
  \centering
  \begin{minipage}[b]{0.4\textwidth}
  \includegraphics[width=\textwidth]{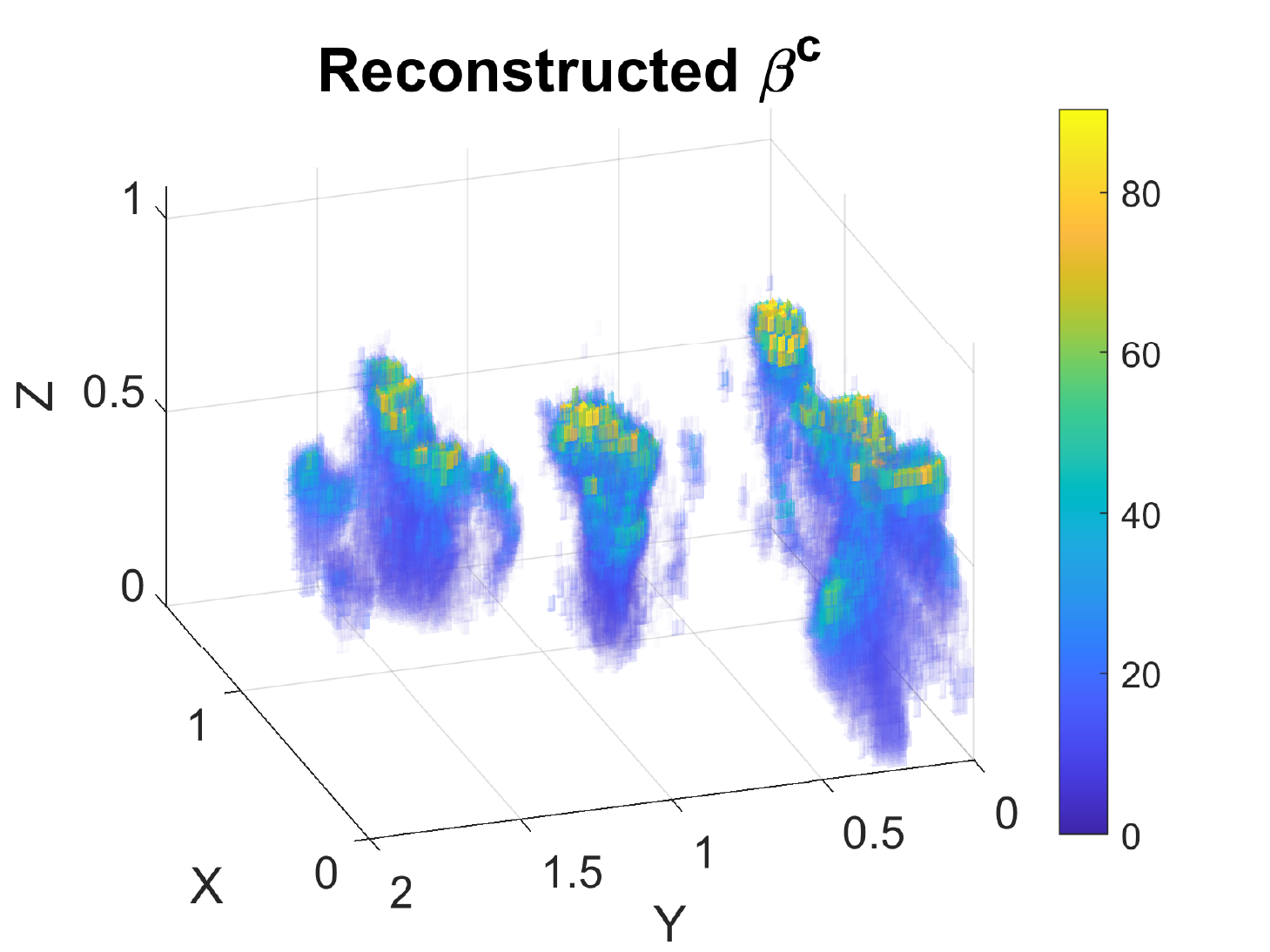}
  \end{minipage}
    \begin{minipage}[b]{0.335\textwidth}
  \includegraphics[trim=0 0 71 0,clip, width=\textwidth]{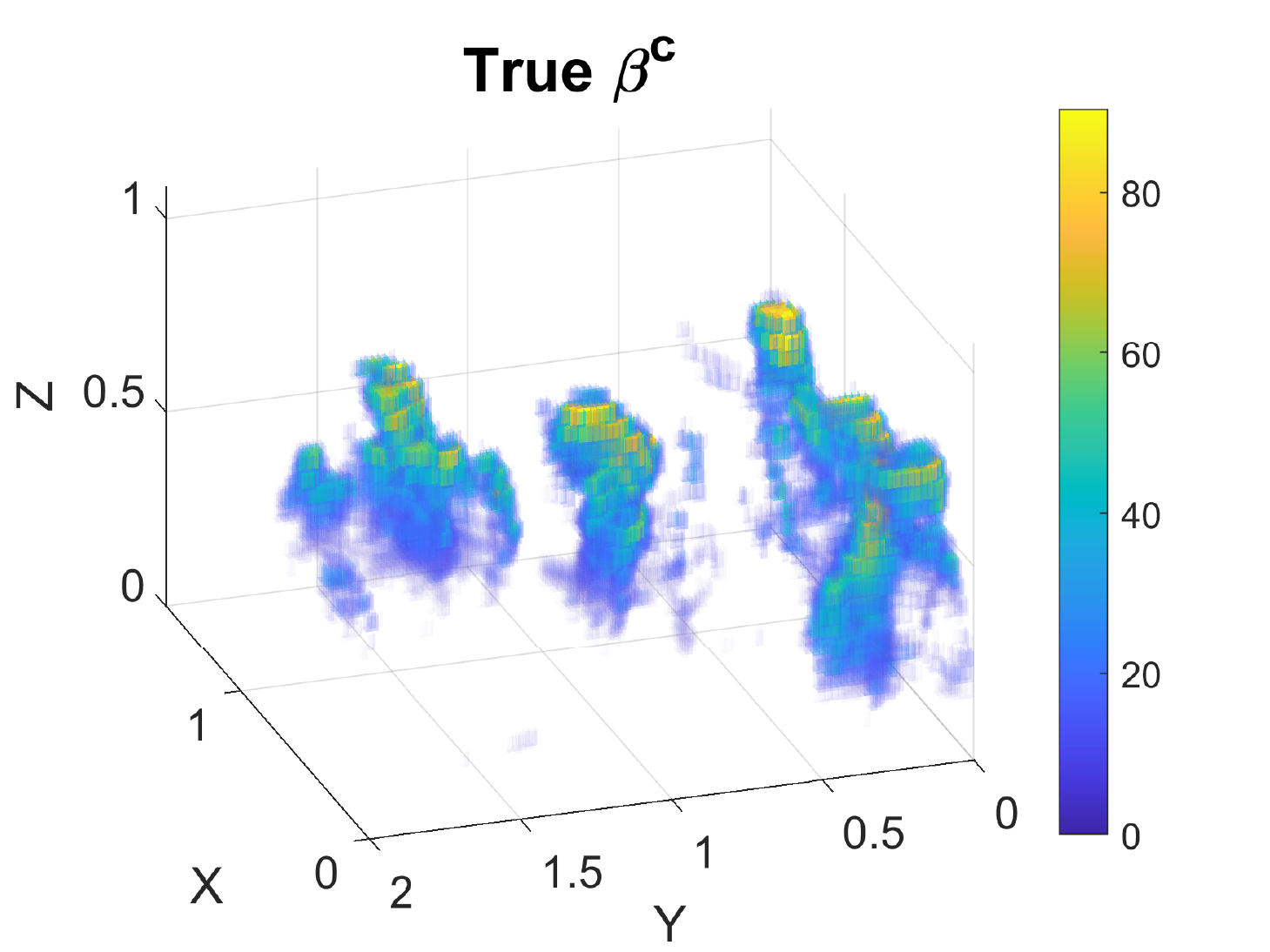}
  \end{minipage}
      \begin{minipage}[b]{0.245\textwidth}
  \includegraphics[width=\textwidth]{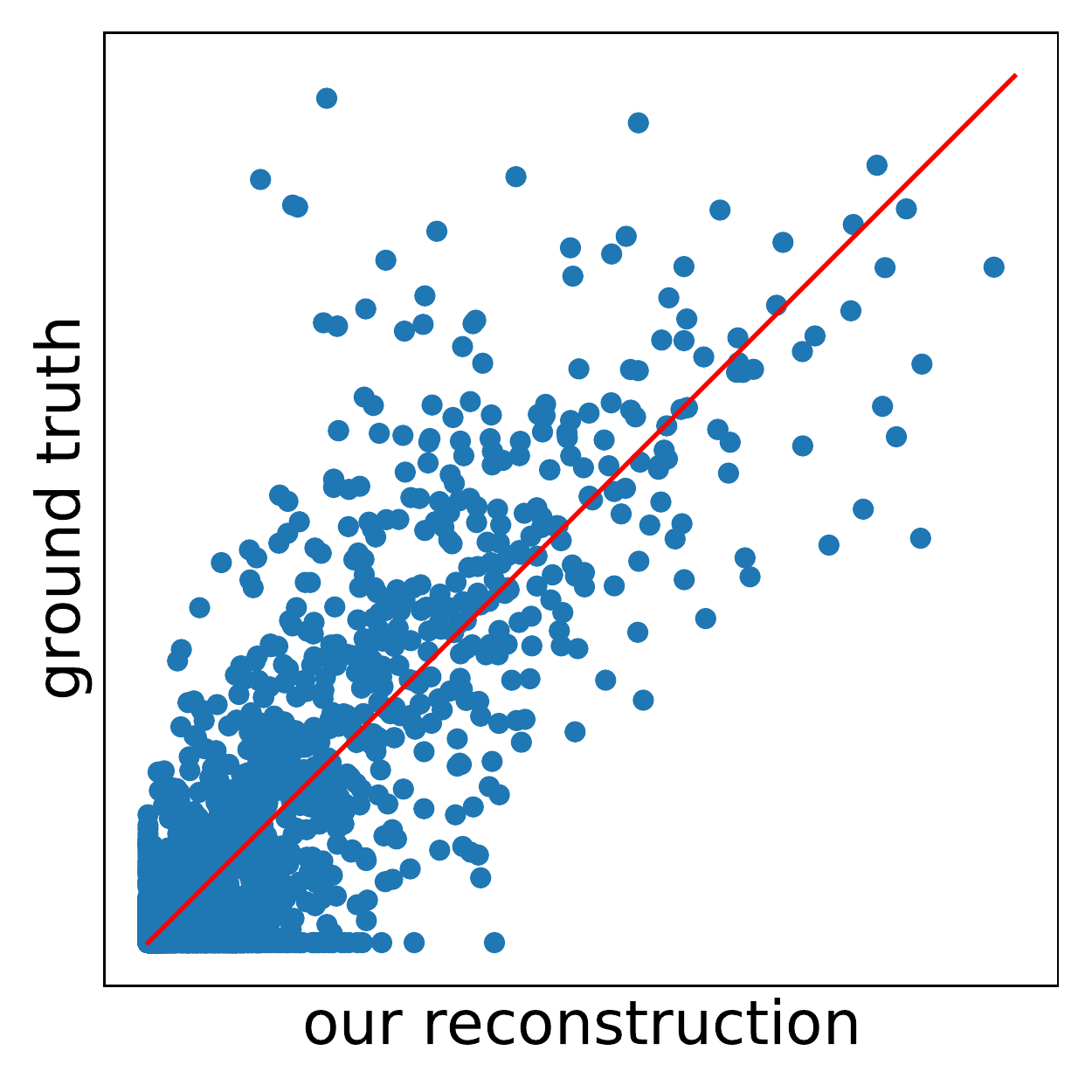}
  \end{minipage}
    \caption{3D plots of a cloud field scene. [Left] The reconstructed medium. [Middle] The true medium. [Right] Scatter plot of 10\% of the data points, randomly selected.}
    \label{fig:cloud_3d}
\end{figure}

\begin{figure}[b!]
    \centering
   \includegraphics[width=0.5\textwidth]{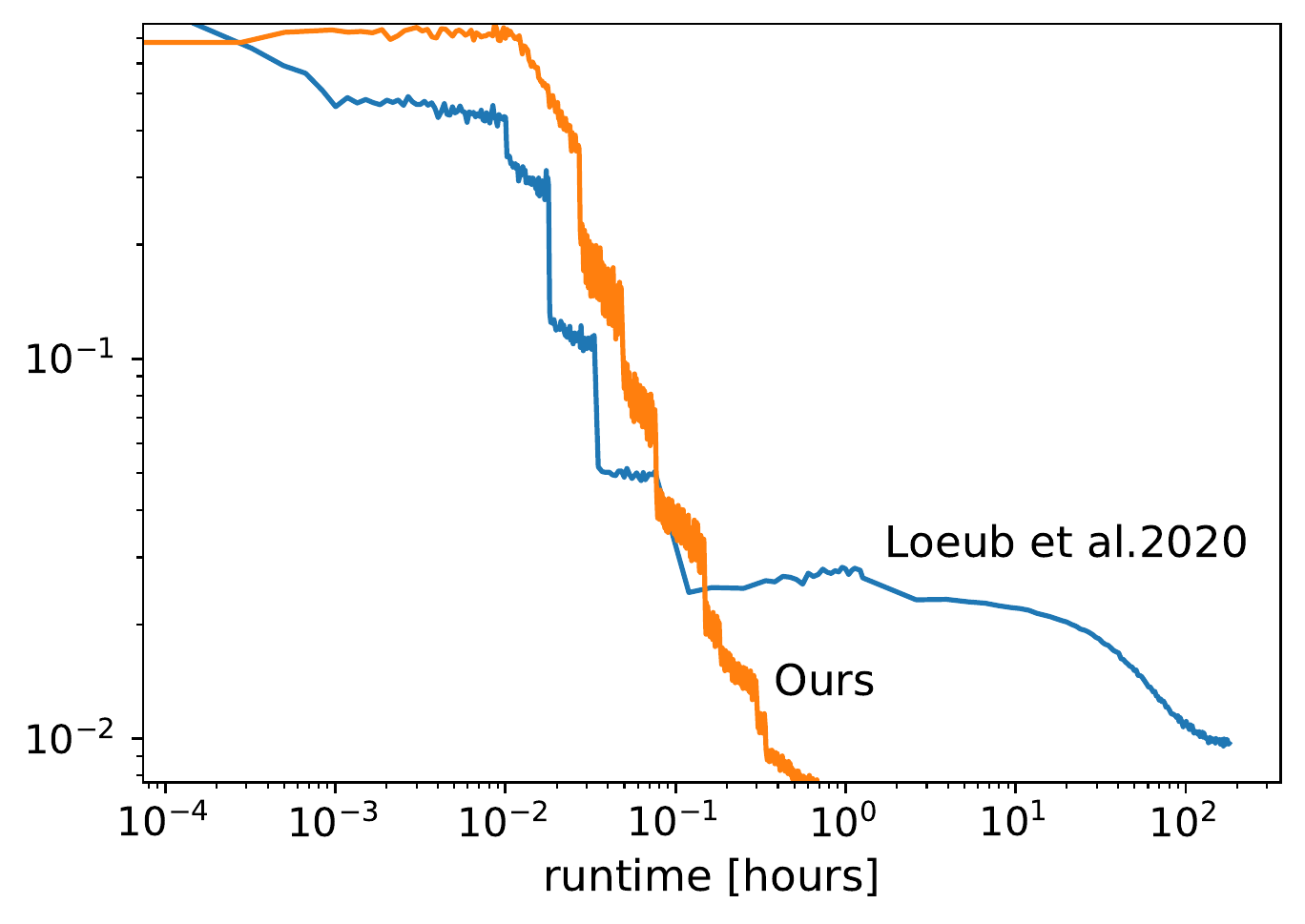}
    \caption{Loss reduction comparison with \citep{loeub2020monotonicity} cloud field scene. Our algorithm requires two orders of magnitude less runtime than \citep{loeub2020monotonicity} to achieve the same reduction in the loss function.}
    \label{fig:loss_reduction_comparison}
\end{figure}

\subsubsection{Results.}
We compare the cloud scenes with \citep{loeub2020monotonicity}. Their code is available at \citep{loeubCode}. We do not compare to the deterministic approach of \citep{levis2015airborne}, since it does not scale to large problems, as discussed in \citep{loeub2020monotonicity}. Quantitative results are displayed in Table \ref{table:quan_res}. Visual results are displayed in \figs{\ref{fig:smoke_convergence_grid}, \ref{fig:cloud_plane}, \ref{fig:cloud_3d}}. A loss reduction comparison with \citep{loeub2020monotonicity} is provided by \fig\ref{fig:loss_reduction_comparison}. Although the plume scene has more unknowns, the reconstructed plume is the most accurate. This is due to two reasons. One reason is that the wider zenith angle of the camera ring yields better sensing of the plume. The other reason is that the plume is much less dense than the clouds (see Table.~\ref{table:scene_settings}), which results in a low standard deviation of the estimators.

We compare our reconstruction of a single cloud with \citep{loeub2020monotonicity}, for the same runtime (\fig\ref{fig:loss_reduction_comparison}). Both the quantitative and visual results of our method are superior. The cloud field reconstruction reached convergence one order of magnitude faster using our method. From this comparison, we conclude that path recycling, together with an efficient \ac{GPU} implementation significantly speed up cloud tomography.

\begin{table}[t!]
\centering
\begin{tabular}{c | c | c | c} 
 \hline
 Scene & $\epsilon$ & $\delta$ & runtime [hours]\\ [0.5ex] 
 \hline\hline 

 Plume  & \smokereldist \% (ours)  & \smokerelbias \%  & 3  \\
 \hline
 
 Single Cloud & \jplreldist \% (ours)   & \jplrelbias \%  & 6.5  \\
  & 84.5 \% \citep{loeub2020monotonicity} & -5.5 \%   & 6.5 \\
 
  \hline
 Cloud Field   & \smallcfreldist \% (ours)   & \smallcfrelbias \%  & 127  \\
 & 44.7 \% \citep{loeub2020monotonicity} & -28.8 \%   & 15 \\
 \hline
\end{tabular}
 \caption{Quantitative results. This table provides our quantitative results and runtime for all three scenes together with a comparison to \citep{loeub2020monotonicity}.}
 \label{table:quan_res}
\end{table}

\begin{figure}[t]
  \centering
  \begin{minipage}[b]{0.5\textwidth}
  \includegraphics[width=\textwidth]{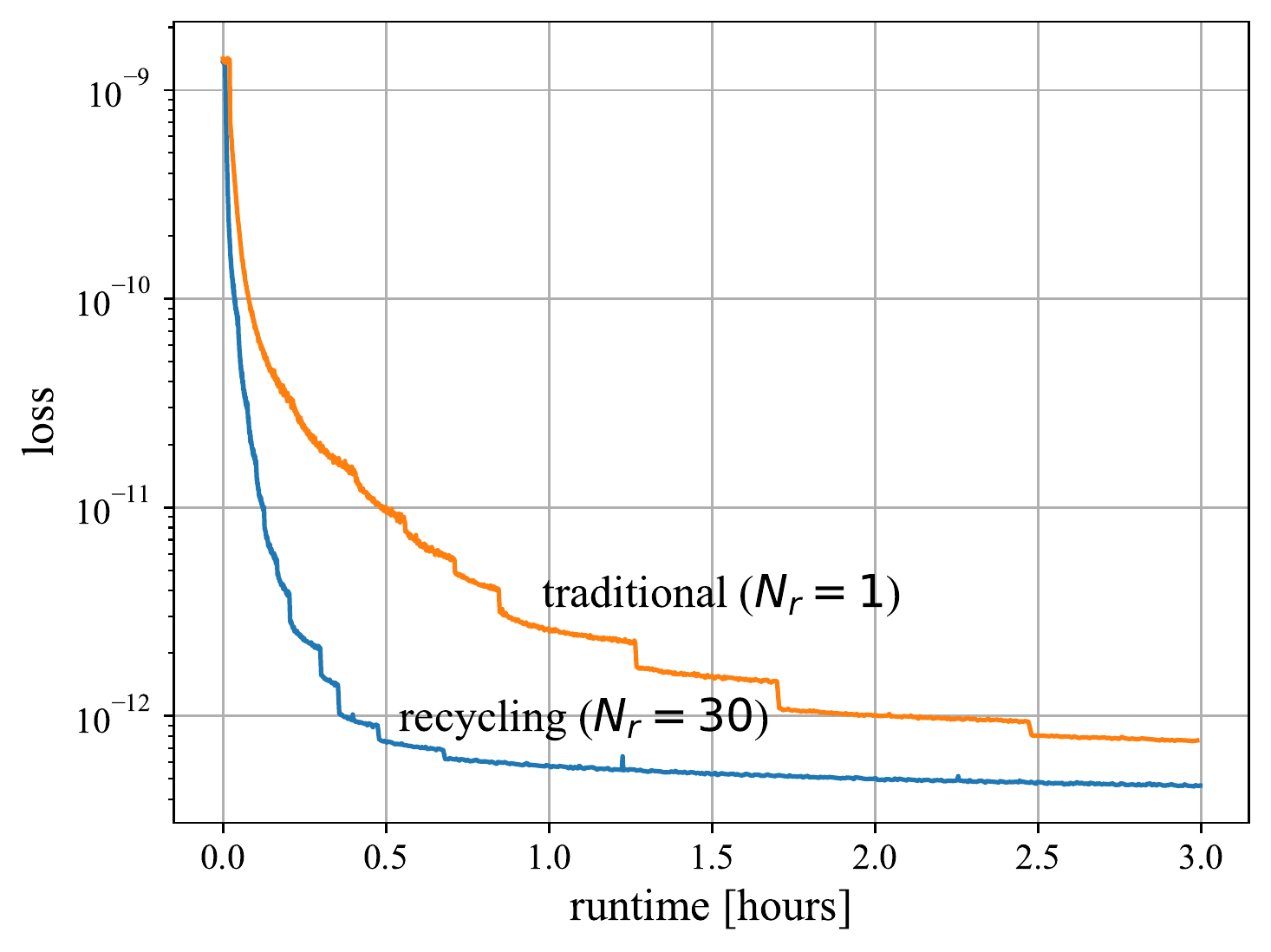}
  \end{minipage}
    \begin{minipage}[b]{0.48\textwidth}
   \includegraphics[width=\textwidth]{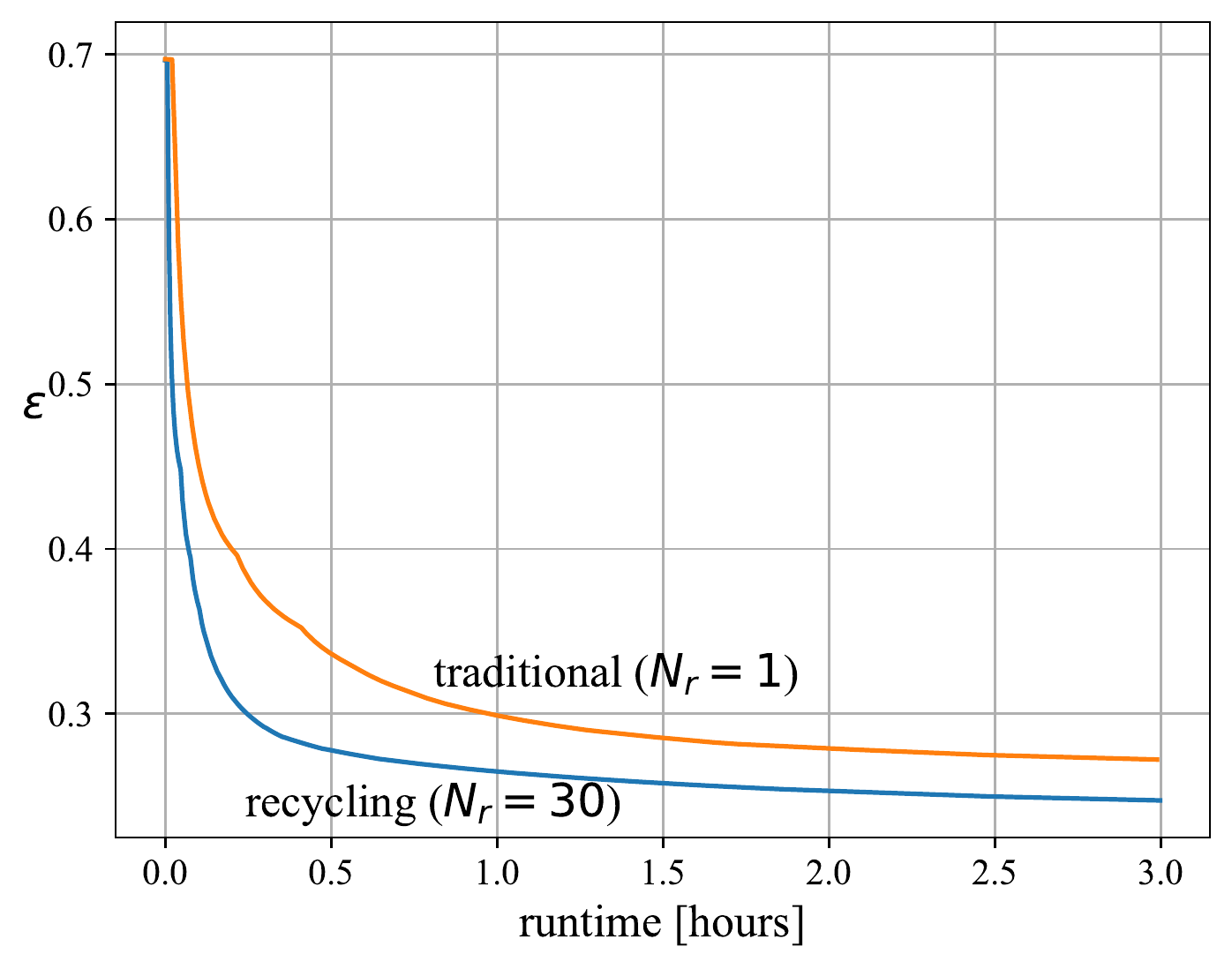}
  \end{minipage}
    \caption{Path recycling speed-up in the plume scene. [Left] Loss vs time. [Right] $\reldist$ vs time. Both simulations use our rendering engine. While for $\Nr=1$, each iteration contributes more to the reduction of the error measures, with $\Nr=30$ we can run more iteration per second.}
    \label{fig:loss_reldist_convergence}
\end{figure}

To validate that path recycling itself contributes to runtime reduction (and not only a \ac{GPU}), we ran an identical reconstruction simulation of the plume scene but with recycling period $\Nr=1$. The results are displayed in \fig\ref{fig:loss_reldist_convergence}.

\subsection{Reflectometry.}

\begin{figure}[t!]
  \centering
  \includegraphics[width=0.9\linewidth]{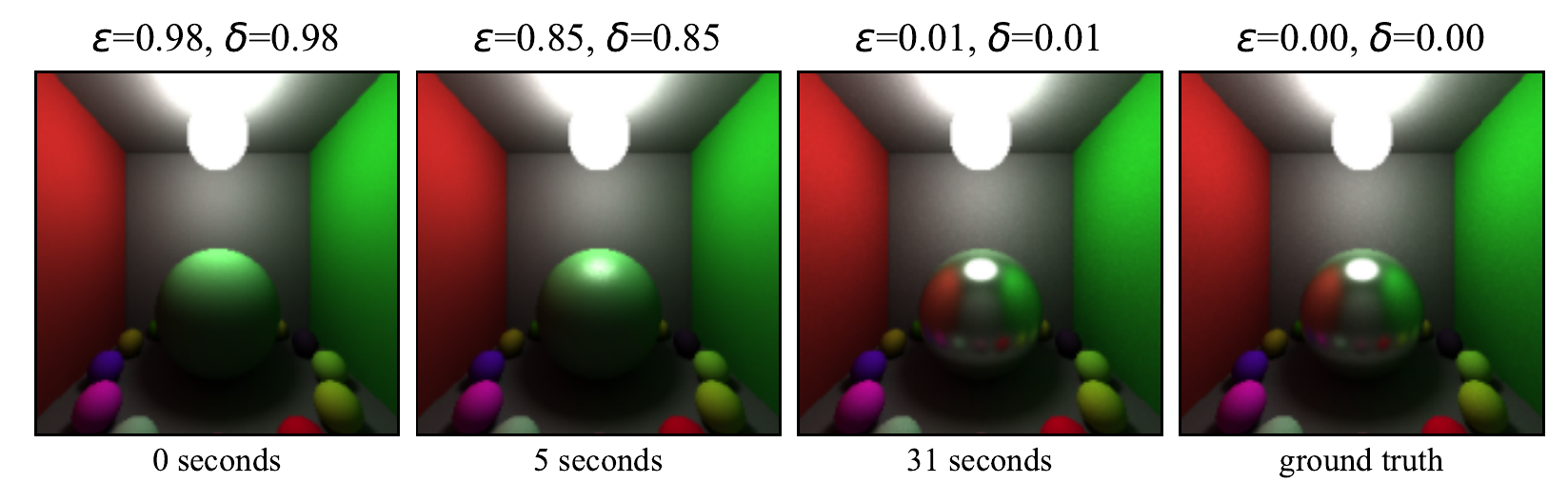} 
    \caption{Surface scene convergence. each image shows a different stage of the optimization. Convergence is reached after only 31 seconds with $\reldist=1\%$. Path sorting was not included in this test.}
    \label{fig:surface_convergence}
\end{figure} 
\begin{figure}[b!]
  \centering
  \begin{minipage}[b]{0.5\textwidth}
  \includegraphics[width=\textwidth]{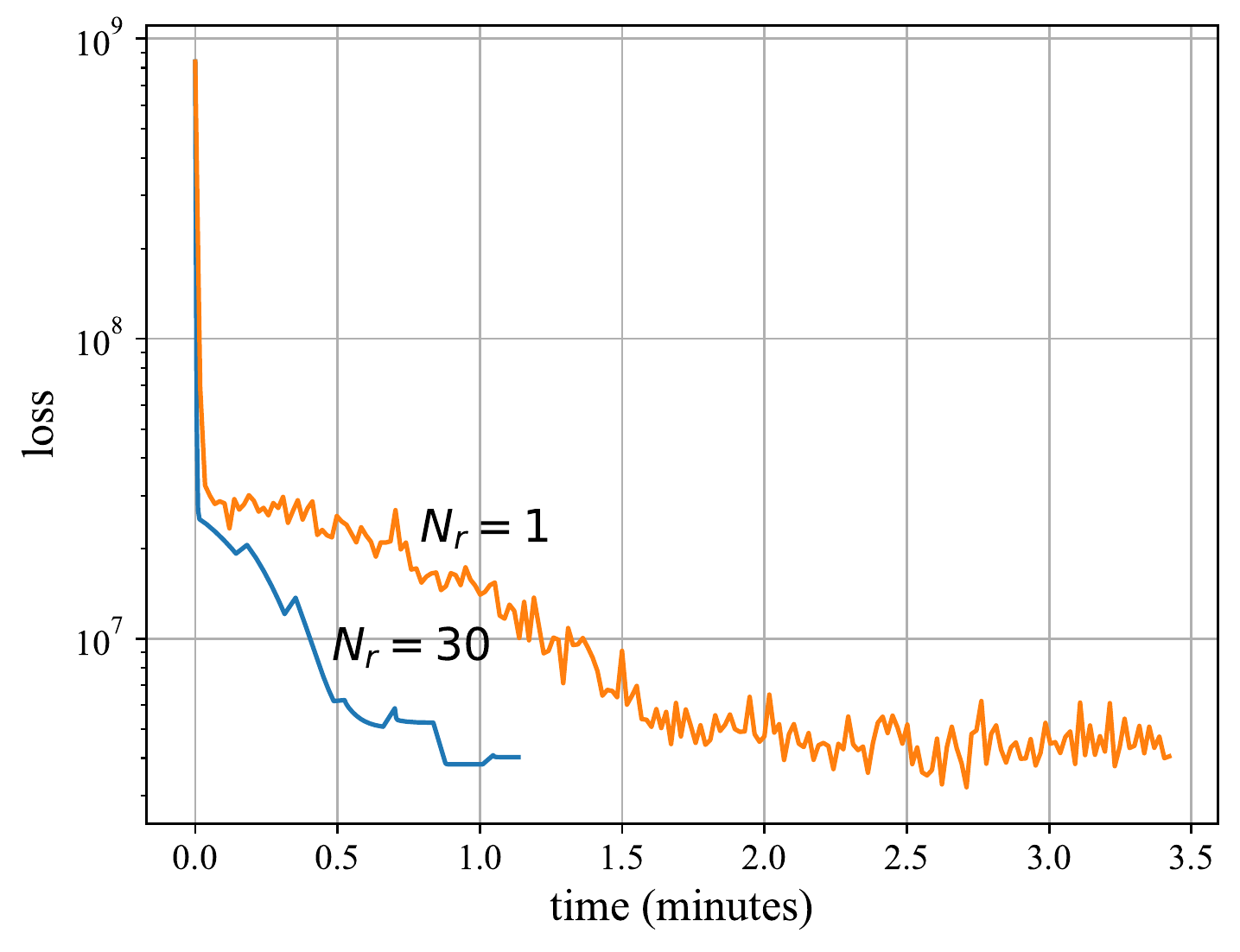}
  \end{minipage}
    \begin{minipage}[b]{0.48\textwidth}
   \includegraphics[width=\textwidth]{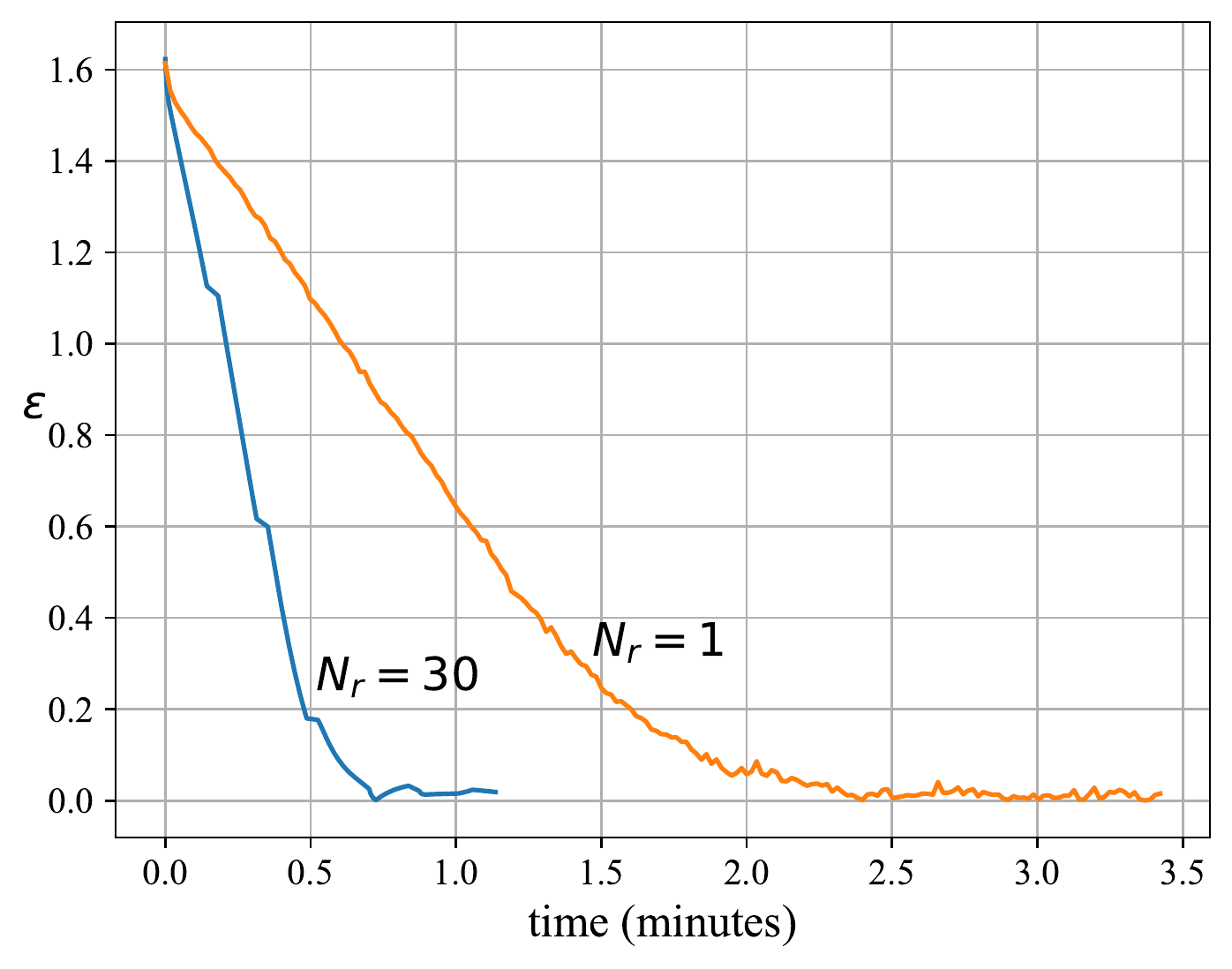}
  \end{minipage}
    \caption{Path recycling speed-up in the surface scene. [Left] Loss vs time. [Right] $\reldist$ vs time. In this test we used $\kappa_s=0.5$ and $\gamma=30$ as the true values for the \ac{BRDF}.}
    \label{fig:loss_reldist_convergence_surfaces}
\end{figure}
In this test, we consider a small box that contains 14 small diffuse spheres \citep{Smallpaint}. At the center, a large sphere is located with a variable \ac{BRDF}. We use the Phong reflection model \citep{phong1975illumination} with no absorption, which has the following parametric \ac{BRDF}:
\begin{equation}
\label{eq:phong}
\brdf\regpar{\locationc, \direction\rightarrow\direction'} = 1-\kappa_s + \kappa_s \cdot (\direction\cdot\direction')^\gamma,
\end{equation}
where $\gamma\geq0$ and $0\leq \kappa_s \leq 1$. The goal is to reconstruct the \ac{BRDF} of the large central sphere, by estimating $\kappa_s$ and $\gamma$. In this experiment, we set the true values to $\kappa_s=0.7$ and $\gamma=50$. We capture the scene using one camera having $60\times60$ pixels and use $N=1.8\times10^6$ light paths. The results are displayed in \fig\ref{fig:surface_convergence}. The plots in \fig\ref{fig:loss_reldist_convergence_surfaces} similarly show that $\Nr=30$ converges faster than $\Nr=1$. While this test is simple, it indicates that path recycling is potentially beneficial for reflectometry.

\section{Discussion}
\label{sec:conclusion}
Using paths that were sampled according to different scene parameters increases the standard deviation of a rendering or gradient estimator, as if less light path are used. However, recycling allows a \ac{GPU} to efficiently run iterated simulations. A different set of light paths is needed for a non-biased gradient (\eqsnopar{\ref{eq:gradient_simualtion}, \ref{eq:forward_simualtion}}). On the other hand, we use the same paths for several iterations, hence biasing the gradient. However, in practice, this bias is not noticeable in the stochastic optimization process. 

Our path recycling method is orthogonal to \citep{nimier2019mitsuba, nimier2020radiative} and our rendering engine does not attempt to serve as an alternative to Mitsuba 2. We have implemented our own engine because adaptation of the public code of \citep{nimier2020radiative} for heterogeneous scattering media is not simple. In principle, our method can be implemented over Mitsuba 2 and thus be used in most of its applications.

We believe that this approach can speed a diverse range of inverse rendering tasks, which were not demonstrated in this paper. For example, X-ray computed tomography through scattering \citep{geva2018x} can benefit from our framework. \citep{geva2018x} uses \ac{MC} for performing X-ray \ac{CT} of the human body while considering the scattered radiation. We believe that the analogies between cloud \ac{CT} and X-ray \ac{CT} can exploit path recycling for these benefits as well. Another example is texture reconstruction \citep{nimier2020radiative}, which seeks to retrieve a texture of a specific object in the scene. In this problem, the nature of the light path does not change between iterations. This is because the texture does not affect the geometry of the path. In applications as such, path recycling could be highly beneficial since the light paths do not lose their relevancy throughout the optimization. 

\section{ACKNOWLEDGMENTS}
This research is funded by the European Research Council (ERC) under the European Unions Horizon 2020 research and innovation program (grant agreement No 810370: CloudCT).
\newpage
\begingroup
\renewcommand\arraystretch{1.4}
\begin{center}
\begin{longtable}[c]{|c|c|} 
 \hline
 Symbol & Definition \\
 \hline
 \hline
 
$A$ & Rectangle area \\ \hline



$D$ & Foreshortening factor \\ \hline
$d\ppath$ & Path differential \\ \hline 

$\mathbb{E}$ & Expected value \\ \hline

$\forwardmodel$ & Forward model vector \\ \hline
$\forwardmodel_{\nm}$ & $\nm$'th measurement Forward model \\ \hline
$\hat{\forwardmodel}$ & Estimated forward model vector\\ \hline
$\contfunc$ & Contribution function\\ \hline
$\contfunc_{\nm}$ & Contribution function of the $\nm$'th measurement\\ \hline
$\phasefunc$ & Phase function\\ \hline
$\brdf$ & \ac{BRDF}\\ \hline
$\anglefunc$ & Scattering function\\ \hline
 
$G$ & Geometric term\\ \hline
$\gradest$ & Gradient estimation \\ \hline
$\segfunc_{\segind}$ & Segment Contribution function \\ \hline


$i$ & Sample index \\ \hline
$I$ & Pixel measurement\\ \hline
$\gtmeasurements$ & Ground-Truth Measurements \\ \hline 

$\ptype$ & Particle type index \\ \hline

$\segind$ & Segment index \\ \hline
$\segind'$ & Next event index \\ \hline
$\pathsize$ & Path size \\ \hline

$\radiance$ & Radiance \\ \hline
$\radiancee$ & Emitted radiance \\ \hline
$\radiances$ & in-scattered radiance \\ \hline
$\lossfunc$ & Loss function \\ \hline
$\distance$ & Distance \\ \hline
$\distance_{\segind,\voxel}$ & Distance of segment $\segind$ and voxel $\voxel$\\ \hline

$\sceneparam$ & Scene unknown vector\\ \hline
$\sceneparamt$ & Scene unknown vector at iteration $\iter$\\ \hline
$\sceneparef$ & Reference of a scene unknown vector \\ \hline
$\sceneparam^*$ & True Scene unknown vector\\ \hline
$\Hat{\sceneparam}$ & Estimated Scene unknown vector\\ \hline
$\scenescalar_{\unknownindex}$ & Scene unknown \\ \hline

$\nm$ & Measurement index \\ \hline
$\normal$ & Surface normal \\ \hline
$\Np$ & Number of samples\\ \hline
$\Nm$ & Number of scene unknowns \\ \hline
$\Npar$ & Particle types number\\ \hline
$\Nr$ & Recycling period\\ \hline


$P$ & Particle probability function \\ \hline
$\nexteventpdf$ & Local Estimation \ac{PDF} \\ \hline
$\unipdf$ & Uniform \ac{PDF} \\ \hline
$\pathset$ & Path space \ac{PDF} \\ \hline

$Q_n$ & Local estimation contribution \\ \hline
$\correctfact_{\nm}$ & Correction factor \\ \hline
$\ray$ & Ray \\ \hline

$\sphere$ & Unit sphere \\ \hline
$\scorefunc_{\unknownindex,\nm}$ & Path score function \\ \hline

$\iter$ & Iteration \\ \hline
$\maxiter$ & Max iteration \\ \hline
$T$ & Transmittance \\ \hline

$\mcpar$ & Uniform random variable \\ \hline
$\mcpar_i$ & Uniform samples \\ \hline
$\mathcal{U}$ & Uniform distribution \\ \hline

$\unknownindex$ & Unknown index \\ \hline
$\voxelset_{\nm}$ & Domain of voxel $\voxel$ \\ \hline

$\pixmeasure$ & Sensor response function \\ \hline

$\location$ & 3D Location\\ \hline
$\sensloc$ & Sensor location\\ \hline
$\location_k$ & Path vertex\\ \hline
$\lastloc$ & Last scattering location of the $i$'th path \\ \hline
$\lightloc$ & Light source location \\ \hline
$\ppath$ & Light path\\ \hline

$\locb$ & Line parameterization \\ \hline
$\locationb$ & 3D Location\\ \hline

$\locc$ & Line parameterization \\ \hline
$\locationc$ & Surface 3D location\\ \hline

$\stepsize$ & Step Size \\ \hline

$\extt$ & Extinction coefficient\\ \hline
$\exta$ & Absorption coefficient\\ \hline
$\exts$ & Scattering coefficient\\ \hline
$\extt^{\regpar{j}}$ & Extinction coefficient of particle type $\ptype$ \\ \hline
$\extcloud_{\unknownindex}$ & Cloud extinction coefficient of voxel $\unknownindex$\\ \hline
$\extair_{\unknownindex}$ & Air extinction coefficient of voxel $\unknownindex$\\ \hline
$\gamma$ & Phone glossiness coefficient\\ \hline

$\relbias$ & Relative bias \\ \hline 

$\reldist$ & Relative distance \\ \hline



$\dangle$ & Angle\\ \hline
$\dangle_{\segind}$ & Scattering angle of $\segind$'th segment\\ \hline
$\dangle_{\segind,d}$ & Scattering angle towards the sensor \\ \hline

$\kappa_s$ & Phong specular coefficient\\ \hline


$\ispdf_{\nm}$ & Importance sampling \ac{PDF} \\ \hline







$\tau$ & Optical distance \\ \hline 





$\direction$ & Irradiance direction\\ \hline
$\direction'$ & Radiance direction\\ \hline
$\direction_k$ & Scattering direction of $k$'th segment\\ \hline
$\hemisphere$ & Unit hemisphere \\ \hline
$\sinscat$ & Single scattering albedo \\ \hline
$\sinscat$ & Single scattering albedo \\ \hline
$\sinscatpar{j}$ & Single scattering albedo of particle type $\ptype$ \\ \hline
\end{longtable}
\end{center}
\endgroup
\newpage

\bibliographystyle{unsrtnat}
\bibliography{references}  





\newpage
\appendix
\section{Monte-Carlo Integration}
\label{sec:Monte-Carlo}
In this section, we give several examples that illustrate \ac{MC} integration. In the following examples, we consider a one-dimensional function, but the generalization to an arbitrary dimension is simple.

Let $f:\squarepar{0,1}\rightarrow \mathbb{R} $ be a one dimensional function. We are interested in calculating the following integral:
\begin{equation}
\label{eq:one_d_integral}
I = \intop_0^1{f(\mcpar)d\mcpar}.
\end{equation}
\subsection{Riemann Sum.}
\begin{figure}[t]
  \centering
  \includegraphics[width=0.6\linewidth]{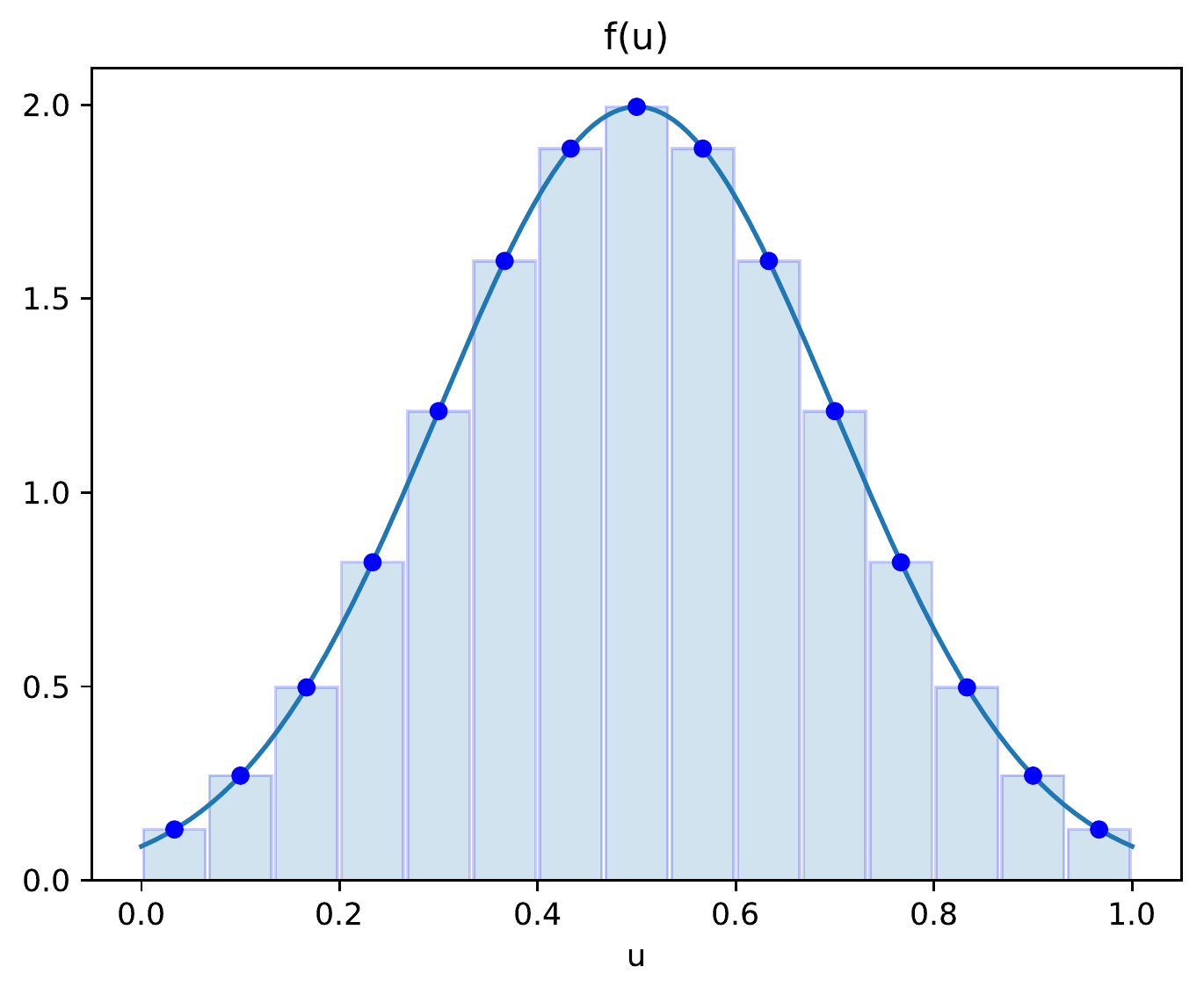}
    \caption{Approximating the integral of the function $f\regpar{\mcpar}$ (dark blue) by dividing the segment $\squarepar{0,1}$ to 15 rectangles (light blue).}
    \label{fig:riemann}
\end{figure}

Before suggesting a \ac{MC} technique, one can trivially approximate the integral in equation \eq{\ref{eq:one_d_integral}} by dividing the segment $[0,1]$ to $N$ partitions of the same length $\frac{1}{N}$, as illustrated in \fig\ref{fig:riemann}. Then, the integral can be approximated by:
\begin{equation}
\label{eq:one_d_sum}
I \approx \sum_{\pathindex=1}^N{A_i},
\end{equation}
where $A_i$ is the area of a rectangle whose width is the $i$'th partition and its length is bounded by the corresponding function value.
The limit $N\rightarrow \infty$ yields the integral definition as a Riemann sum. While this method converges fast for 1-dimensional integrals, it suffers from a slow convergence rate for higher dimensions. This is where \ac{MC} shines. 

\subsection{Uniform Sampling.}
\begin{figure}[t]
  \centering
  \includegraphics[width=0.6\linewidth]{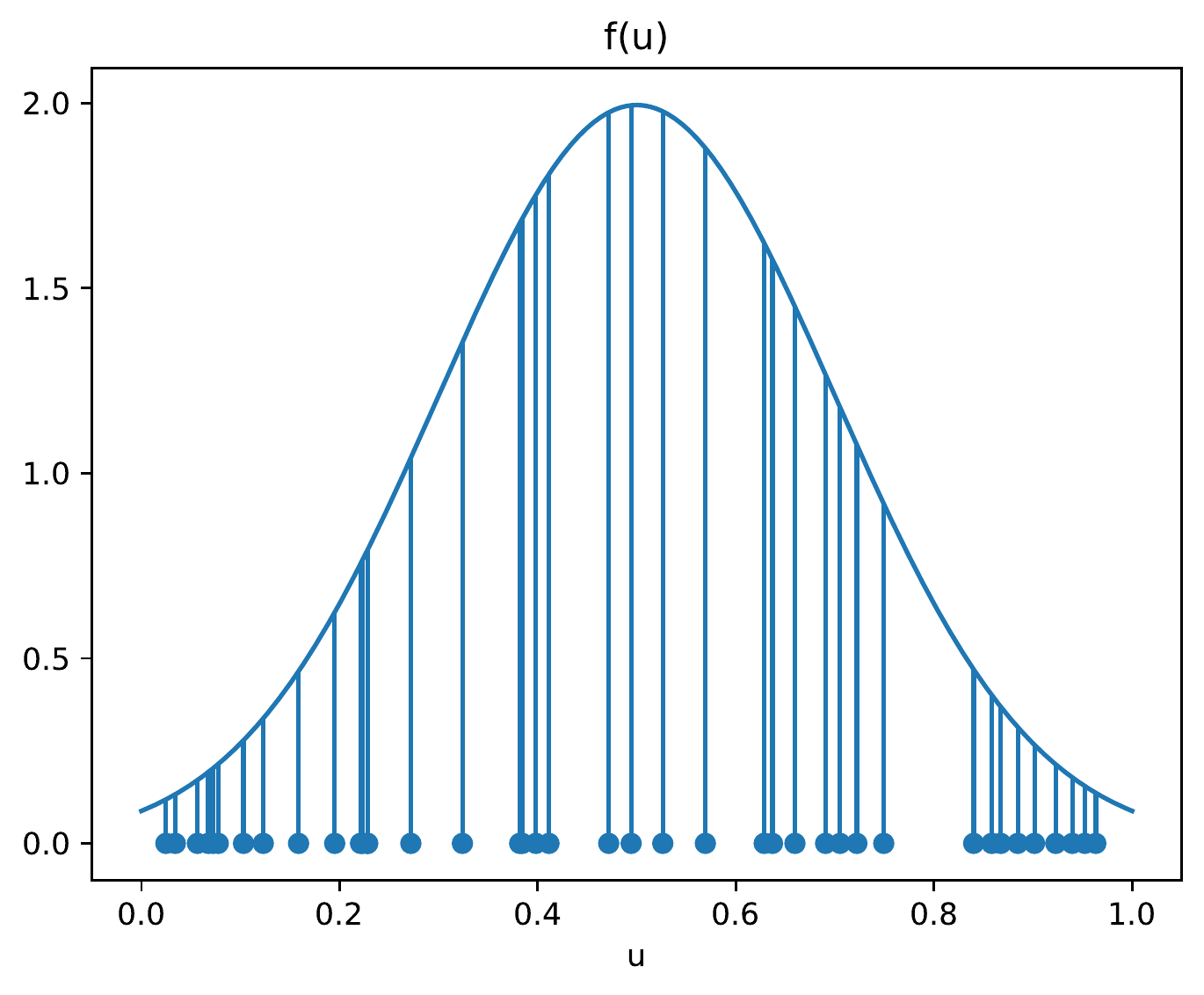}
    \caption{Estimating the integral of the function $f\regpar{\mcpar}$ using uniform sampling.}
    \label{fig:MC}
\end{figure}

\ac{MC} integration estimates the integral by exploiting the similarity between a deterministic integral and the definition of a random variable expectation. Consider $N$ \ac{I.I.D} random samples from a uniform distribution over the segment $[0,1]$, which are defined as follows:
\begin{equation}
\curlpar{\mcpar_i}_{\pathindex=1}^N \simiid \mathcal{U}[0,1].
\end{equation}
The corresponding \ac{PDF} is:
   \begin{equation}
  \unipdf\regpar{\mcpar}=
  \begin{cases}
   1 & \text{if } 0 \leq \mcpar \leq1 \\
   0 & \text{else}
  \end{cases}.
  \end{equation}
  \ac{MC} suggests \citep{veach1997robust} the following estimator, illustrated in \fig\ref{fig:MC}: 

\begin{equation}
\label{eq:MC_1d_estimator}
I \approx \frac{1}{N}\sum_{\pathindex=1}^N{f\regpar{\mcpar_i}}.
\end{equation}
 This estimator can be analogously compared to \eq{\ref{eq:one_d_sum}} but with rectangles at random locations. Denote $\mathbb{E}\squarepar{\cdot}$ as the expected value operator. \ac{MC} relies on the law of large numbers from probability theory, which claims that:
 \begin{equation}
 \label{eq:mc_lim}
    \lim_{N\rightarrow\infty}\frac{1}{N}\sum_{\pathindex=1}^N{f\regpar{\mcpar_i}} = \mathbb{E}\squarepar{f\regpar{\mcpar}}.
 \end{equation}
By plugging the expectation definition we achieve convergence
  \begin{equation}
     \mathbb{E}\squarepar{f\regpar{\mcpar}}=\intop_{-\infty}^{\infty}{f\regpar{\mcpar}\cdot \unipdf\regpar{\mcpar}d\mcpar} = \intop_0^1{f\regpar{\mcpar}d\mcpar} = I .
 \end{equation}
 
\subsection{Importance Sampling.}
\label{sec:IS}
\begin{figure}[t]
  \centering
  \begin{minipage}[b]{0.44\textwidth}
  \includegraphics[width=\textwidth]{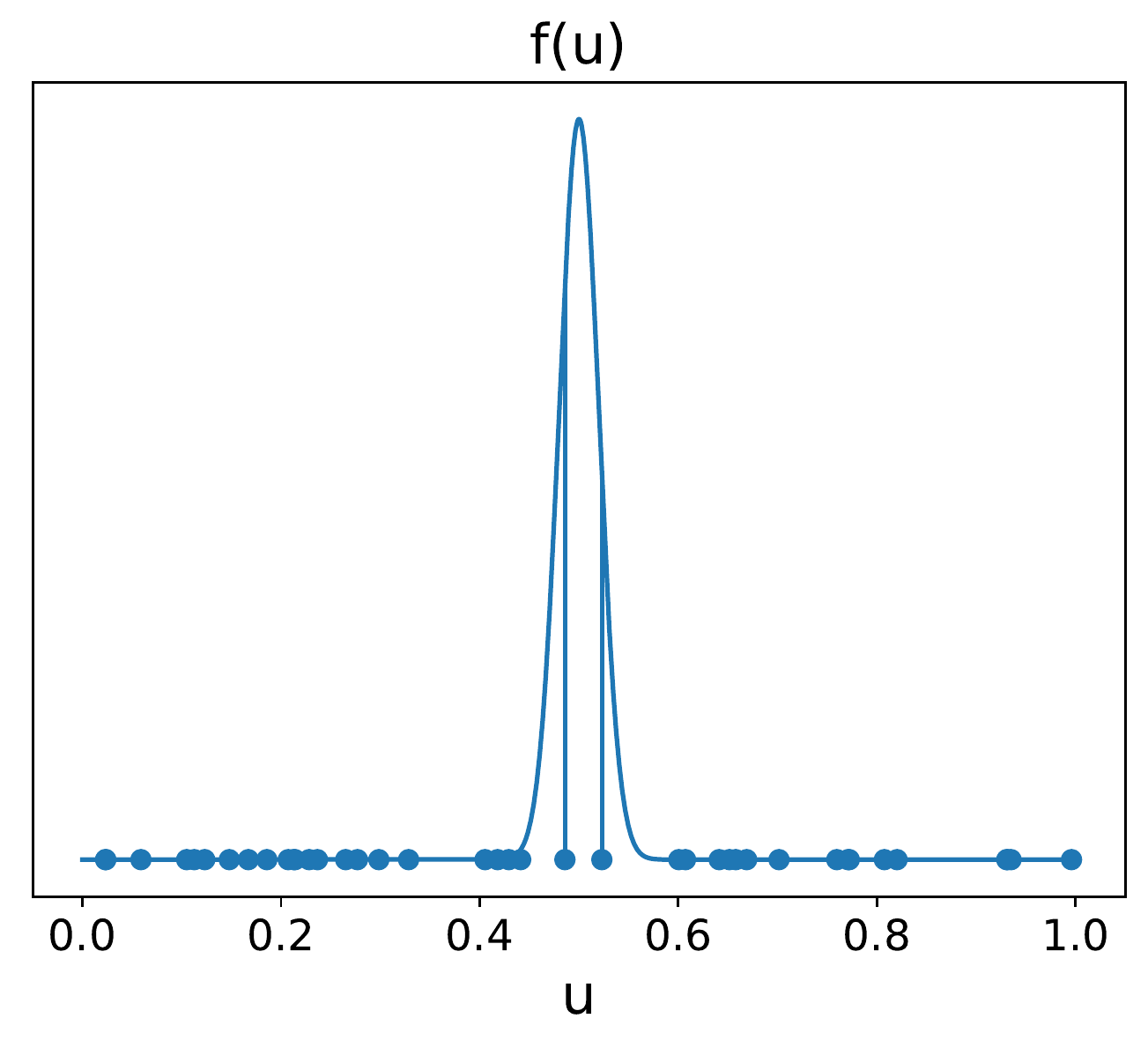}
  \end{minipage}
    \begin{minipage}[b]{0.44\textwidth}
  \includegraphics[width=\textwidth]{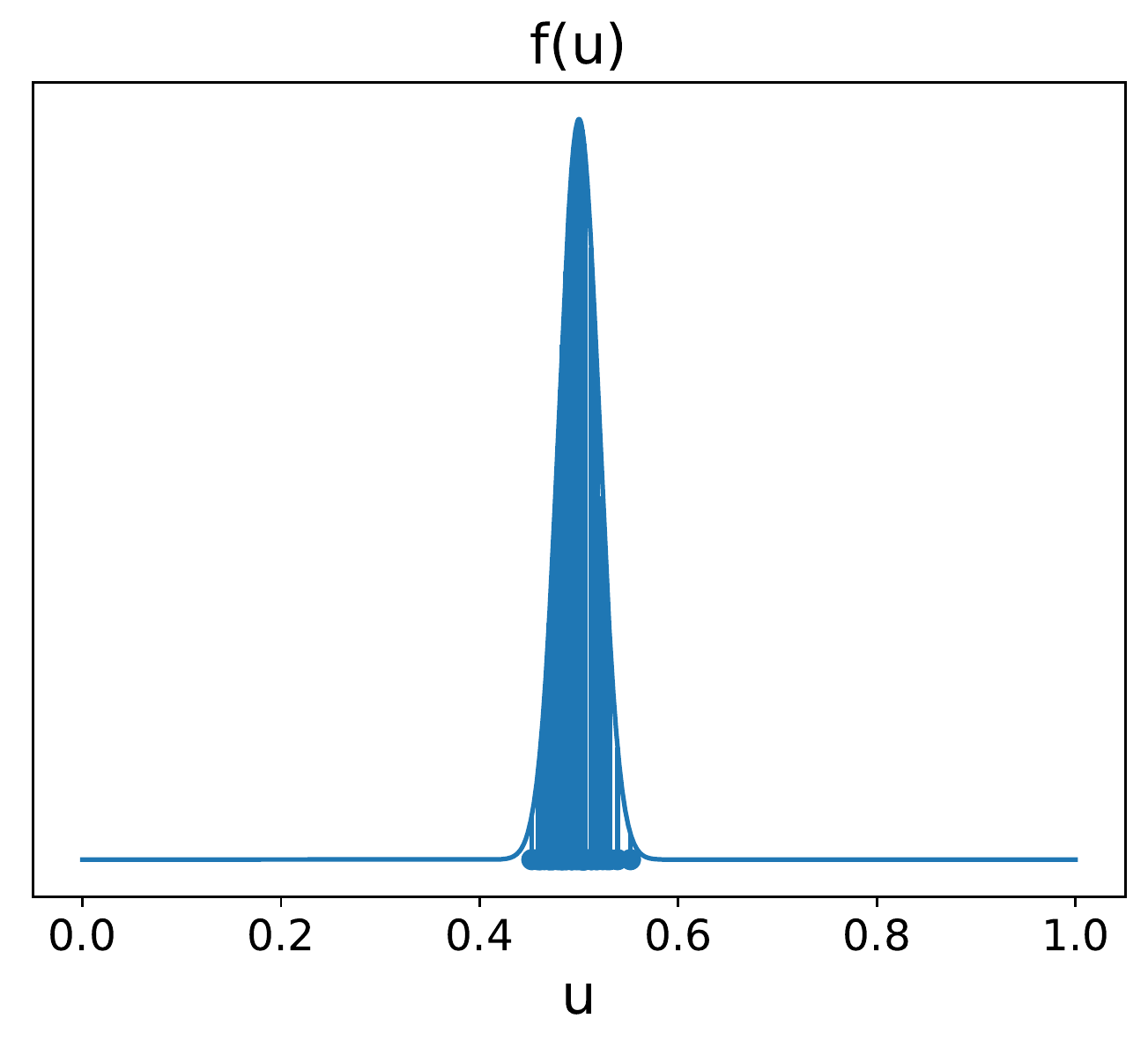}
  \end{minipage}
    \caption{Illustration of a function that is not suitable for uniform sampling. [Top] Uniform sampling misses the important features of $f\regpar{\mcpar}$. [Bottom] Importance sampling samples the random variable from a distribution $\ispdf$ that is similar to $f$. 
    }
    \label{fig:IS}
\end{figure}

Uniform sampling is an acceptable choice for well-behaved functions. However, in cases where the function $f$ has a range including both extremely low and high values, the convergence rate of \eq{\ref{eq:MC_1d_estimator},\ref{eq:mc_lim}} significantly slows down. This is seen in \fig\ref{fig:IS}. This problem can be solved by sampling the random variables from a different \ac{PDF} that considers the shape of $f$.

Let $\ispdf$ be a general \ac{PDF}. Let $\curlpar{\mcpar_i}_{\pathindex=1}^N \simiid \ispdf$ be $N$ \ac{I.I.D} random samples drawn from $\ispdf$. \ac{MC} suggests the following estimator:
\begin{equation}
\label{eq:IS_1d_estimator}
I \approx \frac{1}{N}\sum_{\pathindex=1}^N{\frac{f\regpar{\mcpar_i}}{\ispdf\regpar{\mcpar_i}}}.
\end{equation}
This is a generalization of \eq{\ref{eq:MC_1d_estimator}}. \eq{\ref{eq:IS_1d_estimator}} uses an arbitrary \ac{PDF} to sample the random variables. The division by $\ispdf\regpar{\mcpar_i}$ is a correction factor, due to the fact that some samples have a higher probability to be sampled than other samples. If $\ispdf$ yields high probability of drawing a sample, the correction factor reduces the contribution of this particular sample (and vice versa).

Notice that the estimator in \eq{\ref{eq:IS_1d_estimator}} is unbiased since:
\begin{equation}
\label{eq:unbiasedness}
\begin{split}
\mathbb{E}\squarepar{\frac{1}{N}\sum_{\pathindex=1}^N{\frac{f\regpar{\mcpar_i}}{\ispdf\regpar{\mcpar_i}}}} & = \frac{1}{N}\sum_{\pathindex=1}^N{\mathbb{E}\squarepar{\frac{f\regpar{\mcpar_i}}{\ispdf\regpar{\mcpar_i}}}} \\
& = \frac{1}{N}\sum_{\pathindex=1}^N{\intop_0^1{\frac{f\regpar{\mcpar}}{\cancel{\ispdf\regpar{\mcpar}}}\cdot \cancel{\ispdf\regpar{\mcpar}}d\mcpar}} \\
& = \frac{1}{N}\sum_{\pathindex=1}^N{I} = I .
\end{split}
\end{equation}
\eq{\ref{eq:unbiasedness}} proves that any choice of $\ispdf$ results in an unbiased estimator of $I$; however, we are interested in minimizing the estimation error. It can be shown \citep{veach1997robust} that choosing $\ispdf$ to be similar to $f$ reduces the estimation error. In other words, we want to choose $\ispdf$ that draws in high probability samples that attain high values of $f$. This procedure is called \textit{Importance Sampling}.


\section{Derivation of \ac{PDF}s in {\em Path Tracing}}
This section extends the derivation of $\ispdf_{\nm}$ from Section \ref{sec:MC_est_path_integral} by deriving \eq{\ref{eq:final_mu}}. 
\label{sec:ptpdf}
The \ac{PDF} that corresponds to the distance sampling procedure from Section.~\ref{sec:distance_sampling} is:
\begin{equation}
    \ispdf_{\distance}\regpar{\distance_{\segind}|\location_{\segind-1},\direction_{\segind-1}}  = \extt\regpar{\location_{\segind-1}+\distance_{\segind}\direction_{\segind-1}} \trans{\location_{\segind-1}}{\location_{\segind-1}+\distance_{\segind}\direction_{\segind-1}}.
\end{equation}
Following distance sampling, we set 
\begin{equation}
\label{eq:dist_set_app}
    \location_{\segind}=\location_{\segind-1} + \distance_{\segind}\direction_{\segind}
\end{equation}
Then, the \ac{PDF} that corresponds to the direction sampling procedure
from Section.~\ref{sec:direction_sampling} is: 
\begin{equation}
    \ispdf_{\direction}\regpar{\direction_{\segind}|\direction_{\segind-1},\location_{\segind}}  = \phasefunc\regpar{\location_{\segind},\cos\dangle_{\segind-1,\segind}}.
\end{equation}
Therefore, from \eqs{\ref{eq:geomterm},\ref{eq:semgent_sampling},\ref{eq:dist_set_app}},
\begin{equation}
\ispdf_{\text{ray}}\regpar{\ray_{\segind}|\ray_{\segind-1}} = \extt\regpar{\location_{\segind}}\phasefunc\regpar{\location_{\segind},\cos\dangle_{\segind-1,\segind}}\cdot  \trans{\location_{\segind-1}}{\location_{\segind}}\geomterm\regpar{\location_{\segind-1},\location_{\segind}}.
\end{equation}
Finally, the total path \ac{PDF} is:
\begin{equation}
\label{eq:final_mu_app}
\ispdf_{\nm}\regpar{\ppath} =\ispdf_{\text{ray}}\regpar{\ray_0}\prod_{\segind=1}^{\pathsize}{\ispdf_{\text{ray}}\regpar{\ray_{\segind}|\ray_{\segind-1}}}=\frac{1}{4\pi}\prod_{\segind=1}^{\pathsize}{\extt\regpar{\location_{\segind}}\phasefunc\regpar{\location_{\segind},\dangle_{\segind-1,\segind}} \cdot\trans{\location_{\segind-1}}{\location_{\segind}}\cdot\geomterm\regpar{\location_{\segind-1},\location_{\segind}}}.
\end{equation}
Then, the forward model can be estimated by:
\begin{equation}
\label{eq:forward_approax_app}
I_{\nm} \approx \frac{1}{\Np}\sum_{\pathindex=1}^{\Np}{\frac{\contfunc_{\nm}\regpar{\pathi}}{\ispdf_{\nm}\regpar{\pathi}}}.
\end{equation}
Plugging \eqs{\ref{eq:cont_func_scattering},\ref{eq:final_mu}} in \eq{\ref{eq:forward_approax_app}}:
\begin{equation}
\label{eq:forward_approax2_app}
I_{\nm} \approx \frac{4\pi}{\Np}\sum_{\pathindex=1}^{\Np}{
\radiancee\regpar{\location_0,\directioni_0}\pixmeasure_{\nm}\regpar{\lastloc,\sensloc} \cdot  \prod_{\segind=1}^{\pathsize_i}{
\frac{\exts\regpar{\locationi_{\segind}}}{ \extt\regpar{\locationi_{\segind}}}}}.
\end{equation}
Plugging \eq{\ref{eq:eff_sinscat}} in \eq{\ref{eq:forward_approax2_app}}:
\begin{equation}
I_{\nm} \approx \frac{4\pi}{\Np}\sum_{\pathindex=1}^{\Np}{
\radiancee\regpar{\location_0,\directioni_0}\pixmeasure_{\nm}\regpar{\lastloc,\sensloc} \cdot \prod_{\segind=1}^{\pathsize_i}{
\sinscat\regpar{\locationi_{\segind}}}}.
\end{equation}

\section{Partial Derivatives of the Forward Model}
\label{sec:partial_deriv_app}
As discussed in Section \ref{sec:inverse}, computing the gradient of the forward model $\forwardmodel_{\nm}$ requires an analytical form of its partial derivatives with respect to the scene parameters $\sceneparam$.
Following \citep{gkioulekas2016evaluation}, this section provides a method for deriving $\frac{\forwardmodel_{\nm}\regpar{\sceneparam}}{\partial m_{\unknownindex}}$ .

Differentiating \eq{\ref{eq:forward_model}} with respect to $\scenescalar_{\unknownindex}$ results in:
\begin{equation}
\label{eq:int_derivative_app}
\frac{\partial \forwardmodel_{\nm}\regpar{\sceneparam}}{\partial \scenescalar_{\unknownindex}} = \frac{\partial}{\partial \scenescalar_{\unknownindex}}\intop_{\pathset}{\contfunc_{\nm}\regpar{\ppath|\sceneparam}d\ppath}
= \intop_{\pathset}{\frac{\partial \contfunc_{\nm}\regpar{\ppath|\sceneparam}}{\partial \scenescalar_{\unknownindex}}d\ppath}.
\end{equation}
Hence, deriving the derivative of the model degenerates to deriving the derivative of the \ac{MCF}. Recall that the \ac{MCF} is a multiplication of $\pathsize+1$ functions (\eqnopar{\ref{eq:cont_func}}), where $\pathsize$ is the path size. We follow \citep{gkioulekas2016evaluation} to derive a compact form that holds for an arbitrary $\pathsize$. 
First, consider the case where $\pathsize=2$. Then, $\contfunc = \pixmeasure_{\nm}  \segfunc_{0} \segfunc_{1}$ and
\begin{equation}
\derivx{\contfunc_{\nm}\regpar{\ppath|\sceneparam}}{\scenescalar_{\unknownindex}}  = \deriv{\scenescalar_{\unknownindex}}\regpar{ \pixmeasure_{\nm}  \segfunc_{0} \segfunc_{1}} = \derivx{\pixmeasure_{\nm}}{\scenescalar_{\unknownindex}}  \segfunc_{0}  \segfunc_{1} + \pixmeasure_{\nm}  \derivx{\segfunc_{0}}{\scenescalar_{\unknownindex}} \segfunc_{1} + \pixmeasure_{\nm}  \segfunc_{0}  \derivx{\segfunc_{1}}{\scenescalar_{\unknownindex}}.
\end{equation}
Thus
\begin{equation}
\label{eq:derivative_example_app}
\begin{split}
\derivx{\contfunc_{\nm}\regpar{\ppath|\sceneparam}}{\scenescalar_{\unknownindex}} &=\pixmeasure_{\nm}  \segfunc_{0}  \segfunc_{1}  \dfrac{\derivx{\pixmeasure_{\nm}}{\scenescalar_{\unknownindex}}}{\pixmeasure_{\nm}} + \pixmeasure_{\nm}  \segfunc_{0}  \segfunc_{1}  \dfrac{\derivx{\segfunc_{0}}{\scenescalar_{\unknownindex}}}{\segfunc_{0}}+ \pixmeasure_{\nm}\segfunc_{0}  \segfunc_{1}  \dfrac{ \derivx{\segfunc_{1}}{\scenescalar_{\unknownindex}}}{\segfunc_{1}} \\
&= \contfunc_{\nm}\regpar{\ppath|\sceneparam}  \regpar{\dfrac{\derivx{\pixmeasure_{\nm}}{\scenescalar_{\unknownindex}}}{\pixmeasure_{\nm}} + \frac{\derivx{\segfunc_0}{\scenescalar_{\unknownindex}}}{\segfunc_0} + \frac{\derivx{\segfunc_1}{\scenescalar_{\unknownindex}}}{\segfunc_1}}
\end{split}.
\end{equation}
Hence, the derivative is the \ac{MCF} multiplied by a term that depends on the \ac{SCF}s and their derivatives. Using the \ac{PSF} definition in \eq{\ref{eq:total_derivative_factor}}, for an arbitrary $\pathsize$, \eq{\ref{eq:int_derivative_app}} generalizes to
\begin{equation}
\label{eq:cont_derivative_app}
\begin{split}
\derivx{\contfunc_{\nm}\regpar{\ppath|\sceneparam}}{\scenescalar_{\unknownindex}} & = \deriv{\scenescalar_{\unknownindex}}\regpar{\pixmeasure_{\nm} \prod_{\segind=0}^{\pathsize-1}{\segfunc_{\segind}}} = \contfunc_{\nm}\regpar{\ppath|\sceneparam}\regpar{ \frac{\derivx{\pixmeasure_{\nm}}{\scenescalar_{\unknownindex}}}{\pixmeasure_{\nm}}+\sum_{\segind=0}^{\pathsize-1}{  \frac{\derivx{\segfunc_{\segind}}{\scenescalar_{\unknownindex}}}{\segfunc_{\segind}}}} \\ 
& = \contfunc_{\nm}\regpar{\ppath|\sceneparam} \cdot \scorefunc_{\unknownindex,\nm}\regpar{\ppath|\sceneparam},
\end{split}
\end{equation}
where
\begin{equation}
\label{eq:total_derivative_factor_app}
\scorefunc_{\unknownindex, \nm}\regpar{\ppath|\sceneparam} \triangleq \frac{\deriv{\scenescalar_{\unknownindex}} \pixmeasure_{\nm}\regpar{\location_{\pathsize-1},\location_{\pathsize}}}{ \pixmeasure_{\nm}\regpar{\location_{\pathsize-1},\location_{\pathsize}}}+ \sum_{\segind=0}^{\pathsize-1}{\frac{\deriv{\scenescalar_{\unknownindex}}\segfunc_{\segind}\regpar{\location_{\segind-1}, \location_{\segind}, \location_{\segind+1}}}{\segfunc_{\segind}\regpar{\location_{\segind-1}, \location_{\segind}, \location_{\segind+1}}}}.
\end{equation}
Plugging back $\derivx{\contfunc_{\nm}\regpar{\ppath|\sceneparam}}{\scenescalar_{\unknownindex}}$ in \eq{\ref{eq:int_derivative_app}}:
\begin{equation}
\label{eq:int_derivative_full_app}
\frac{\partial \forwardmodel_{\nm}\regpar{\sceneparam}}{\partial \scenescalar_{\unknownindex}} =  \intop_{\pathset}{\contfunc_{\nm}\regpar{\ppath|\sceneparam}\cdot \scorefunc_{\unknownindex,\nm}\regpar{\ppath|\sceneparam} d\ppath}.
\end{equation}
\subsection{\ac{PSF} of Scattering Tomography}
\label{sec:psf_st}
We now provide a detailed derivation of $\scorefunc_{\unknownindex, \nm}\regpar{\ppath|\sceneparam}$. The transmittance of the medium on the line segment $\linesegment{\location_{\segind}}{\location_{\segind+1}}$ is
\begin{equation}
    \trans{\location_{\segind}}{\location_{\segind+1}} \approx \exp{\regpar{-\sum_{\voxel=1}^{\Nu}{\regpar{\extair_{\voxel}+\extcloud_{\voxel}}\cdot\distance_{\segind,\voxel}}}},
\end{equation}
Since a pixel response function $\pixmeasure_{\nm}$ does not depend on $\extcloud_{\unknownindex}$:
\begin{equation}
\scorefunc_{\unknownindex, \nm}\regpar{\ppath|\sceneparam} = \sum_{\segind=0}^{\pathsize-1}{\frac{\deriv{\extcloud_{\unknownindex}}\segfunc_{\segind}\regpar{\location_{\segind-1}, \location_{\segind}, \location_{\segind+1}}}{\segfunc_{\segind}\regpar{\location_{\segind-1}, \location_{\segind}, \location_{\segind+1}}}}.
\end{equation}
We evaluate each summand separately. For $\segind=0$ (\eqnopar{\ref{eq:g0K}}), we get:
\begin{equation}
\frac{\deriv{\extcloud_\unknownindex}\segfunc_{0}\regpar{\location_{-1},\location_{0},\location_{1}}}{\segfunc_{0}\regpar{\location_{-1},\location_{0},\location_{1}}} = \frac{\deriv{\extcloud_{\unknownindex}}\squarepar{\Bar{\radiancee}\regpar{\location_{0},\location_{1}}\trans{\location_{0}}{\location_{1}}G\regpar{\location_0,\location_{1}}}}{\Bar{\radiancee}\regpar{\location_{0},\location_{1}}\trans{\location_{0}}{\location_{1}}G\regpar{\location_0,\location_{1}}}.
\end{equation}
Notice that $\trans{\location_{0}}{\location_{1}}$ is the only term that may depend on $\extcloud_{\unknownindex}$, therefore:
\begin{equation}
\label{eq:Sm0}
\frac{\deriv{\extcloud_\unknownindex}\segfunc_{0}\regpar{\location_{-1},\location_{0},\location_{1}}}{\segfunc_{0}\regpar{\location_{-1},\location_{0},\location_{1}}} = \frac{\Bar{\radiancee}\regpar{\location_{0},\location_{1}}\deriv{\extcloud_{\unknownindex}}\squarepar{\trans{\location_{0}}{\location_{1}}}G\regpar{\location_0,\location_{1}}}{\Bar{\radiancee}\regpar{\location_{0},\location_{1}}\trans{\location_{0}}{\location_{1}}G\regpar{\location_0,\location_{1}}} = \frac{\deriv{\extcloud_\unknownindex}\trans{\location_{0}}{\location_{1}}}{\trans{\location_{0}}{\location_{1}}}
\end{equation}
The transmittance derivative is:
\begin{equation}
\label{eq:trans_derivative}
\deriv{\extcloud_\unknownindex}\trans{\location_{0}}{\location_{1}} = 
-\distance_{0, \unknownindex} \cdot \trans{\location_{0}}{\location_{1}}
\end{equation}
Plugging back the transmittance derivative in \eq{\ref{eq:Sm0}}:
\begin{equation}
\label{eq:trans_derivative2}
\frac{\deriv{\extcloud_\unknownindex}\segfunc_{0}\regpar{\location_{-1},\location_{0},\location_{1}}}{\segfunc_{0}\regpar{\location_{-1},\location_{0},\location_{1}}} = -\distance_{0, \unknownindex}
\end{equation}
For $\segind\in\curlpar{1,...,\pathsize-1}$ (\eqnopar{\ref{eq:gk_scatter}}):
\begin{equation}
\frac{\deriv{\extcloud_\unknownindex}\segfunc_{\segind}\regpar{\location_{\segind-1},\location_{\segind},\location_{\segind+1}}}{\segfunc_{\segind}\regpar{\location_{\segind-1},\location_{\segind},\location_{\segind+1}}} = \frac{\deriv{\extcloud_\unknownindex}\squarepar{\anglefunc\regpar{\location_{\segind-1},\location_{\segind},\location_{\segind+1}}\trans{\location_{\segind}}{\location_{\segind+1}}G\regpar{\location_{\segind},\location_{\segind+1}}}}{\anglefunc\regpar{\location_{\segind-1},\location_{\segind},\location_{\segind+1}}\trans{\location_{\segind}}{\location_{\segind+1}}\geomterm\regpar{\location_{\segind},\location_{\segind+1}}}.
\end{equation}
Since $\geomterm\regpar{\location_{\segind},\location_{\segind+1}}$ does not depend on $\extcloud_{\unknownindex}$:
\begin{equation}
\label{eq:gb_deriv}
\frac{\deriv{\extcloud_\unknownindex}\segfunc_{\segind}\regpar{\location_{\segind-1},\location_{\segind},\location_{\segind+1}}}{\segfunc_{\segind}\regpar{\location_{\segind-1},\location_{\segind},\location_{\segind+1}}} = \frac{\deriv{\extcloud_\unknownindex}\squarepar{\anglefunc\regpar{\location_{\segind-1},\location_{\segind},\location_{\segind+1}}\trans{\location_{\segind}}{\location_{\segind+1}}}}{\anglefunc\regpar{\location_{\segind-1},\location_{\segind},\location_{\segind+1}}\trans{\location_{\segind}}{\location_{\segind+1}}}.
\end{equation}
In analogy to \eqs{\ref{eq:trans_derivative},\ref{eq:trans_derivative2}}, the second term is given by:
\begin{equation}
\label{eq:trans_deriv_b}
\frac{\deriv{\extcloud_\unknownindex}\trans{\location_{\segind}}{\location_{\segind+1}}}{\trans{\location_{\segind}}{\location_{\segind+1}}} = 
-\distance_{\segind, \unknownindex}.
\end{equation}
Recall the definition of $\anglefunc$ for a scattering medium (\eqnopar{\ref{eq:fs}}). Then,
\begin{equation}
\begin{split}
\anglefunc\regpar{\location_{\segind-1},\location_{\segind},\location_{\segind+1}}&=\exts\regpar{\location_{\segind}}\cdot\phasefunc\regpar{\location_{\segind},\dangle_{\segind}}\\
&=\exts\regpar{\location_{\segind}}\cdot\frac{\extscloud\regpar{\location_{\segind}}\cdot\phasefunccloud\regpar{\dangle_{\segind-1,\segind}}+\extsair\regpar{\location_{\segind}}\cdot\phasefuncair\regpar{\dangle_{\segind-1,\segind}}}{\exts\regpar{\location_{\segind}}}\\
&=\extscloud\regpar{\location_{\segind}}\cdot\phasefunccloud\regpar{\dangle_{\segind-1,\segind}}+\extsair\regpar{\location_{\segind}}\cdot\phasefuncair\regpar{\dangle_{\segind-1,\segind}}.
\end{split}
\label{eq:fs_scattering}
\end{equation}
Here, $\exts$ is the total scattering coefficient and  $\exts^{\text{c}}, \exts^{\text{a}}$ are the cloud and air scattering coefficients, respectively. We need to write \eq{\ref{eq:fs_scattering}} using the unknowns. If $\location_{\segind}\in\voxelset_{\voxel}$, then
\begin{equation}
\label{eq:unknown_term}
\exts^{\text{c}}\regpar{\location_{\segind}} = \sinscatcloud\extcloud_{\voxel}, \quad \exts^{\text{a}}\regpar{\location_{\segind}} = \sinscatair\extair_{\voxel}.
\end{equation}
Plugging \eq{\ref{eq:unknown_term}} in \eq{\ref{eq:fs_scattering}}:
\begin{equation}
\label{eq:fs_using_unknowns}
\anglefunc\regpar{\location_{\segind-1},\location_{\segind},\location_{\segind+1}} = \sinscatcloud\extcloud_{\voxel}\cdot\phasefunc^{\text{c}}\regpar{\dangle_{\segind-1,\segind}}+\sinscatair\extair_{\voxel}\cdot\phasefunc^{\text{a}}\regpar{\dangle_{\segind-1,\segind}}
\end{equation}
Thus:
\begin{equation}
\label{eq:fs_deriv}
\frac{\deriv{\extcloud_\unknownindex}\anglefunc\regpar{\location_{\segind-1},\location_{\segind},\location_{\segind+1}}}{\anglefunc\regpar{\location_{\segind-1},\location_{\segind},\location_{\segind+1}}}=
\dfrac{\sinscatcloud\phasefunc^{\text{c}}\regpar{\dangle_{\segind-1,\segind}}}{\anglefunc\regpar{\location_{\segind-1},\location_{\segind},\location_{\segind+1}}} =
\regpar{\extcloud_{\voxel}+\dfrac{\sinscatair\phasefunc^{\text{a}}\regpar{\dangle_{\segind-1,\segind}}}{\sinscatcloud\phasefunc^{\text{c}}\regpar{\dangle_{\segind-1,\segind}}}\extair_{\voxel}} ^ {-1}.
\end{equation}
Otherwise\footnote{We note that \citep{loeub2020monotonicity} provides similar derivations. However, the factor $\dfrac{\sinscatair\phasefunc^{\text{a}}\regpar{\dangle_{\segind-1,\segind}}}{\sinscatcloud\phasefunc^{\text{c}}\regpar{\dangle_{\segind-1,\segind}}}$ from \eq{\ref{eq:fs_deriv}} is missing from their derivation.}, if $\location_{\segind}\notin\voxelset_{\voxel}$, then
\begin{equation}
\label{eq:fs_deriv0}
    \frac{\deriv{\extcloud_\unknownindex}\anglefunc\regpar{\location_{\segind-1},\location_{\segind},\location_{\segind+1}}}{\anglefunc\regpar{\location_{\segind-1},\location_{\segind},\location_{\segind+1}}}=0.
\end{equation}
Plugging \eq{\ref{eq:trans_deriv_b}} and \eqs{\ref{eq:fs_deriv},\ref{eq:fs_deriv0}} in \eq{\ref{eq:gb_deriv}}:
\begin{equation}
\frac{\deriv{\extcloud_\unknownindex}\segfunc_{\segind}\regpar{\location_{\segind-1},\location_{\segind},\location_{\segind+1}}}{\segfunc_{\segind}\regpar{\location_{\segind-1},\location_{\segind},\location_{\segind+1}}} = -\distance_{\segind, \unknownindex}
+
\begin{cases}
\regpar{\extcloud_{\voxel}+\dfrac{\sinscatair\phasefunc^{\text{a}}\regpar{\dangle_{\segind-1,\segind}}}{\sinscatcloud\phasefunc^{\text{c}}\regpar{\dangle_{\segind-1,\segind}}}\extair_{\voxel}} ^ {-1} & \text{if $\location_{\segind}\in\voxelset_{\voxel}$} \\
0 & \text{else}
\end{cases}.
\end{equation}
To conclude:
\begin{equation}
\scorefunc_{\unknownindex, \nm}\regpar{\ppath|\sceneparam} = -\sum_{\segind=0}^{\pathsize}{\distance_{\segind,\voxel}}+
\sum_{\segind=0}^{\pathsize-1}{
\begin{cases}
\regpar{\extcloud_{\voxel}+\dfrac{\sinscatair\phasefunc^{\text{a}}\regpar{\dangle_{\segind-1,\segind}}}{\sinscatcloud\phasefunc^{\text{c}}\regpar{\dangle_{\segind-1,\segind}}}\extair_{\voxel}} ^ {-1} & \text{if $\location_{\segind}\in\voxelset_{\voxel}$} \\
0 & \text{else}
\end{cases}}.
\end{equation}

\end{document}